\documentclass[journal]{IEEEtran}
\usepackage{amsmath,amsfonts}
\usepackage[ruled,vlined,linesnumbered]{algorithm2e}
\usepackage{array}
\usepackage[caption=false,font=normalsize,labelfont=sf,textfont=sf]{subfig}
\usepackage{textcomp} 
\usepackage{stfloats}
\usepackage{needspace}
\usepackage{url}
\usepackage{verbatim}
\usepackage{graphicx}
\usepackage{cite}
\usepackage{amsthm}
\newtheorem{theorem}{Theorem}
\newtheorem{lemma}{Lemma}
\newtheorem{assumption}{Assumption}

\newtheorem{remark}{Remark}

\usepackage{threeparttable}

\usepackage{tabularray}
\usepackage{multicol}
\usepackage{multirow}

\usepackage{pifont}
\usepackage{xcolor}
\usepackage{diagbox}
\usepackage{amssymb}
\usepackage{makecell}
\usepackage{booktabs}
\usepackage{xfrac}
\UseTblrLibrary{amsmath}
\usepackage[normalem]{ulem} %

\providecommand{\SHOWCHANGES}{0}
\definecolor{tyqrevcolor}{RGB}{0,105,105} %

\ifnum\SHOWCHANGES=1
  \newcommand{\rev}[1]{\textcolor{red}{#1}}
  \newenvironment{revblock}{\begingroup\color{red}}{\endgroup}
  
  \newenvironment{tyqrevblock}{\begingroup\color{tyqrevcolor}}{\endgroup}
  \newcommand{\old}[1]{\textcolor{blue}{\sout{#1}}}
\else
  \newcommand{\rev}[1]{#1}
  \newenvironment{revblock}{}{}

  \newcommand{\old}[1]{}
\fi

\ifnum\SHOWCHANGES=1
  \newenvironment{revtable}{\begingroup\color{red}}{\endgroup}
\else
  \newenvironment{revtable}{\begingroup}{\endgroup}
\fi

\begin{document}
\bstctlcite{IEEEref:BSTcontrol}

\title{
Preserving Full 6-DOF Actuation Under Abrupt Total Rotor Failures: Passive Fault-Tolerant Flight Control Using a Biaxial-Tilt Hexacopter
}

\author{

Yipeng Yang, Yiqiao Tang, Hao~Zhang, Jinqi~Jiang, Jianfeng~He, Rumo~Chen, Xinghu~Yu, Zhan~Li, Huijun~Gao

}

\maketitle

\begin{abstract}

\begin{revblock}
Abrupt total rotor failures can invalidate nominal allocation despite remaining wrench capability.
This paper develops an allocator-layer passive fault-tolerant control (AL-PFTC) scheme for overactuated vector-thrust aerial vehicles.
Analysis of convexified attainable force and torque spaces at horizontal hover compares biaxial- and uniaxial-tilt configurations and screens fault cases using prescribed reserve thresholds.
Instantaneous wrench loss relative to these reserves characterizes the initial redistribution burden.
AL-PFTC adapts the allocation matrix using a virtual error derived from filtered wrench commands and an onboard external wrench estimate, with adaptive parameter bounds constructed from right inverses.
Joint analysis including estimator residuals and filter dynamics establishes local uniform ultimate boundedness for position and attitude regulation under the stated conditions.
Controller-layer passive fault-tolerant control (CL-PFTC) provides a reference with nominal allocation and observer compensation.
Both share a nominal tracking baseline and run at 600~Hz on an STM32H753 without explicit fault identification, fault-mode switching, or online optimization.
Comparisons show that AL-PFTC satisfies the recovery criterion in more tested hovering cases than CL-PFTC.
Additional experiments using only onboard sensing, including indoor tracking under wind disturbance, outdoor tracking under wind and low-temperature conditions, narrow-frame traversal, and contact-based aerial writing, further validate the robustness of AL-PFTC in complex operational environments.
\end{revblock}

\end{abstract}

\begin{IEEEkeywords}
Aerial systems: mechanics and control, failure detection and recovery, robust/adaptive control of robotic systems, overactuated multirotor.
\end{IEEEkeywords}

\section{Introduction}
\begin{figure}[t]
    \centering
    \captionsetup[subfigure]{font=footnotesize}
     \subfloat[]{
        \includegraphics[width=2.7cm]{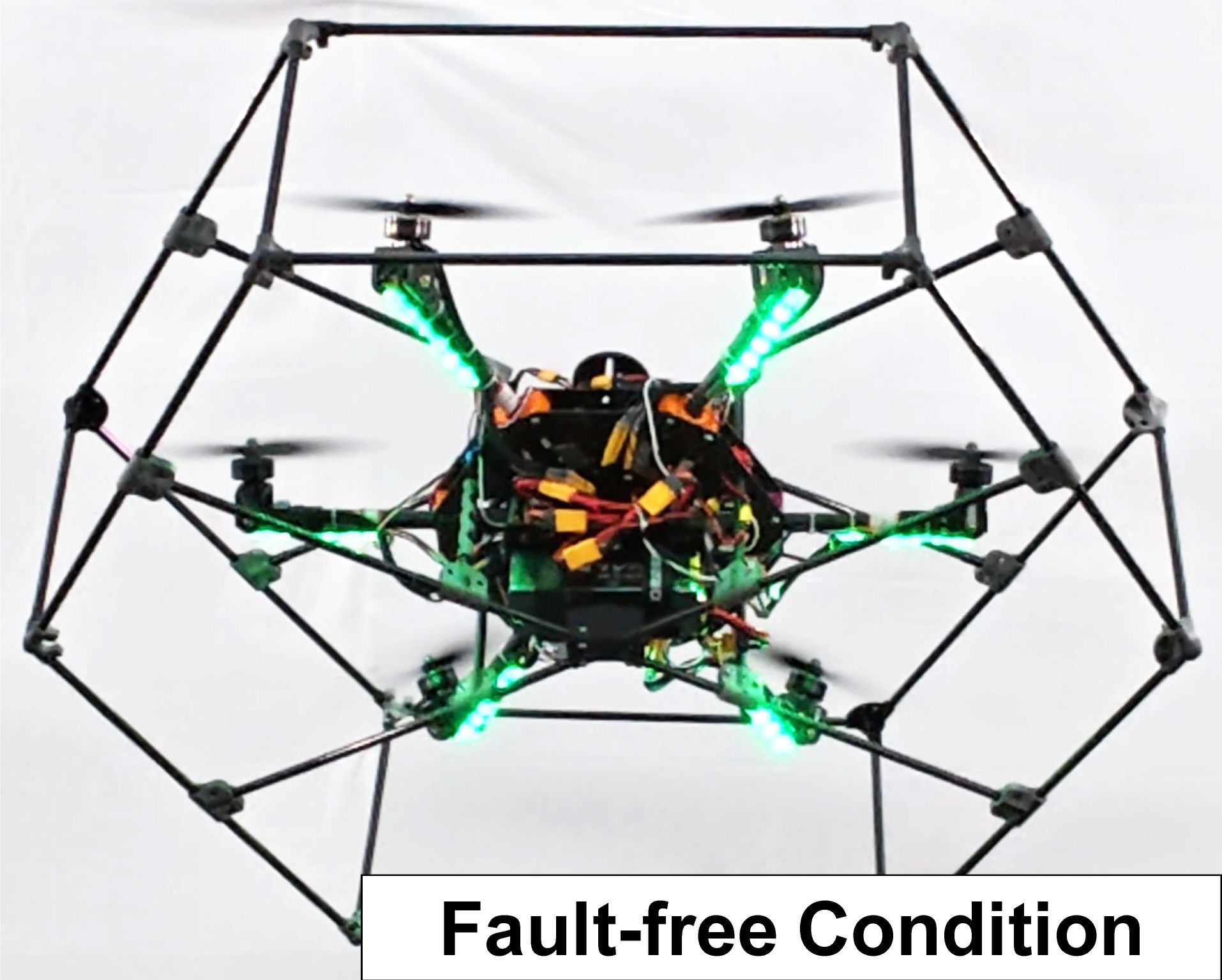}
    }
    \subfloat[]{
        \includegraphics[width=2.7cm]{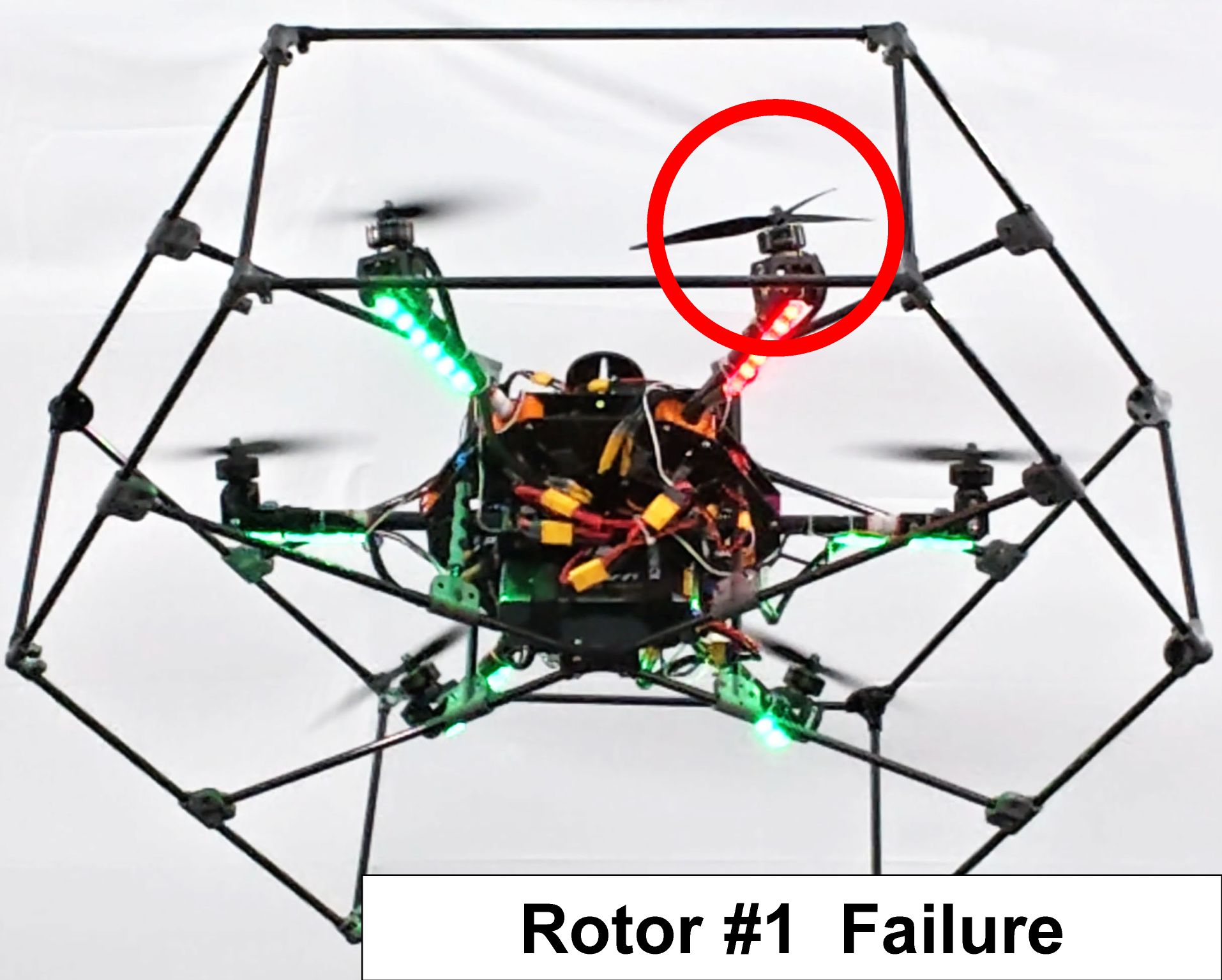}
    }
    \subfloat[]{
        \includegraphics[width=2.7cm]{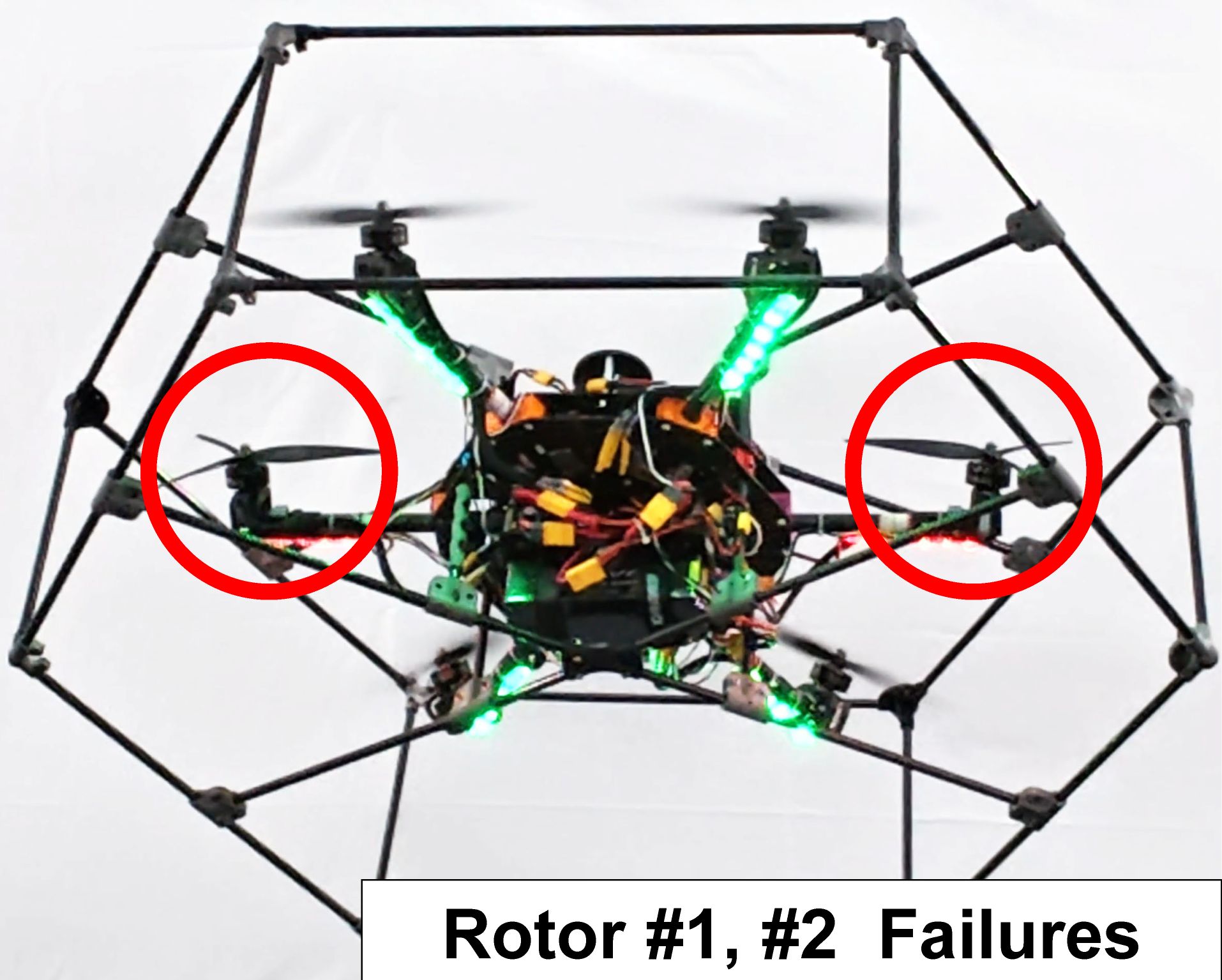}
    }\\
    \subfloat[]{
        \includegraphics[width=2.7cm]{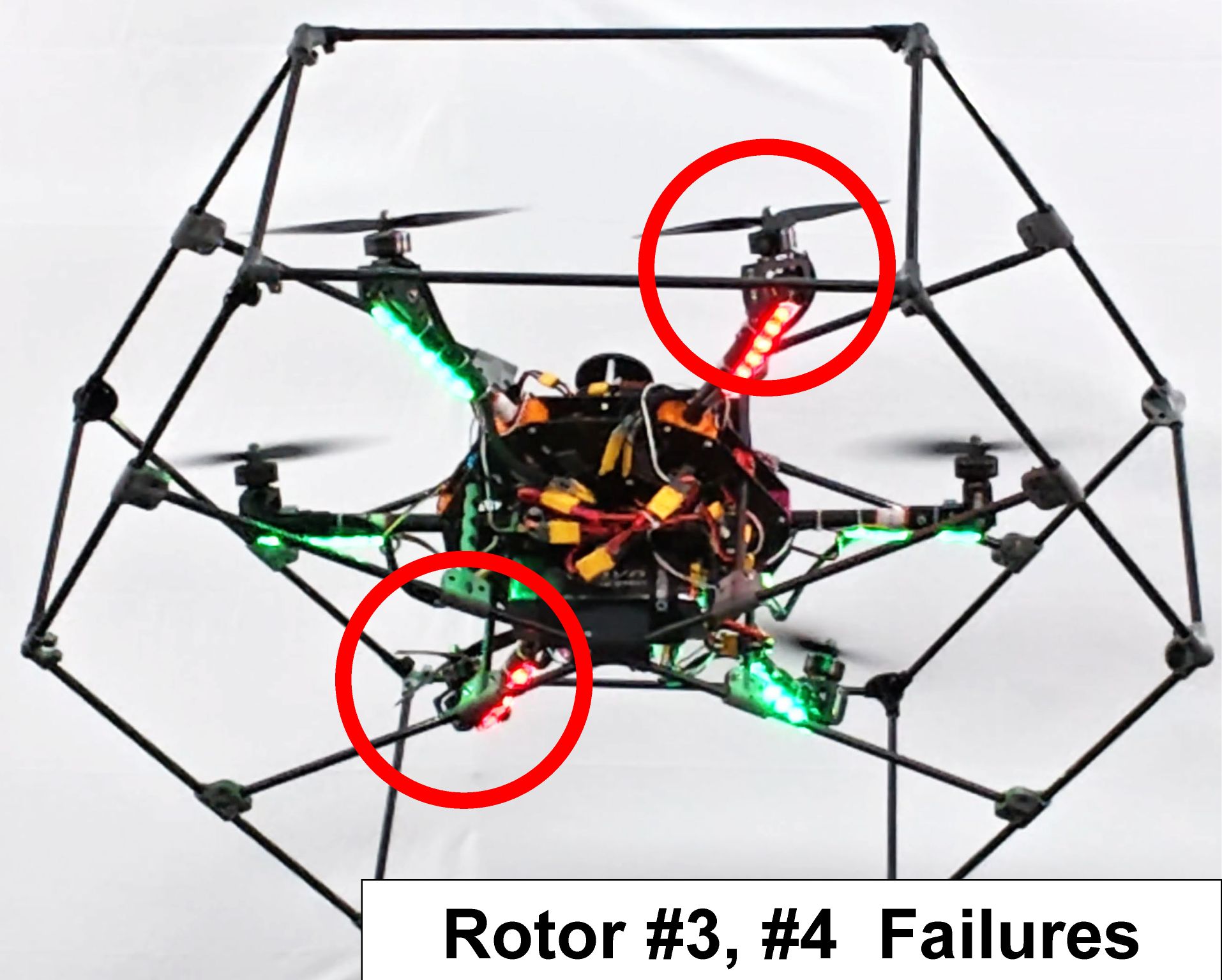}
    }
    \subfloat[]{
        \includegraphics[width=2.7cm]{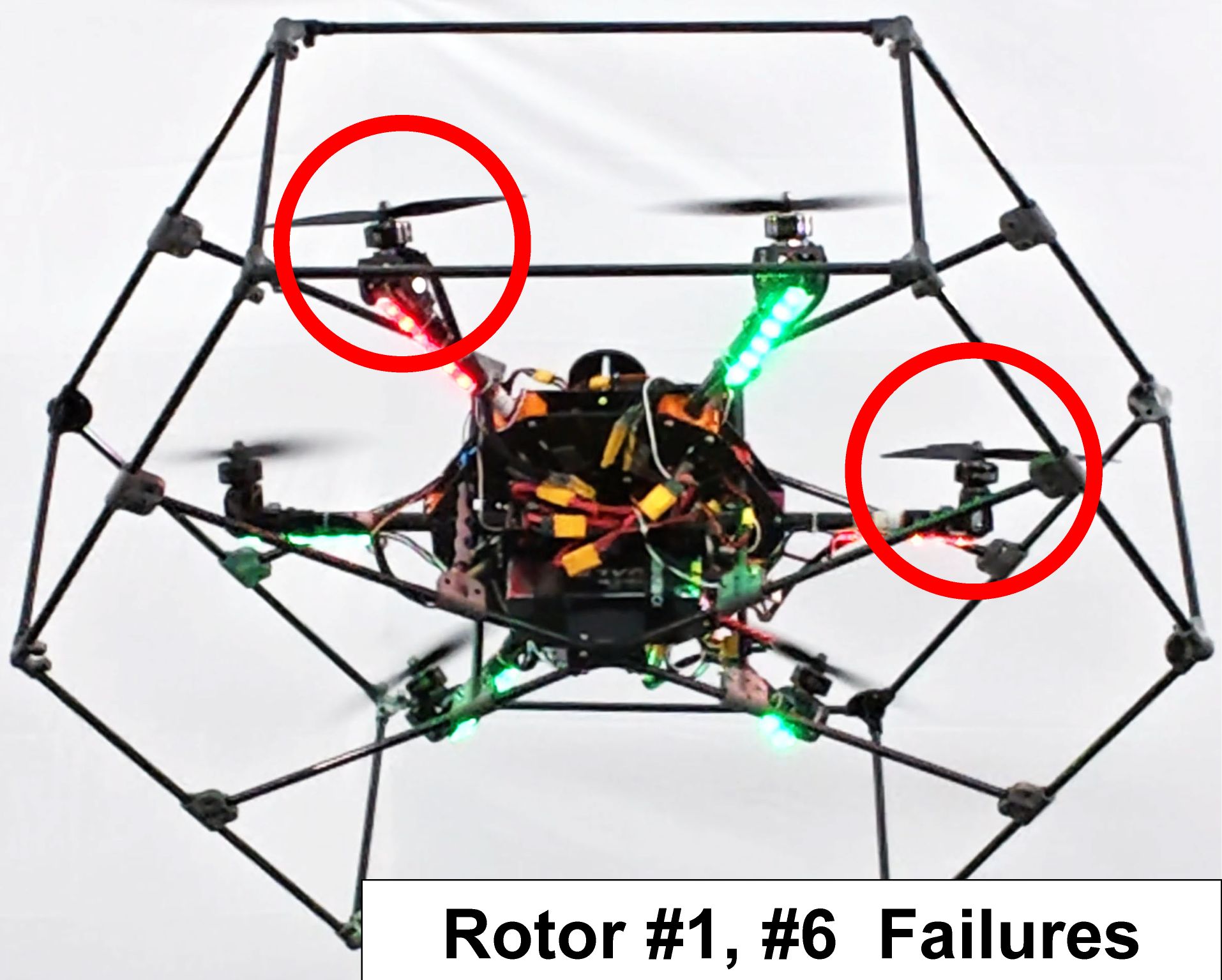}
    }
    \subfloat[]{
        \includegraphics[width=2.7cm]{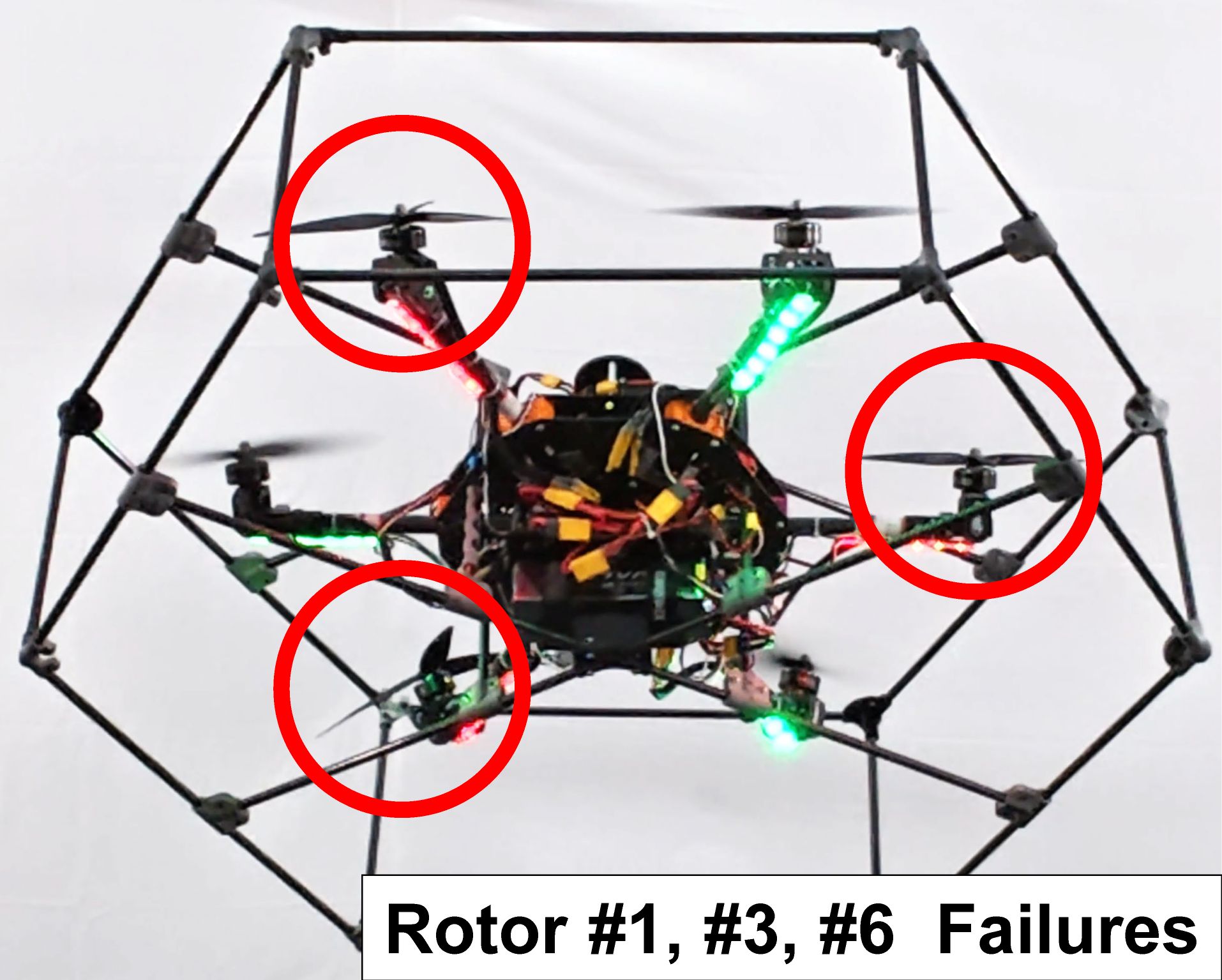}
    }
    \caption{Fault-tolerant hovering of the \rev{biaxial-tilt overactuated hexacopter} (BTO)
    \rev{using allocator-layer passive fault-tolerant control} (AL-PFTC)
    at \rev{a} roll \rev{angle of} $45^\circ$ under different abrupt total rotor failure conditions.
    Red circles \rev{mark} failed rotors.
    (a) Fault-free. (b) \rev{Failure of} rotor \#1. (c) \rev{Failures of rotors} \#1 \rev{and} \#2.
    (d) \rev{Failures of rotors} \#3 \rev{and} \#4. (e) \rev{Failures of rotors} \#1 \rev{and} \#6. (f) \rev{Failures of rotors} \#1, \#3, \rev{and} \#6.
    See the supplementary video for complete flight demonstrations.
    Anonymous supplementary \rev{review} materials are available at \emph{https://anonymous.4open.science/w/tilthex-rotorout-2883}.}
    \label{fig:al-bto-hovering-fault-cases}
\end{figure}

\begin{revblock}
\textcolor{black}{Unmanned aerial vehicles (UAVs) have demonstrated a broad range of capabilities,
including collaborative exploration~\cite{zhou2023racer}, aerial interaction and manipulation~\cite{ollero2022present}, payload transportation~\cite{wang2024impactaware}, and omnidirectional flight enabled by reconfigurable aerial platforms~\cite{allenspach2020design,zheng2020tiltdrone}.}
\textcolor{black}{Demanding maneuvers such as aerial robotic perching~\cite{zhang2019perching} further highlight the} importance of accurate motion control during task execution.
However, abrupt total rotor failures can destabilize the vehicle and  interrupt ongoing tasks.
This work considers flight tasks that require independent position and attitude control after rotor failures, and aims to restore this control using the remaining actuators.
The identities of the failed rotors are \textcolor{black}{unknown to the controller}.
Fig.~\ref{fig:al-bto-hovering-fault-cases} shows representative hovering results at a roll angle of $45^\circ$ following abrupt total rotor failures.
\end{revblock}

\begin{figure}[t]
    \centering
    \captionsetup[subfigure]{font=footnotesize}
    \subfloat[]{
        \includegraphics[width=2.7cm]{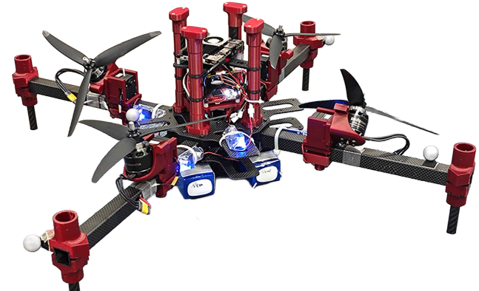}
    }
    \subfloat[]{
        \includegraphics[width=2.7cm]{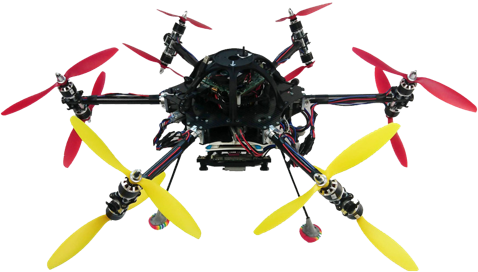}
    }
    \subfloat[]{
        \includegraphics[width=2.7cm]{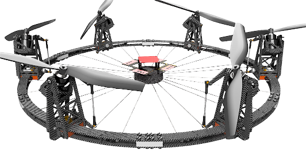}
    }\\
    \subfloat[]{
        \includegraphics[width=2.7cm]{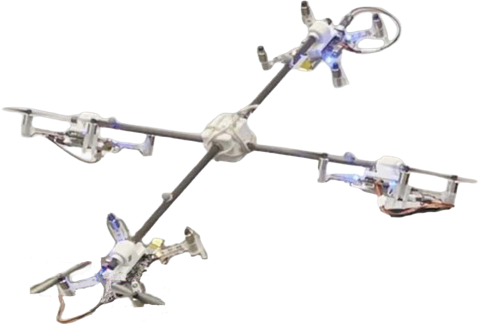}
    }
    \subfloat[]{
        \includegraphics[width=2.7cm]{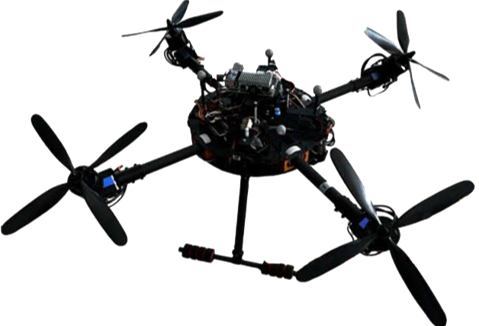}
    }
    \subfloat[]{
        \includegraphics[width=2.7cm]{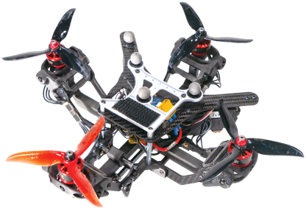}
    }
    \caption{\rev{Representative aerial robots equipped with tilting-rotor actuator units}. (a) Quadcopter with UTAUs~\cite{li2024servo}. (b) Hexacopter with UTAUs~\cite{allenspach2020design}. (c) Hexacopter with synchronized UTAUs~\cite{ryll2016modeling}. (d) Quadcopter with UTAUs \rev{based on micro quadcopters}~\cite{su2024faulttolerant}. (e) Quadcopter with BTAUs~\cite{yang2024new}. (f) Quadcopter with synchronized BTAUs~\cite{zheng2020tiltdrone}.}
    \label{fig:tilting-actuator-platforms}
\end{figure}

\begin{revblock}
Restoring independent position and attitude control after rotor failures is physically constrained by the forces and torques that the remaining actuators can generate.
In conventional coplanar multirotors, the thrust direction is fixed relative to the airframe, coupling translational motion to vehicle attitude.
For quadrotors, recovery methods can retain position control after rotor failures while relinquishing yaw control\textcolor{black}{~\cite{sun2021incremental,nan2022nonlinear}}.
Platforms with tilting rotors can instead vary the thrust direction relative to the airframe\textcolor{black}{~\cite{chen2024design}}.
These platforms use uniaxial-tilt actuator units (UTAUs) or biaxial-tilt actuator units (BTAUs), as illustrated in Fig.~\ref{fig:tilting-actuator-platforms}.
A UTAU constrains its thrust vector to a plane\textcolor{black}{~\cite{allenspach2020design,kamel2018voliro,li2024servo,ryll2016modeling,su2024faulttolerant}}, whereas a BTAU adds an orthogonal tilt axis and permits thrust vectors outside that plane\textcolor{black}{~\cite{zheng2020tiltdrone,singh2022quadplus,yang2024new}}.
For a given rotor arrangement, the second tilt axis provides additional freedom in allocating the required forces and torques among the actuators.
Both types have been used in platforms that independently control position and attitude in all six degrees of freedom (6-DOF)\textcolor{black}{~\cite{ryll2016modeling,yang2024new}}.
In such overactuated tilting-rotor aerial robots, the actuator redundancy can preserve full 6-DOF actuation under certain rotor failure conditions, providing a physical basis for motion recovery.
Realizing this recovery requires coordinating the remaining actuators to compensate for the lost contributions.
\end{revblock}

\begin{revblock}
Fault-tolerant control (FTC) addresses this problem, with approaches differing in their reliance on explicit fault identification.
Active approaches use fault diagnosis to reconfigure the controller or allocator\textcolor{black}{~\cite{jiang2012faulttolerant,gao2015survey,nguyen2021design}}.
Lee et al.\textcolor{black}{~\cite{lee2020failsafe}} restore 4-DOF flight after a single motor failure through fault detection and control reconfiguration.
Passive FTC (PFTC) instead accommodates faults \textcolor{black}{without explicit fault identification, typically through robust design or disturbance rejection.}
Ke et al.\textcolor{black}{~\cite{ke2023uniform}} develop a uniform passive controller for rotor failures in quadrotors, while Estrada et al.\textcolor{black}{~\cite{estrada2024passive}} use Lyapunov redesign with sliding modes to construct a passive controller.

In overactuated aerial platforms, control allocation distributes the forces and torques requested by the flight controller among redundant actuators.
Adjusting this distribution can help compensate for the wrench mismatch caused by rotor failures.
Nullspace methods adjust actuator commands to handle actuator limits while preserving the wrench predicted by the allocation model\textcolor{black}{~\cite{su2021nullspacebased}}.
These methods have also been extended to account for downwash interactions\textcolor{black}{~\cite{su2022downwashaware}}.
Su et al.\textcolor{black}{~\cite{su2024faulttolerant}} combine nullspace allocation with controller reconfiguration and disturbance compensation to accommodate rotor failures.
Their method assumes accurate identification of the failed propellers.
Mehndiratta et al.\textcolor{black}{~\cite{mehndiratta2021recedinga}} likewise use fault information to reconfigure a nonlinear model predictive controller and recover pose control after a single propeller failure in simulation.

Adaptive methods provide another way to accommodate uncertain actuator effectiveness.

Park et al.\textcolor{black}{~\cite{parkAdaptiveFaultTolerant2021}}
use an $\mathcal L_1$ adaptive scheme to estimate uncertainty in the vehicle dynamics and actuator effectiveness.
At the allocation level,
Tohidi et al.\textcolor{black}{~\cite{tohidi2020adaptive}} update the allocation matrix without explicitly estimating actuator effectiveness. 
Measured or estimated generalized forces provide feedback to a virtual allocation system, 
whose error relative to a reference model drives allocation-matrix updates within prescribed parameter bounds. 
The formulations in\textcolor{black}{~\cite{tohidi2020adaptive,tohidiDiscreteAdaptiveControl2021}} assume strictly positive actuator effectiveness.

Complete actuator failures have also been addressed by combining adaptive fault-tolerant control with control allocation. 
For spacecraft attitude tracking, Shen et al.~\cite{shen2015inertiafree} combine inertia-free adaptive control with allocation based on estimated actuator effectiveness, 
while their robust allocation approach~\cite{shen2017robust} explicitly accounts for fault-estimation uncertainty within the allocation process. 
Wang and Zhang~\cite{wang2018adaptive} combine adaptive sliding-mode control with control reallocation and demonstrate octorotor flight under a complete actuator failure,
 both alone and combined with partial effectiveness loss in another actuator. 
 These approaches use actuator fault information in the allocation process.

Without explicit fault identification, the control problem is how to use the remaining actuation capability to recover the required motion.
After total rotor failures, even if the wrench map retains full row rank, the required equilibrium wrench may still be infeasible under actuator limits\textcolor{black}{~\cite{franchi2018fullpose,du2015controllability,michieletto2018fundamental}}.
Beyond equilibrium feasibility, the remaining force and torque
reserves also matter for continued tracking and disturbance rejection.
With nominal allocation retained, compensation in the flight controller changes the requested wrench while its distribution among actuators remains based on the nominal model.
This may require larger commands in directions weakened by failure and drive individual actuators to their limits, even when another feasible allocation exists.

In this work, we consider rotor failures under which the aerial robot retains full 6-DOF actuation.
We study this problem on a reconfigurable \textcolor{black}{biaxial-tilt overactuated hexacopter (BTO)}, extending the platform with four BTAUs in\textcolor{black}{~\cite{yang2024new}} to six independently commanded units.
Locking the outer tilt axes yields a \textcolor{black}{uniaxial-tilt overactuated hexacopter (UTO)}, which shares the airframe, sensing, and computing hardware.
We first analyze \textcolor{black}{the attainable wrench space (AWS)} of the BTO and UTO at horizontal hover to 
identify fault conditions that retain prescribed force and torque reserves for continued task execution.
Within this scope, we develop the \textcolor{black}{allocator-layer PFTC (AL-PFTC)}, 
which adapts the allocation matrix to compensate for wrench mismatch after rotor failures.
\textcolor{black}{Controller-layer PFTC (CL-PFTC)} serves as a reference, retaining nominal allocation and using a \textcolor{black}{linear extended state observer (LESO)} to compensate for lumped disturbances.
Both methods share the nominal \textcolor{black}{high-order fully actuated (HOFA)} tracking baseline.
Neither method requires online optimization or receives the identities of the failed rotors.

\textcolor{black}{The main contributions are summarized as follows:}

1) \textbf{\textit{Adaptive allocation for passive fault tolerance}}:
We develop the AL-PFTC framework to support continued pose control under abrupt total rotor
failures without explicit fault identification.
We construct common offline parameter bounds for the prescribed effectiveness set and use filtered external-wrench estimates obtained from onboard motion measurements as allocation-layer feedback.
The analysis retains estimator residuals and filter dynamics, yielding an explicit sufficient bandwidth condition for allocation-error dissipation.

2) \textbf{\textit{Characterization of actuation capability under rotor failures}}:
We quantify the post-failure force and torque reserves of BTO and UTO using convexified attainable force and torque spaces at horizontal hover.
Furthermore, we provide a reserve-based fault set for the adaptive design aimed at continued task execution.
An offline fault-severity assessment further enables comparison of the
initial redistribution burden across the retained BTO cases.

3) \textbf{\textit{Experimental validation of fault-tolerant flight and task execution}}:
We validate AL-PFTC through comparative flight experiments and task-level demonstrations.
Comparisons on the same hardware show that AL-PFTC achieves hovering recovery in more tested fault cases than CL-PFTC.
In the task-level trials, the BTO with AL-PFTC demonstrates 6-DOF tracking under wind disturbances and completes narrow-frame traversal at nonzero roll and contact-based aerial writing under unmeasured contact disturbances, where an abrupt total rotor failure is injected during execution and sustained until completion.
The onboard implementation operates at 600 Hz without online optimization.

\textcolor{black}{The rest of this article is organized as follows.
Section~II provides the dynamic model of BTO.
The comparative AWS analysis is discussed in Section~III.
In Section~IV, the CL-PFTC framework is formulated \rev{as a reference}.
Section~V details the design and stability analysis of the AL-PFTC framework.
In Section~VI, we present the simulation results.
Section~VII reports the experimental results, including hovering tests, trajectory-tracking tests, and autonomous flight-task demonstrations under various fault conditions.
Finally, Section~VIII draws conclusions and discusses future work.}
\end{revblock}

\section{Dynamic Modeling} \label{sec:dynamics-modeling}
\begin{figure}[htbp]
	\centering
	\centering
    \includegraphics[width=8.5cm]{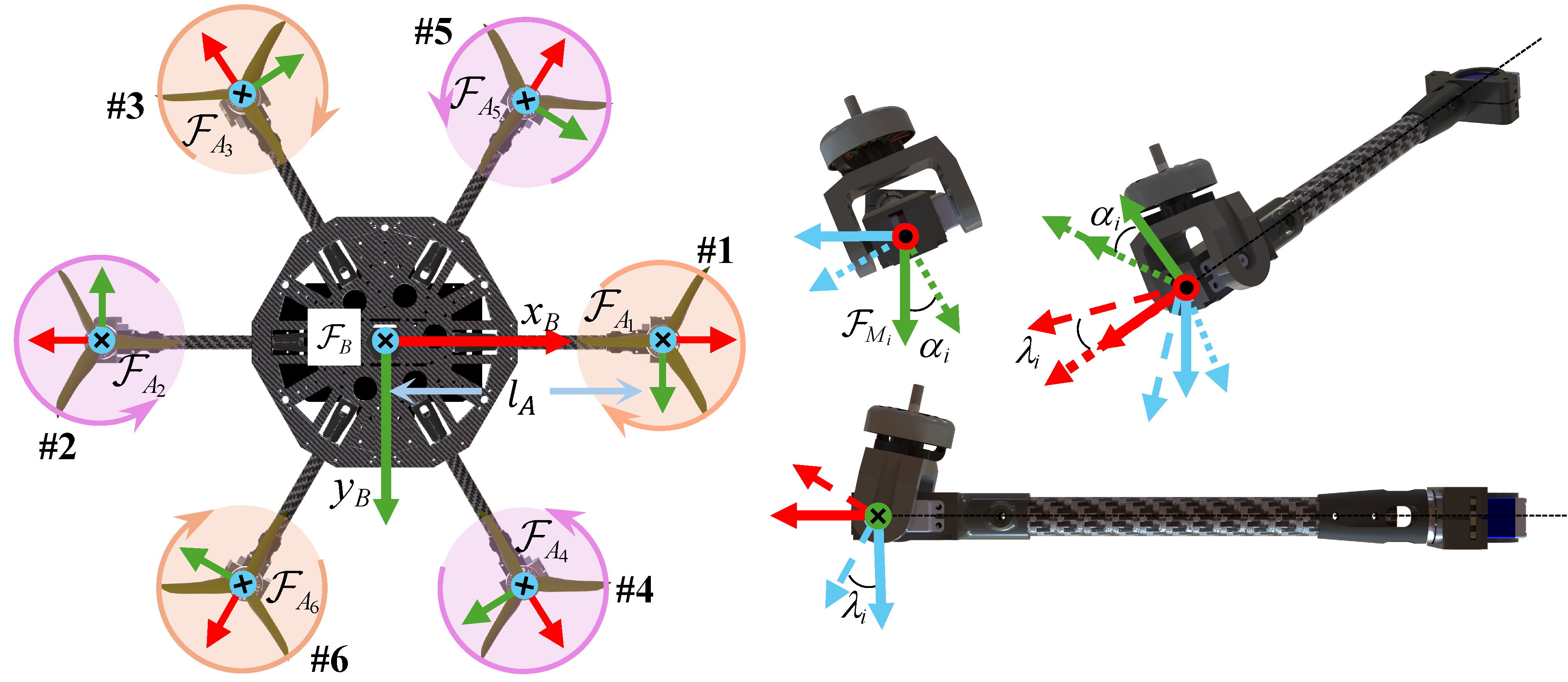}
	\caption{The structure and coordinate definitions of the BTO prototype.}
	\label{fig:bto-structure-coordinates}
\end{figure}

\begin{revblock}
    This section develops an aerodynamic wrench model for each BTAU and a rigid-body model of the BTO.
    Table~{\ref{tab:dynamics-main-notation}} summarizes the main symbols and notation in this section.
    Fig.~\ref{fig:bto-structure-coordinates} defines the north-east-down (NED) inertial frame
    $\mathcal F_E=\{O_E,x_E,y_E,z_E\}$ and the forward-right-down (FRD) body frame
    $\mathcal F_B=\{O_B,x_B,y_B,z_B\}$, with $O_B$ at the geometric center of the airframe.
\end{revblock}
\begin{table}[t]
\caption{\rev{Main Symbols and Definitions for Dynamic Modeling}}
\label{tab:dynamics-main-notation}
\suppressfloats[t]
\centering
\footnotesize
\setlength{\tabcolsep}{2pt}
\renewcommand{\arraystretch}{1.00}
    \begin{revtable}
        \begin{tabular}{@{}
            >{\centering\arraybackslash}m{0.15\columnwidth}
            >{\raggedright\arraybackslash}m{0.77\columnwidth}
            @{}}
        \toprule
        \textbf{Symbol} & \textbf{Definition} \\
        \midrule

        $\boldsymbol p$
        & Position of $O_B$ expressed in $\mathcal F_E$;
        $\boldsymbol p\in\mathbb R^3$ [m]. \\

        $\boldsymbol q$
        & Unit quaternion of $\mathcal F_B$ relative to $\mathcal F_E$;
        $\boldsymbol q\in\mathbb S^3$. \\

        $\boldsymbol\omega$
        & Angular velocity of $\mathcal F_B$ relative to $\mathcal F_E$, expressed in $\mathcal F_B$;
        $\boldsymbol\omega\in\mathbb R^3$ [rad/s]. \\

        $ m $
        & Total mass of the BTO [kg]. \\

        $\mathbf J_B$
        & Inertia tensor about $O_B$, expressed in $\mathcal F_B$;
        $\mathbf J_B\in\mathbb R^{3\times3}$ [$\mathrm{kg\cdot m^2}$]. \\

        $\alpha_i$
        & Tilt angle of the inner servo in actuator unit $i$ [rad]. \\

        $\lambda_i$
        & Tilt angle of the outer servo in actuator unit $i$ [rad]. \\

        $\dot{\sigma}_i$
        & Angular speed of rotor $i$ [rad/s]. \\

        $\mathbf B_0$
        & Nominal control effectiveness matrix;
        $\mathbf B_0\in\mathbb R^{6\times18}$. \\

        ${}^{B}\!\boldsymbol w_{\mathrm{ae}}$
        & Actual aerodynamic wrench;
        ${}^{B}\!\boldsymbol w_{\mathrm{ae}}\in\mathbb R^6$. \\

        ${}^{B}\!\boldsymbol w_d$
        & External and unmodeled disturbance wrench;
        ${}^{B}\!\boldsymbol w_d\in\mathbb R^6$. \\

        $\mathbb I$
        & Actuator index set, $\mathbb I=\{1,\ldots,6\}$. \\

        $\boldsymbol\Lambda$
        & Rotor effectiveness matrix;
        $\boldsymbol\Lambda\in\mathbb R^{18\times18}$. \\
        
        ${}^{X}(\cdot)$
        & Quantity expressed in frame $\mathcal F_X$. \\

        ${}^{X}\!\mathbf R_Y$
        & Rotation matrix from $\mathcal F_Y$ to $\mathcal F_X$;
            ${}^{X}\!\mathbf R_Y\in\mathrm{SO}(3)$. \\

        $[\boldsymbol x]_\times$
        & Skew-symmetric matrix of vector $\boldsymbol x$,
            $[\boldsymbol x]_\times\boldsymbol y
            =\boldsymbol x\times\boldsymbol y$. \\

        $\operatorname{col}(\cdot)$
        & Operator for vertical concatenation of vectors. \\

        $\operatorname{blkdiag}(\cdot)$
        & Operator for block-diagonal concatenation. \\

        \bottomrule
        \end{tabular}
    \end{revtable}
\end{table}

\subsection{Aerodynamic Wrench Modeling for BTAUs}\label{subsec:btau-aerodynamic-wrench}

\begin{revblock}
    This subsection models the wrench generated by each BTAU and combines the six contributions while accounting for reduced rotor effectiveness.

    As shown in Fig.~\ref{fig:bto-structure-coordinates}, the six BTAUs are mounted in the
    $x_B$-$y_B$ plane at azimuths
    $\delta_i(\delta_1=0,\delta_2=\pi,\delta_3=-\tfrac{2\pi}{3},\delta_4=\tfrac{\pi}{3},\delta_5=-\tfrac{\pi}{3},\delta_6=\tfrac{2\pi}{3})$.
    Frame $\mathcal F_{A_i}$ is attached to BTAU $i$, with installation rotation
    ${}^{B}\!\mathbf R_{A_i}=\mathbf R_z(\delta_i)$.
    Let $\boldsymbol r_A=[l_A,0,0]^\top$, where $l_A$ is the distance from $O_B$
    to the origin of $\mathcal F_{A_i}$.
    Then ${}^{B}\!\boldsymbol r_{A_i}={}^{B}\!\mathbf R_{A_i}\boldsymbol r_A$
    is the position of that origin expressed in $\mathcal F_B$.
    Relative to $\mathcal F_{A_i}$, the rotor frame $\mathcal F_{M_i}$ is obtained by
    successive intrinsic rotations about the local $x$- and $y$-axes, yielding
    ${}^{A_i}\!\mathbf R_{M_i}=\mathbf R_x(\alpha_i)\mathbf R_y(\lambda_i)$.
\end{revblock}

\rev{The thrust and reaction-torque magnitudes of rotor $i$ are
$T_i=c_f\dot{\sigma}_i^2$ and $\tau_i=c_t\dot{\sigma}_i^2$,}
where $c_f$ and $c_t$ are the corresponding aerodynamic coefficients.
\rev{Expressed in $\mathcal F_{A_i}$, the force and reaction torque are}
\begin{revblock}
    \begin{equation}
            \begin{aligned}
            {}^{A_i}\!\boldsymbol f_{\mathrm{ae}} &= {}^{A_i}\!\mathbf R_{M_i}\begin{bmatrix}
                                                                            0 \\[-1mm]
                                                                            0 \\[-1mm]
                                                                            -T_i \\
                                                                    \end{bmatrix}            
                                         =  \begin{bmatrix}
                                                -\sin\lambda_i\\[-1mm]
                                                \sin\alpha_i\cos\lambda_i\\[-1mm]
                                                -\cos\alpha_i\cos\lambda_i
                                            \end{bmatrix} T_i \\
            {}^{A_i}\!\boldsymbol t_{\mathrm{ae}} &= \epsilon_i c_\tau {}^{A_i}\!\boldsymbol f_{\mathrm{ae}},\qquad \boldsymbol \epsilon=(1,-1,1,-1,-1,1)
            \end{aligned}
        \label{eq:btau-aerodynamic-force-torque}
    \end{equation}
where $c_\tau\triangleq\tfrac{c_t}{c_f}$ is the torque ratio, and
$\epsilon_i$ represents the rotation direction of rotor $i$.
When viewed from the $-z_{M_i}$ side along the $+z_{M_i}$ direction,
$\epsilon_i=1$ and $\epsilon_i=-1$ denote clockwise and counterclockwise rotation, respectively.
\end{revblock}

Transforming \eqref{eq:btau-aerodynamic-force-torque} to $\mathcal F_B$
gives the force and torque generated by BTAU $i$:
\begin{revblock}
    \begin{equation}
            \begin{aligned}
                &{}^{B}\!\boldsymbol{f}_{\mathrm{ae},i}={}^{B}\!\mathbf{R}_{A_i}{}^{A_i}\!\boldsymbol{f}_{\mathrm{ae}}
            \\
                &{}^{B}\!\boldsymbol{t}_{\mathrm{ae},i}={}^{B}\!\mathbf{R}_{A_i} {}^{A_i}\!\boldsymbol{t}_{\mathrm{ae}}+{}^{B}\!\boldsymbol{r}_{A_i}\times{}^{B}\!\boldsymbol{f}_{\mathrm{ae},i}
            \end{aligned}
        \label{eq:btau-body-frame-wrench}
    \end{equation}
\end{revblock}

\begin{revblock}
    Summing the six BTAU wrenches gives
    \begin{equation}
        \begin{aligned}
            {}^{B}\!\boldsymbol w_{\mathrm{ae}} &= \sum_{i=1}^{6} \eta_i\operatorname{col} \left({}^{B}\boldsymbol f_{\mathrm{ae},i},{}^{B}\boldsymbol t_{\mathrm{ae},i}\right)
                                        = \mathbf B_0 \boldsymbol\Lambda \boldsymbol u_f^c \\
            \boldsymbol u_f^c &= \operatorname{col}\left({}^{A_1}\!\boldsymbol f_{\mathrm{ae}},\ldots, {}^{A_6}\!\boldsymbol f_{\mathrm{ae}}\right), 
            \mathbf B_0 = \begin{bmatrix}
                            \mathbf b_1 \!&\! \cdots \!&\! \mathbf b_6
                        \end{bmatrix}\\
            \mathbf b_i &= \begin{bmatrix}
                                {}^{B}\!\mathbf R_{A_i}\\
                                {}^{B}\!\mathbf R_{A_i}\left(\epsilon_i c_\tau\mathbf I_3+[\boldsymbol r_A]_\times\right)
                            \end{bmatrix} \in \mathbb R^{6\times3} \\
            \boldsymbol\Lambda &= \operatorname{blkdiag}\left(\eta_1\mathbf I_3,\ldots,\eta_6\mathbf I_3\right)
        \end{aligned}
    \label{eq:bto-total-aerodynamic-wrench}
    \end{equation}
where ${}^{B}\!\boldsymbol w_{\mathrm{ae}} \in \mathbb R^{6}$ is the actual aerodynamic wrench. 
The allocator output $\boldsymbol u_f^c\in\mathbb R^{18}$ stacks the six BTAU force commands,
which are converted into tilt-angle and rotor-speed references tracked by the actuator inner loops.
The matrix $\mathbf B_0\in \mathbb R^{6\times 18}$ maps the BTAU force commands to the nominal wrench.
The matrix $\boldsymbol\Lambda\in \mathbb R^{18\times 18}$ scales each force command by the corresponding rotor effectiveness $\eta_i\in[0,1]$.
An abrupt total failure of rotor $i$ changes $\eta_i$ from one to zero.
\end{revblock}
\subsection{Dynamic Modeling of the BTO}
\begin{revblock}
    The BTO is modeled as a rigid body governed by the following Newton--Euler equations:
\end{revblock}
\begin{revblock}
    \begin{equation}
            \begin{aligned}
            \mathbf M_B\!\begin{bmatrix}
                            \ddot{\boldsymbol p}     \\
                            \dot{\boldsymbol\omega} \\
                         \end{bmatrix} \!&=\! {}^{B}\!\boldsymbol w_{\mathrm{gra}} + {}^{B}\!\boldsymbol w_{\mathrm{rot}}
                                        + {}^{B}\!\boldsymbol w_{\mathrm{ae}} + {}^{B}\!\boldsymbol w_d \\
            \mathbf M_B \!&=\! \begin{bmatrix}
                                m {}^{E}\!\mathbf R_B^{\top}                                              \!&\! -m[{}^{B}\!\boldsymbol r_{\mathrm{com}}]_\times \\
                                m[{}^{B}\!\boldsymbol r_{\mathrm{com}}]_\times {}^{E}\!\mathbf R_B^{\top} \!&\! \mathbf J_B
                            \end{bmatrix} \in \mathbb R^{6\times 6}\\
            {}^{B}\!\boldsymbol w_{\mathrm{gra}} \!&=\! \operatorname{col}\left(
                                                                m {}^{B}\!\boldsymbol g_E,
                                                                m [{}^{B}\!\boldsymbol r_{\mathrm{com}}]_\times{}^{B}\!\boldsymbol g_E
                                                            \right) \in \mathbb R^6\\
            {}^{B}\!\boldsymbol w_{\mathrm{rot}} \!&=\! \operatorname{col}\left(
                                                                -m[\boldsymbol\omega]_\times^2{}^{B}\!\boldsymbol r_{\mathrm{com}},
                                                                -[\boldsymbol\omega]_\times\mathbf J_B\boldsymbol\omega
                                                            \right) \in \mathbb R^6
            \end{aligned}
    \label{eq:bto-newton-euler-dynamics}
    \end{equation}
where ${}^{B}\!\boldsymbol r_{\mathrm{com}}\in\mathbb R^3$ is the center-of-mass 
position relative to $O_B$, and ${}^{E}\!\mathbf R_B$ is the rotation matrix
associated with $\boldsymbol q$.
The body-frame gravity vector is
${}^{B}\!\boldsymbol g_E={}^{B}\!\mathbf R_E[0,0,\mathrm g]^\top\in\mathbb R^3$,
with $\mathrm g=9.81\,\mathrm{m/s^2}$.
${}^{B}\!\boldsymbol w_{\mathrm{gra}}$ is the gravity wrench, and
${}^{B}\!\boldsymbol w_{\mathrm{rot}}$ contains the centrifugal-force and gyroscopic-torque terms.
\end{revblock}

\begin{revblock}
    Following\textcolor{black}{~\cite{yang2024new}}, the bounded disturbance wrench
    ${}^{B}\!\boldsymbol w_d\in\mathbb R^6$ collects residual effects of
    the angular velocities and accelerations of the tilting mechanisms and rotors,
    tilt-induced variations in ${}^{B}\!\boldsymbol r_{\mathrm{com}}$ and $\mathbf J_B$,
    actuator dynamics, unmodeled aerodynamics, wind, and other external disturbances.
    Total rotor failures caused by faults in the drive system are modeled by effectiveness losses in
    \eqref{eq:bto-total-aerodynamic-wrench}, with residual actuator and aerodynamic
    effects included in ${}^{B}\!\boldsymbol w_d$\textcolor{black}{~\cite{sun2021incremental,ke2023uniform,nguyen2021design}}.
    For prescribed symmetric blade separation at $6000\,\mathrm{rpm}$,
    Abaqus impulse estimates bound nominal vehicle speed and body-rate changes by
    $0.16\,\mathrm{mm/s}$ and $0.11\,\mathrm{deg/s}$, respectively,
    supporting omission of the mechanical impulse.
\end{revblock}

\section{AWS Analysis with Faults}\label{sec:aws-fault-analysis}
\begin{revblock}
    The attainable wrench space (AWS) contains all wrenches achievable under actuator constraints.
    Using the nominal parameters in Table~\ref{tab:aws-nominal-parameters},
    this section compares the wrench reserves remaining after rotor failures,
    identifies fault cases that meet prescribed reserve thresholds,
    and assesses fault severity using loss-to-margin ratios.
\end{revblock}
\begin{table}[!htb]
    \caption{Nominal Parameters of the BTO}
    \label{tab:aws-nominal-parameters}
    \centering
    \footnotesize
    \setlength{\tabcolsep}{2pt}
    \renewcommand{\arraystretch}{1.00}
    \begin{tabular}{@{}
        >{\centering\arraybackslash}m{0.2\columnwidth}
        >{\raggedright\arraybackslash}m{0.5\columnwidth}
        @{}}
        \toprule
        \textbf{Parameter}                      & \textbf{Value}           \\
        \midrule

        $ m $                                   &
        $3.2\,\mathrm{kg}$                                                  \\

        $\rev{\mathbf{J}_B}$                          &
        $\mathrm{diag}\left(0.083,0.088,0.145\right)\,\mathrm{kg\cdot m^2}$ \\

        $l_A$                                   &
        $0.23\,\mathrm{m}$                                                  \\

        $c_f$                                   &
        $\rev{2.46\times 10^{-6}\,\mathrm{N/(rad/s)^2}}$                          \\

        $\rev{c_\tau}$                                &
        $\rev{0.02\,\mathrm{m}}$                                                  \\

        $\rev{{}^{B}\!\boldsymbol{r}_{\mathrm{com}}}$ &
        $\rev{(5.750, -2.750, 1.716)\times 10^{-3}\,\mathrm{m}}$                  \\

        \bottomrule
    \end{tabular}
\end{table}
\subsection{\rev{Wrench Margins and Task-Based Screening}}
\rev{Let ${}^{B}\!\boldsymbol{w}_{\mathrm{eq}}=\operatorname{col}(\boldsymbol f_{\mathrm{eq}},\boldsymbol t_{\mathrm{eq}})\in\mathbb R^6$ denote the
aerodynamic wrench required at the equilibrium point.
The attainable force space $\mathbb F_P$ (AFS) and attainable torque space $\mathbb T_P$ (ATS) characterize
force and torque reserve while preserving equilibrium in the complementary channel:}
\begin{revblock}
    \begin{equation}
        \begin{aligned}
            \mathbb F_P(\boldsymbol \Lambda,\boldsymbol t_{\mathrm{eq}}) \! & =\! \left\{\!
            \boldsymbol f \!\in\!\mathbb R^3 \middle|
            \exists\boldsymbol u_f^c\!\in\!\mathbb U_P\!\!:\! \mathbf B_0\boldsymbol\Lambda\boldsymbol u_f^c \!=\! \operatorname{col} (\boldsymbol f,\boldsymbol t_{\mathrm{eq}})
            \!\right\}                                                             \\
            \mathbb T_P(\boldsymbol \Lambda,\boldsymbol f_{\mathrm{eq}}) \! & =\! \left\{\!
            \boldsymbol t\!\in\!\mathbb R^3 \middle|
            \exists\boldsymbol u_f^c\!\in\!\mathbb U_P\!\!:\! \mathbf B_0\boldsymbol\Lambda\boldsymbol u_f^c \!=\! \operatorname{col} (\boldsymbol f_{\mathrm{eq}},\boldsymbol t)
            \!\right\}
        \end{aligned}
        \label{eq:aws-force-torque-space-definition}
    \end{equation}
    where $P\in\{\mathrm{BTO},\mathrm{UTO},\mathrm{CCU}\}$ is the platform configuration, and $\mathbb U_P=\mathbb U_{P,1}\times\cdots\times\mathbb U_{P,6}\subset\mathbb R^{18}$ is the Cartesian product of the six independently constrained actuator-force sets:
    \begin{equation}
        \mathbb U_{P,i} = \left\{ {}^{A_i}\!\boldsymbol f_{\mathrm{ae}}^{P} \,\middle|\,
        \begin{array}{l}
            \alpha_i\in[-\alpha_{\max},\alpha_{\max}]   \\[-0.5mm]
            \lambda_i\in[\lambda_{\min},\lambda_{\max}] \\[-0.5mm]
            T_i\in[0,T_{\max}]
        \end{array}
        \right\}\subset\mathbb R^3
        \label{eq:aws-actuator-force-set}
    \end{equation}
\end{revblock}
\rev{where the bounds $\alpha_{\max}\!=\!\pi$, $\lambda_{\min}\!=\!-\pi$, $\lambda_{\max}=\pi/12$, and $T_{\max}\!=\!16\,\mathrm N$ are set by cable-routing, collision, and thrust constraints.}
\rev{The three configurations share $\mathbf B_0$ but differ in their actuator-force sets. 
The BTO follows \eqref{eq:btau-aerodynamic-force-torque}, while the UTO and conventional coplanar underactuated hexacopter (CCU) impose $\lambda_i=0$ and $\alpha_i=\lambda_i=0$, respectively:}
\begin{revblock}
    \begin{equation}
        {}^{A_i}\!\boldsymbol f_{\mathrm{ae}}^{\mathrm{UTO}} = \begin{bmatrix}
            0             \\[-1mm]
            \sin\alpha_i  \\[-1mm]
            -\cos\alpha_i \\
        \end{bmatrix} T_i , \qquad
        {}^{A_i}\!\boldsymbol f_{\mathrm{ae}}^{\mathrm{CCU}} = \begin{bmatrix}
            0  \\[-1mm]
            0  \\[-1mm]
            -1 \\
        \end{bmatrix} T_i
        \label{eq:aws-uto-ccu-force-models}
    \end{equation}

    Under bounds \eqref{eq:aws-actuator-force-set}, the equivalent BTO actuator-force set is
    \begin{equation}
        \mathbb U_{\mathrm{BTO},i} =
        \left\{
        \boldsymbol u_i\in\mathbb R^3
        \middle|
        \begin{array}{l}
            \|\boldsymbol u_i\|_2\leq T_{\max}, \\[-0.5mm]
            u_{i,x} +\sin\lambda_{\max}\|\boldsymbol u_i\|_2 \geq0
        \end{array}
        \right\}
        \label{eq:aws-bto-vector-force-set}
    \end{equation}
    where $\boldsymbol u_i \triangleq{}^{A_i}\!\boldsymbol f_{\mathrm{ae}}^{\mathrm{BTO}} = (u_{i,x},u_{i,y},u_{i,z})^\top$.

    For horizontal hover without disturbances, set
    ${}^{B}\!\boldsymbol w_{\mathrm{eq}}={}^{B}\!\boldsymbol w_h
    \triangleq\operatorname{col}(\boldsymbol f_h,\boldsymbol t_h)
    =-{}^{B}\!\boldsymbol w_{\mathrm{gra}}$.
    For each platform, we consider all $2^6=64$ endpoint configurations,
    in which each rotor is fully healthy or has suffered a total failure.
    For each configuration, offline sampling uses $N=10242$ unit directions $\boldsymbol d_j$ from a level-5 icosphere,
    maximizing $\boldsymbol d_j^\top\boldsymbol f$ over the AFS and $\boldsymbol d_j^\top\boldsymbol t$ over the ATS at this equilibrium.
    UTO and CCU use
    second-order cone programming (SOCP) and linear programming, respectively.
    BTO uses an SOCP relaxation, with mixed-integer quadratically constrained
    programming to obtain a physically feasible point.
    Upper bounds on directional support values provide outer polytopes for margin checks.

    Let $\boldsymbol f_{P,j}$ and $\boldsymbol t_{P,j}$ denote the solutions passing
    the physical actuator and equilibrium constraint checks, with
    $\mathbb J_F$ and $\mathbb J_T$ indexing the directions yielding such solutions.
    Their convex hulls provide inner approximations of the convexified sections:
    \begin{equation}
        \begin{aligned}
            \widehat{\mathbb F}_{P}\left(\boldsymbol\Lambda,\boldsymbol t_h\right)
             & \triangleq \operatorname{conv}
            \left\{
            \boldsymbol f_{P,j}
            \mid j\in\mathbb J_F
            \right\}
            \\
            \widehat{\mathbb T}_{P}\left(\boldsymbol\Lambda,\boldsymbol f_h\right)
             & \triangleq \operatorname{conv}
            \left\{
            \boldsymbol t_{P,j}
            \mid j\in\mathbb J_T
            \right\}
        \end{aligned}
        \label{eq:aws-support-polytopes}
    \end{equation}
    \begin{figure*}[!p]
        \centering
        \includegraphics[
            width=\textwidth,
            height=0.9\textheight,
            keepaspectratio
        ]{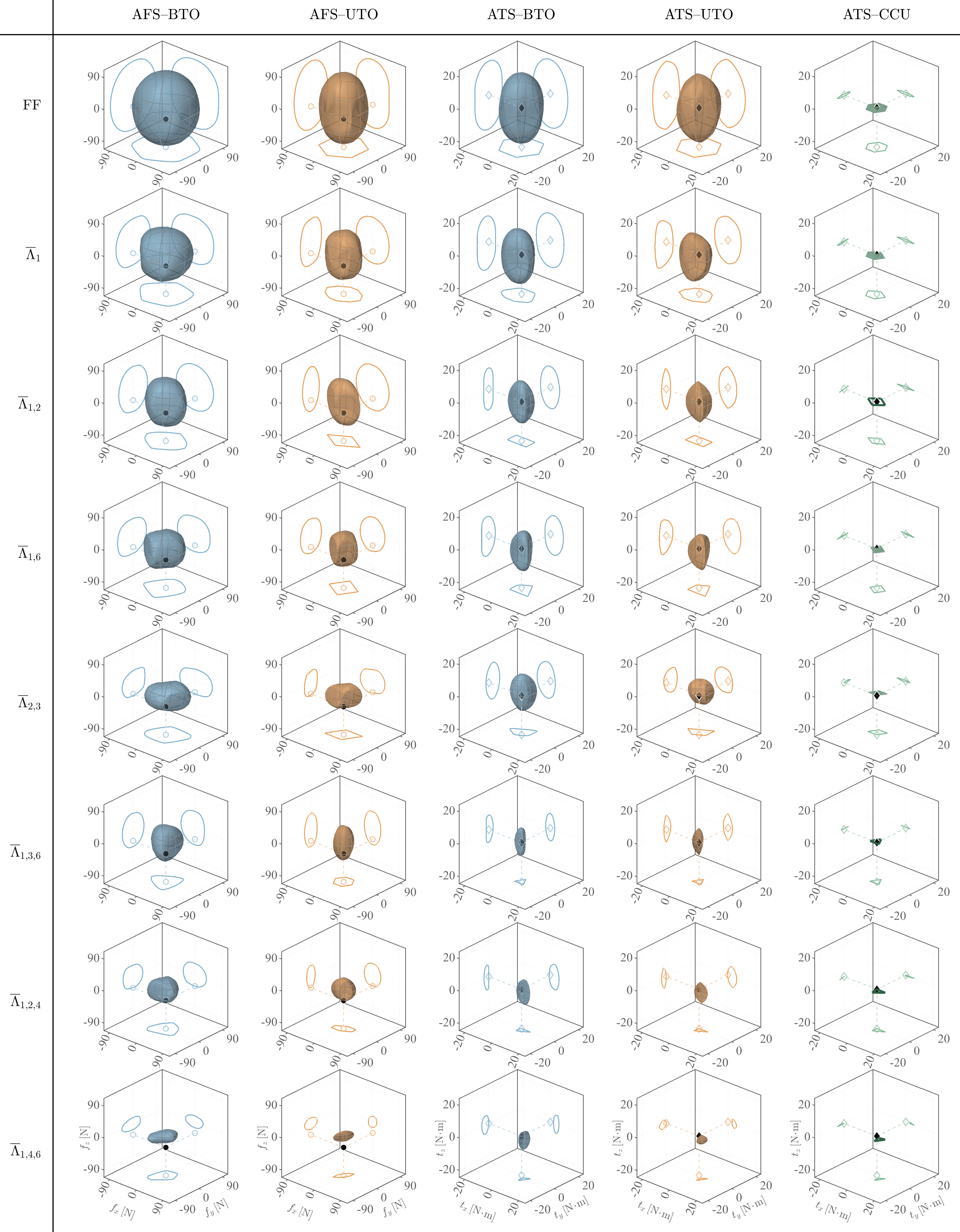}
        \caption{\rev{Sampled convex-hull representations of the equilibrium-constrained AFS of the BTO and UTO and ATS of the BTO, UTO, and CCU under the fault-free condition and seven selected total rotor failures. Filled circles and diamonds denote the hover-equilibrium force $\boldsymbol f_h$ and torque $\boldsymbol t_h$, respectively. Contours and open markers on the coordinate planes are orthogonal projections of the hulls and equilibrium points. The CCU AFS is omitted because it is empty or confined to the body $z$-axis, with no three-dimensional interior.}}
        \label{fig:aws-force-torque-spaces}
    \end{figure*}
    Let $\mathbb I_f=\{i\in\mathbb I\mid\eta_i=0\}$ index the failed rotors,
    with $\overline{\boldsymbol\Lambda}_{\mathbb I_f}$ denoting the corresponding
    total-failure effectiveness matrix; $\mathbf{FF}$ denotes the fault-free case.
    To quantify the remaining reserve,
    we define the force and torque margins following\textcolor{black}{~\cite{liu2023configuration}}:
    \begin{equation}
        \begin{aligned}
            r_{F,P}\left(\mathbb I_f\right) \! & =\!
            \max\! \left\{ r>0 \middle| \mathbb B_3(\boldsymbol f_h,r) \!\subseteq\! \widehat{\mathbb F}_{P}
            \left(\overline{\boldsymbol\Lambda}_{\mathbb I_f},\boldsymbol t_h\right)
            \!\right\}
            \\
            r_{T,P}\left(\mathbb I_f\right) \! & =\!
            \max\! \left\{ r>0 \middle| \mathbb B_3(\boldsymbol t_h,r) \!\subseteq\! \widehat{\mathbb T}_{P}
            \left(\overline{\boldsymbol\Lambda}_{\mathbb I_f},\boldsymbol f_h\right)
            \!\right\}
        \end{aligned}
        \label{eq:aws-force-torque-margins}
    \end{equation}
    where $\mathbb B_3(\boldsymbol c,r)\!=\!\left\{\boldsymbol x\in\mathbb R^3\mid\|\boldsymbol x-\boldsymbol c\|_2\leq r\right\}$.
    The sampled hulls make each positive radius a lower bound on the
    corresponding convexified margin.

    Continued task execution after rotor failures requires wrench reserves
    for command tracking and disturbance rejection.
    Let $\ddot p_{\mathrm{task}}>0$ and $\dot\omega_{\mathrm{task}}>0$
    denote prescribed translational and angular reserve thresholds,
    expressed as equivalent accelerations.
    The nominal mass and largest principal moment of inertia convert
    these thresholds into force and torque requirements, respectively.
    The retained fault set is
    \begin{equation}
        \mathbb A_P\triangleq\left\{
        \mathbb I_f\subseteq\mathbb I\;\middle|\;
        \begin{aligned}
            r_{F,P}(\mathbb I_f)&\geq m\ddot p_{\mathrm{task}},\\
            r_{T,P}(\mathbb I_f)&\geq\|\mathbf J_B\|_2\dot\omega_{\mathrm{task}}
        \end{aligned}
        \right\}
        \label{eq:aws-admissible-fault-set}
    \end{equation}
    where endpoints with undefined margins are excluded.
    For all configurations, we set $\ddot p_{\mathrm{task}}=2\,\mathrm{m/s^2}$
    and $\dot\omega_{\mathrm{task}}=1\,\mathrm{rad/s^2}$,
    corresponding to reserves of $6.4\,\mathrm N$
    (approximately $20\%$ of the vehicle weight)
    and $0.145\,\mathrm{N\cdot m}$ beyond hover.
    The CCU cannot provide the required lateral force because its thrust
    acts only along the body $z$ axis.
    
\end{revblock}
\begin{table}[t]
    \centering
    \caption{\rev{Wrench Margins and Task-Based Fault Screening}}
    \label{tab:aws-force-torque-margins}
    \footnotesize
    \setlength{\tabcolsep}{3pt}
    \renewcommand{\arraystretch}{0.84}
    \setlength{\aboverulesep}{1pt}
    \setlength{\belowrulesep}{1.5pt}
    \begin{revtable}
    \begin{tabular*}{0.92\columnwidth}{@{\extracolsep{\fill}}ccccccc@{}}
        \toprule
        \multirow{2}{*}{Fault} & \multicolumn{3}{c}{$r_F$ [N]} & \multicolumn{3}{c}{$r_T$ [$\mathrm{N\cdot m}$]} \\
        \cmidrule(lr){2-4}\cmidrule(lr){5-7}
        & BTO & UTO & CCU & BTO & UTO & CCU \\
        \midrule
        $\mathbf{FF}$ & \textbf{64.225} & 55.347 & \textemdash & \textbf{11.139} & 10.995 & 0.628 \\
        \midrule
        $\overline{\boldsymbol\Lambda}_{1}$ & \textbf{32.559} & 31.395 & \textemdash & \textbf{5.622} & 5.441 & \textemdash \\
        $\overline{\boldsymbol\Lambda}_{2}$ & \textbf{34.122} & 32.936 & \textemdash & \textbf{5.983} & 5.801 & 0.031$^{\dagger}$ \\
        $\overline{\boldsymbol\Lambda}_{3}$ & \textbf{33.367} & 32.052 & \textemdash & \textbf{5.818} & 5.640 & 0.003$^{\dagger}$ \\
        $\overline{\boldsymbol\Lambda}_{4}$ & \textbf{33.235} & 31.914 & \textemdash & \textbf{5.789} & 5.609 & \textemdash \\
        $\overline{\boldsymbol\Lambda}_{5}$ & \textbf{32.618} & 31.429 & \textemdash & \textbf{5.638} & 5.460 & \textemdash \\
        $\overline{\boldsymbol\Lambda}_{6}$ & \textbf{34.037} & 32.840 & \textemdash & \textbf{5.969} & 5.789 & 0.028$^{\dagger}$ \\
        \midrule
        $\overline{\boldsymbol\Lambda}_{1,2}$ & \textbf{31.508} & 25.656 & \textemdash & \textbf{3.678} & 3.586 & \textemdash \\
        $\overline{\boldsymbol\Lambda}_{1,3}$ & \textbf{16.798} & 15.415 & \textemdash & \textbf{3.593} & 3.133 & \textemdash \\
        $\overline{\boldsymbol\Lambda}_{1,6}$ & \textbf{17.415} & 15.852 & \textemdash & \textbf{3.570} & 3.221 & \textemdash \\
        $\overline{\boldsymbol\Lambda}_{2,3}$ & \textbf{4.513}$^{\dagger}$ & 1.627$^{\dagger}$ & \textemdash & \textbf{0.902} & 0.329 & 0.003$^{\dagger}$ \\
        $\overline{\boldsymbol\Lambda}_{2,4}$ & \textbf{18.218} & 16.826 & \textemdash & \textbf{3.880} & 3.420 & \textemdash \\
        $\overline{\boldsymbol\Lambda}_{2,5}$ & \textbf{17.548} & 15.985 & \textemdash & \textbf{3.598} & 3.248 & \textemdash \\
        $\overline{\boldsymbol\Lambda}_{2,6}$ & \textbf{3.846}$^{\dagger}$ & 2.053$^{\dagger}$ & \textemdash & \textbf{0.777} & 0.416 & 0.028$^{\dagger}$ \\
        $\overline{\boldsymbol\Lambda}_{3,4}$ & \textbf{32.167} & 26.277 & \textemdash & \textbf{3.844} & 3.754 & \textemdash \\
        $\overline{\boldsymbol\Lambda}_{3,6}$ & \textbf{18.299} & 16.954 & \textemdash & \textbf{3.883} & 3.446 & 0.003$^{\dagger}$ \\
        $\overline{\boldsymbol\Lambda}_{4,5}$ & \textbf{16.736} & 15.413 & \textemdash & \textbf{3.569} & 3.132 & \textemdash \\
        $\overline{\boldsymbol\Lambda}_{4,6}$ & \textbf{4.379}$^{\dagger}$ & 1.493$^{\dagger}$ & \textemdash & \textbf{0.875} & 0.303 & \textemdash \\
        $\overline{\boldsymbol\Lambda}_{5,6}$ & \textbf{31.651} & 25.720 & \textemdash & \textbf{3.694} & 3.604 & \textemdash \\
        \midrule
        $\overline{\boldsymbol\Lambda}_{1,2,4}$ & \textemdash & \textemdash & \textemdash & \textemdash & \textemdash & \textemdash \\
        $\overline{\boldsymbol\Lambda}_{1,3,6}$ & \textbf{14.946} & 14.543 & \textemdash & \textbf{1.745} & 1.735 & \textemdash \\
        $\overline{\boldsymbol\Lambda}_{1,4,6}$ & \textemdash & \textemdash & \textemdash & \textemdash & \textemdash & \textemdash \\
        $\overline{\boldsymbol\Lambda}_{2,4,5}$ & \textbf{14.813} & 14.439 & \textemdash & \textbf{1.729} & 1.720 & \textemdash \\
        \bottomrule
    \end{tabular*}
    \par\vspace{2pt}
    \begin{minipage}{0.92\columnwidth}
        \footnotesize
        Entries marked with $\dagger$ are positive margins below the corresponding reserve thresholds.
        \textemdash\ denotes that no positive margin was
        established.
    \end{minipage}
    \end{revtable}
\end{table}
\begin{revblock}
    BTO and UTO retain the same 18 endpoints, including $\mathbf{FF}$,
    listed alongside selected excluded cases in
    Table~\ref{tab:aws-force-torque-margins}.
    The adjacent pairs $\{2,3\}$, $\{2,6\}$, and $\{4,6\}$ retain
    positive margins, but even the upper bounds on their force margins
    remain below the threshold.
    Across all 64 endpoints, screening with either lower or upper margin
    bounds retains this set for
    $\ddot p_{\mathrm{task}}\in[1.5,4]\,\mathrm{m/s^2}$ and
    $\dot\omega_{\mathrm{task}}\in[0.5,5]\,\mathrm{rad/s^2}$.

    For the illustrated double failures, 
    Fig.~\ref{fig:aws-force-torque-spaces} and Table~\ref{tab:aws-force-torque-margins} show 
    decreasing BTO force reserve from the diametrically opposed pair 
    $\overline{\boldsymbol\Lambda}_{1,2}$ ($180^\circ$), to $\overline{\boldsymbol\Lambda}_{1,6}$ ($120^\circ$),
    and then to the adjacent pair $\overline{\boldsymbol\Lambda}_{2,3}$ ($60^\circ$).
    The spatial distribution of the surviving actuators affects their ability to generate additional force 
    while satisfying the equilibrium torque constraint in \eqref{eq:aws-force-torque-space-definition}.
    Among the illustrated triple failures,
    $\overline{\boldsymbol\Lambda}_{1,3,6}$ involves three equally spaced failed rotors and passes screening,
    whereas no positive margins in three dimensions are established for $\overline{\boldsymbol\Lambda}_{1,2,4}$ or $\overline{\boldsymbol\Lambda}_{1,4,6}$.

    The computed BTO force and torque margins exceed those of UTO by $22.81\%$ and $2.55\%$ under
    $\overline{\boldsymbol\Lambda}_{1,2}$, compared with $9.86\%$ and $10.84\%$ under $\overline{\boldsymbol\Lambda}_{1,6}$.
    For all 18 retained endpoints, positive UTO section margins establish
    a statically attainable wrench neighborhood about $\boldsymbol w_h$
    for both configurations, since the UTO AWS is convex and contained in
    the BTO AWS.

    Extending the endpoints retained by task screening to partial
    effectiveness losses gives
    \begin{equation}
        \mathbb L_P \triangleq \bigcup_{\mathbb I_f\in\mathbb A_P}
        \left\{\boldsymbol\Lambda\middle|
        \overline{\boldsymbol\Lambda}_{\mathbb I_f}\preceq\boldsymbol\Lambda\preceq\mathbf I_{18}\right\},
        \label{eq:aws-admissible-effectiveness-set}
    \end{equation}

    Positive force and torque margins imply that the two sections jointly
    span all six wrench directions, so
    $\operatorname{rank}(\mathbf B_0\overline{\boldsymbol\Lambda}_{\mathbb I_f})=6$
    for every $\mathbb I_f\in\mathbb A_P$.
    Any $\boldsymbol\Lambda\in\mathbb L_P$ preserves these active columns, so $\operatorname{rank}(\mathbf B_0\boldsymbol\Lambda)=6$.
\end{revblock}
\subsection{\rev{Fault Severity Assessment}}
\begin{revblock}
    For BTO, adopt the hover allocation before failure
    $\boldsymbol u_h=\mathbf B_W^\dagger\boldsymbol w_h$ from
    \eqref{eq:cl-weighted-pseudoinverse-allocation}, using
    $\mathbf W=\mathbf I_6\otimes\operatorname{diag}(0.9007,0.6947,1)\in \mathbb R^{18\times 18}$.
    The instantaneous wrench loss for endpoint $\mathbb I_f$ is
    \begin{equation}
        {}^{B}\!\boldsymbol w_{\mathrm{L}} \left(\mathbb I_f\right) = \mathbf B_0\left(\mathbf I_{18}-\overline{\boldsymbol\Lambda}_{\mathbb I_f}\right)\boldsymbol u_h
        \label{eq:aws-instantaneous-wrench-loss}
    \end{equation}
    where ${}^{B}\!\boldsymbol w_{\mathrm{L}}=\operatorname{col}(\boldsymbol f_{\mathrm{L}},\boldsymbol t_{\mathrm{L}})$ is the instantaneous wrench loss.
\end{revblock}

\begin{figure}[!htb]
    \centering
    \includegraphics[width=75mm]{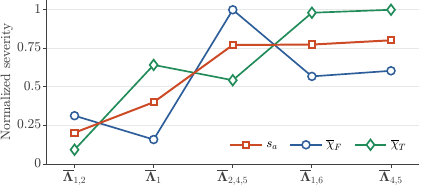}
    \caption{\rev{Normalized ratios and values of $s_a$ for five representative BTO fault cases retained by task screening, sorted by $s_a$. Normalization uses all 18 endpoints in $\mathbb A_{\mathrm{BTO}}$, including $\mathbf{FF}$.}}
    \label{fig:aws-fault-severity}
\end{figure}

\begin{revblock}
    For the BTO endpoint configurations $\mathbb I_f\in\mathbb A_{\mathrm{BTO}}$, 
    define the ratios of the instantaneous force and torque losses
    to the corresponding margins as follows:
    \begin{equation}
        \chi_F(\mathbb I_f) = \frac{\left\|\boldsymbol f_{\mathrm{L}}(\mathbb I_f)\right\|_2}
        {r_{F,\mathrm{BTO}}(\mathbb I_f)},\;
        \chi_T(\mathbb I_f) = \frac{\left\|\boldsymbol t_{\mathrm{L}}(\mathbb I_f)\right\|_2}
        {r_{T,\mathrm{BTO}}(\mathbb I_f)}
        \label{eq:aws-loss-margin-ratios}
    \end{equation}
    Normalize and combine the ratios as
    \begin{equation}
        \left\{
        \begin{aligned}
             & \overline{\chi}_{F}(\mathbb I_f) \!=\! \frac{ \chi_{F}(\mathbb I_f)}{\displaystyle\max_{\mathbb J\in\mathbb A_{\mathrm{BTO}}}\chi_{F}(\mathbb J)},
            \overline{\chi}_{T}(\mathbb I_f) \!=\! \frac{\chi_{T}(\mathbb I_f)}{\displaystyle\max_{\mathbb J\in\mathbb A_{\mathrm{BTO}}}\chi_{T}(\mathbb J)}
            \\
             & s_a(\mathbb I_f) = \frac{\overline{\chi}_{F}(\mathbb I_f) + \overline{\chi}_{T}(\mathbb I_f) }{2},\qquad \mathbb I_f \in \mathbb A_{\mathrm{BTO}}
        \end{aligned}
        \right.
        \label{eq:aws-normalized-fault-severity}
    \end{equation}
    The fault severity index $s_a\in[0,1]$ averages the normalized loss-to-margin ratios under $\boldsymbol u_h$.
    It provides an offline comparison of initial fault effects.
    Table~\ref{tab:aws-loss-margin-ratios} reports the raw ratios, and Fig.~\ref{fig:aws-fault-severity} shows their normalized values and $s_a$.

\end{revblock}
\begin{table}[!htb]
    \centering
    \caption{\rev{Loss-to-Margin Ratios for Representative BTO Fault Cases}}
    \label{tab:aws-loss-margin-ratios}
    \footnotesize
    \renewcommand{\arraystretch}{1.00}
    \begin{revtable}
        \begin{tabular}{cccccc}
            \toprule
            \textrm{Fault}
                     & $\overline{\boldsymbol\Lambda}_{1}$   & $\overline{\boldsymbol\Lambda}_{1,2}$   & $\overline{\boldsymbol\Lambda}_{1,6}$
                     & $\overline{\boldsymbol\Lambda}_{4,5}$ & $\overline{\boldsymbol\Lambda}_{2,4,5}$                                                             \\
            \midrule

            $\chi_F$ & $0.169$                               & $0.332$                                 & $0.602$                               & $0.641$ & $1.060$ \\
            $\chi_T$ & $0.225$                               & $0.033$                                 & $0.344$                               & $0.351$ & $0.191$ \\
            \bottomrule
        \end{tabular}
    \end{revtable}
\end{table}
\begin{revblock}
Under the prescribed allocation before failure, the moment contributions of rotors~1 and~2 largely cancel about $O_B$. The net moment loss under $\overline{\boldsymbol\Lambda}_{1,2}$ is therefore $0.12\,\mathrm{N\cdot m}$, versus $1.27\,\mathrm{N\cdot m}$ under $\overline{\boldsymbol\Lambda}_{1}$, despite a smaller torque margin. After normalization, the decrease in $\bar{\chi}_T$ outweighs the increase in $\bar{\chi}_F$, yielding a lower $s_a$ for $\overline{\boldsymbol\Lambda}_{1,2}$.

The double failures $\overline{\boldsymbol\Lambda}_{1,2}$ and $\overline{\boldsymbol\Lambda}_{1,6}$ cause similar force losses ($10.46$ and $10.49\,\mathrm N$), but the latter has a smaller force margin ($17.41$ versus $31.51\,\mathrm N$). Their torque margins are similar, while the net moment loss increases from $0.12$ to $1.23\,\mathrm{N\cdot m}$. Thus, the higher $\chi_F$ mainly reflects reduced reserve, and the higher $\chi_T$ mainly reflects greater initial moment loss. Although $\overline{\boldsymbol\Lambda}_{2,4,5}$ has the largest $\chi_F$ in the screened set, its smaller $\chi_T$ gives a lower $s_a$ than $\overline{\boldsymbol\Lambda}_{4,5}$. Fault geometry affects both losses and reserves, so the number of failed rotors alone does not determine severity.
\end{revblock}

\section{CL-PFTC Design}
\begin{figure}[!h]
    \centering
    \includegraphics[width=88mm]{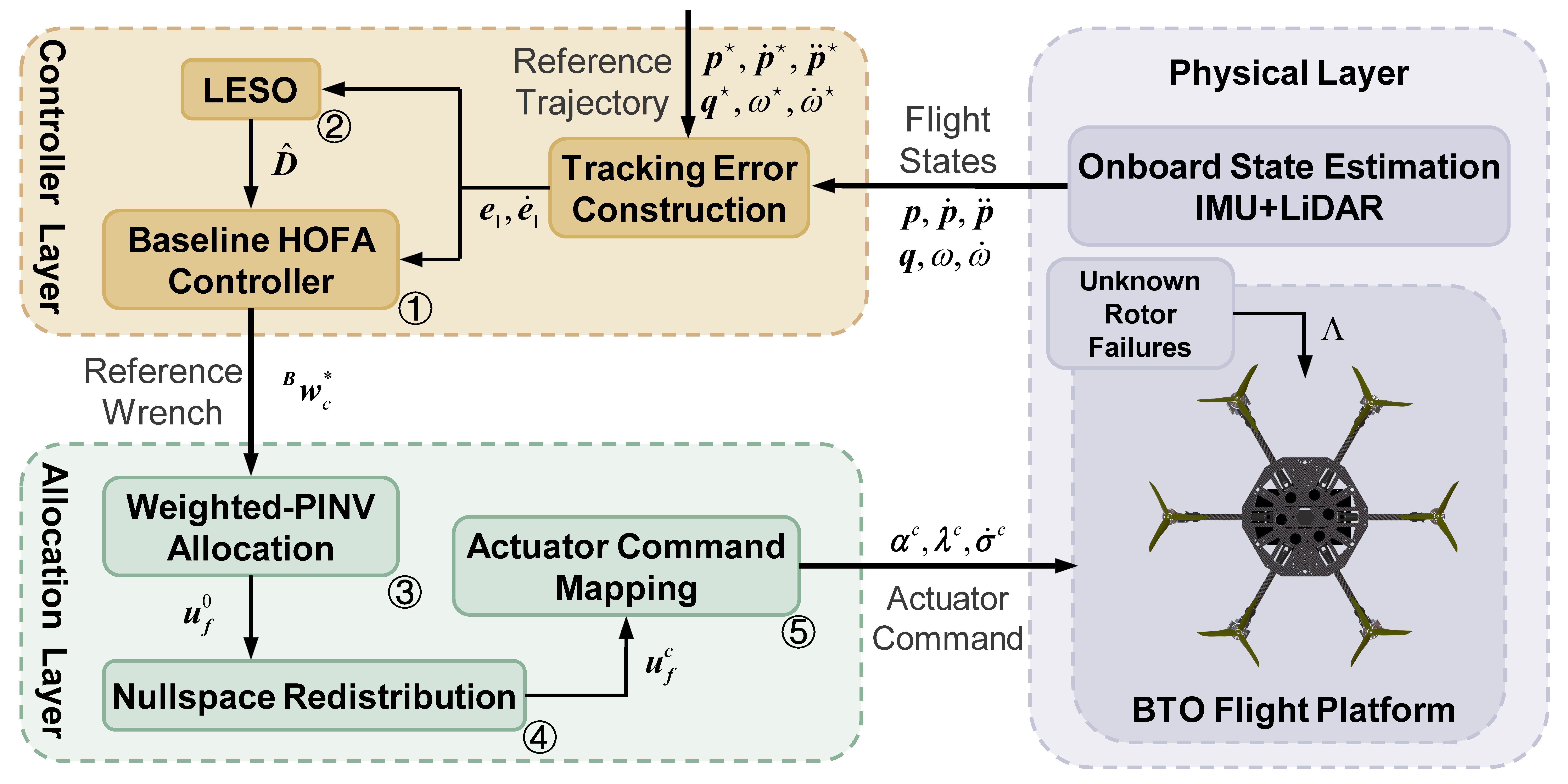}
    \caption{\rev{CL-PFTC architecture comprising
        \ding{172} baseline HOFA control~\eqref{eq:cl-hofa-control-law},
        \ding{173} LESO compensation~\eqref{eq:cl-leso},
        \ding{174} weighted pseudoinverse (PINV) allocation~\eqref{eq:cl-weighted-pseudoinverse-allocation},
        \ding{175} maximum tilt-angle constraint (MATC) nullspace redistribution~\eqref{eq:cl-matc-force-command}, and
        \ding{176} actuator-command mapping~\eqref{eq:cl-actuator-command-conversion}.}}
    \label{fig:cl-pftc-framework}

    \vspace{0.5\baselineskip}
    \begingroup
    \makeatletter
    \def\@captype{table}
    \makeatother
    \caption{\rev{Main Symbols and Notation for CL-PFTC}}
    \label{tab:cl-pftc-notation}
    \footnotesize
    \setlength{\tabcolsep}{2pt}
    \renewcommand{\arraystretch}{1.00}
    \begin{revtable}
        \begin{tabular}{@{}
            >{\centering\arraybackslash}m{0.12\columnwidth}
            >{\raggedright\arraybackslash}m{0.82\columnwidth}
            @{}}
            \toprule
            \textbf{Symbol} & \textbf{Definition}                                                   \\
            \midrule
            $\boldsymbol e_p$
                            & Position tracking error expressed in $\mathcal F_E$;
            $\boldsymbol e_p \in \mathbb R^{3}$ [m].                                                \\

            $\boldsymbol q_e$
                            & Unit quaternion representing attitude tracking error;
            $\boldsymbol q_e \in \mathbb S^3$.                                                      \\

            $\boldsymbol e_1$
                            & Position and attitude tracking error vector;
            $\boldsymbol e_1 \in \mathbb R^6$.                                                      \\

            $\boldsymbol e$
                            & Tracking error state vector $\boldsymbol{e}=\operatorname{col}(\boldsymbol{e}_1,\dot{\boldsymbol{e}}_1)$ \;
            $\boldsymbol e\in \mathbb R^{12}$.                                                      \\

            $\boldsymbol D$
                            & Controller-layer lumped disturbance;
            $\boldsymbol D\in\mathbb R^6$.                                                          \\

            ${}^{B}\!\boldsymbol w_c^\star$
                             & Reference aerodynamic control wrench;
            ${}^{B}\!\boldsymbol w_c^\star\in\mathbb R^6$. \\

            $\mathbf D_w$
                            & Diagonal matrix of wrench limits;
            $\mathbf D_w\in\mathbb R^{6\times6}$.                                                  \\

            $\boldsymbol s_w$
                            & Command clipping error;
            $\boldsymbol s_w\in\mathbb R^6$.                                                       \\

            $(\cdot)^{\star}$
                            & Superscript denoting a reference quantity.                            \\

            $(\cdot)^{\mathrm{n}}$
                            & Superscript denoting a nominal model quantity.                        \\
            \bottomrule
        \end{tabular}
    \end{revtable}
    \endgroup
\end{figure}

\begin{revblock}
    CL-PFTC serves as the controller-layer reference for abrupt total rotor failures with an unknown index set $\mathbb I_f\in\mathbb A_P$.
    As shown in Fig.~\ref{fig:cl-pftc-framework}, its fixed nominal allocator and wrench-level HOFA--LESO loop provide 6-DOF recovery under the assumptions below.
    Table~\ref{tab:cl-pftc-notation} summarizes the notation.
\end{revblock}
\subsection{Fault-Tolerant Controller Design}\label{subsec:cl-controller-design}
\subsubsection{Baseline HOFA Controller Design}\label{subsubsec:cl-baseline-hofa-controller}
\begin{revblock}
    Given $\boldsymbol p^\star\in\mathbb R^3$ and $\boldsymbol q^\star\in\mathbb S^3$, define
    $\boldsymbol e_p=\boldsymbol p^\star-\boldsymbol p$,
    $\boldsymbol q_e=\boldsymbol q^{-1}\otimes\boldsymbol q^\star$, and
    $\boldsymbol e_1\triangleq\operatorname{col}(\boldsymbol e_p,\boldsymbol q_{ev})\in\mathbb R^6$,
    where $q_{e0}$ and $\boldsymbol q_{ev}$ are the scalar and vector parts of $\boldsymbol q_e$.
    Substitution into \eqref{eq:bto-newton-euler-dynamics} gives \eqref{eq:cl-tracking-error-dynamics}.
\end{revblock}
\begin{equation}
    \left\{
    \begin{aligned}
        \ddot{\boldsymbol{e}}_1 & = \rev{\boldsymbol a}
        \!+\!\rev{\boldsymbol \Phi \rev{\mathbf{M}_B^{-1}}}\left(\rev{{}^{B}\!\boldsymbol w_{\mathrm{gra}} \!+\! {}^{B}\!\boldsymbol w_{\mathrm{rot}} \!+\! {}^{B}\!\boldsymbol w_{\mathrm{ae}} \!+\! {}^{B}\!\boldsymbol w_d}\right)
        \\
        \boldsymbol\Omega       & = \tfrac{1}{2}\left(\dot{q}_{e0}\rev{\mathbf I_3}\!-\!\left[\dot{\boldsymbol{q}}_{ev}\right]_\times\right)\boldsymbol{\omega}_e\!+\!\tfrac{1}{2}\left(q_{e0}\rev{\mathbf I_3}\!-\!\left[\boldsymbol{q}_{ev}\right]_\times\right)\rev{{}^{B}\!\dot{\boldsymbol\omega}^{\star}}
        \\
        \boldsymbol \Phi        & =
        \begin{bmatrix}
            -\mathbf{I}_3 & \mathbf{0}_3                                                                        \\
            \mathbf{0}_3  & -\frac{1}{2}\left(q_{e0}\mathbf{I}_3-\left[\boldsymbol{q}_{ev}\right]_\times\right)
        \end{bmatrix}\rev{\in \mathbb R^{6\times 6}}
    \end{aligned}
    \right.
    \label{eq:cl-tracking-error-dynamics}
\end{equation}
\begin{revblock}
    where $\boldsymbol a\triangleq\operatorname{col}(\ddot{\boldsymbol p}^{\star},\boldsymbol\Omega)\in\mathbb R^6$ combines the reference translational acceleration with the nonlinear terms $\boldsymbol\Omega\in\mathbb R^3$ from the attitude error kinematics.
    Here $\boldsymbol\omega_e={}^B\!\boldsymbol\omega^\star-\boldsymbol\omega$,
    $\dot q_{e0}=-\tfrac12\boldsymbol\omega_e^\top\boldsymbol q_{ev}$, and
    $\dot{\boldsymbol q}_{ev}=\tfrac12(q_{e0}\mathbf I_3-[\boldsymbol q_{ev}]_\times)\boldsymbol\omega_e$.
\end{revblock}
\begin{revblock}
    Decompose $\mathbf M_B$ as
    \begin{equation}
        \mathbf M_B  = \mathbf M_B^{\mathrm n} + \Delta\mathbf M_B
        \label{eq:cl-inertia-model-uncertainty}
    \end{equation}
    where $\mathbf M_B^{\mathrm n}\in\mathbb R^{6\times6}$ is its nominal value and $\Delta\mathbf M_B\in\mathbb R^{6\times6}$ captures inertial-parameter uncertainty.
    The corresponding nominal gravity and rotational wrenches are denoted by ${}^{B}\!\boldsymbol w_{\mathrm{gra}}^{\mathrm n}$ and ${}^{B}\!\boldsymbol w_{\mathrm{rot}}^{\mathrm n}$, respectively.
\end{revblock}
\begin{revblock}
    With ${}^{B}\!\boldsymbol w_c^\star$ denoting the reference aerodynamic wrench, \eqref{eq:cl-tracking-error-dynamics} becomes
\end{revblock}
\begin{equation}
    \begin{aligned}
        \ddot{\boldsymbol e}_1 \!=\! \rev{\boldsymbol a} \!+\!
        \boldsymbol\Phi \rev{\left(\mathbf M_B^{\mathrm n}\right)^{-1}}
        \left(\rev{
            {}^{B}\!\boldsymbol w_c^{\star} \!+\! {}^{B}\!\boldsymbol w_{\mathrm{gra}}^{\mathrm n} \!+\! {}^{B}\!\boldsymbol w_{\mathrm{rot}}^{\mathrm n}
        }\right) \!+\! \rev{\boldsymbol D}
    \end{aligned}
    \label{eq:cl-controller-error-dynamics}
\end{equation}
\begin{revblock}
    The lumped disturbance $\boldsymbol D\in\mathbb R^6$ is decomposed as
\end{revblock}
\begin{revblock}
    \begin{equation}
        \left\{
        \begin{aligned}
            \boldsymbol D   & = \boldsymbol D_m + \boldsymbol D_f + \boldsymbol D_d                                                                                                                                      \\
            \boldsymbol D_m & = \boldsymbol\Phi\mathbf M_B^{-1} \left({}^{B}\!\boldsymbol w_{\mathrm{gra}} + {}^{B}\!\boldsymbol w_{\mathrm{rot}} + {}^{B}\!\boldsymbol w_c^{\star}\right)                                                 \\
                            & \quad-\boldsymbol\Phi\left(\mathbf M_B^{\mathrm n}\right)^{-1}\left({}^{B}\!\boldsymbol w_{\mathrm{gra}}^{\mathrm n}+{}^{B}\!\boldsymbol w_{\mathrm{rot}}^{\mathrm n}+{}^{B}\!\boldsymbol w_c^{\star}\right) \\
            \boldsymbol D_f & = \boldsymbol\Phi\mathbf M_B^{-1}\left(\mathbf B_0\boldsymbol\Lambda\boldsymbol u_f^c-{}^{B}\!\boldsymbol w_c^{\star}\right)                                                               \\
            \boldsymbol D_d & = \boldsymbol\Phi\mathbf M_B^{-1}{}^{B}\!\boldsymbol w_d
        \end{aligned}
        \right.
        \label{eq:cl-disturbance-decomposition}
    \end{equation}
    where $\boldsymbol D_m$ collects inertial-parameter uncertainty and the associated model errors.
    $\boldsymbol D_f$ represents the aerodynamic wrench mismatch caused by unknown rotor failures and errors in control allocation (CA).
    $\boldsymbol D_d$ collects the effects of actuator dynamics, unmodeled dynamics, wind, and external disturbances.

    \begin{assumption}[Attitude-error domain]
        \label{ass:attitude-domain}
        The design and analysis are conducted on a fixed quaternion branch within
        $\mathbb Q_{\varepsilon}
            =\{\boldsymbol q_e\in\mathbb S^3:
            |q_{e0}|\geq\varepsilon_q\}$,
        where $0<\varepsilon_q<1$.
    \end{assumption}

    Under Assumption~\ref{ass:attitude-domain}, $\det(\boldsymbol\Phi)=q_{e0}/8\neq0$, so
    \eqref{eq:cl-controller-error-dynamics} has a second-order HOFA form.
    Following\textcolor{black}{ \cite{duanHighorderFullyActuated2021}}, use the control law
\end{revblock}
\begin{equation}
    \left\{
    \begin{aligned}
         & \rev{{}^{B}\!\boldsymbol w_c^{\star}=\operatorname{Sat}_{\mathbf D_w}({}^{B}\!\boldsymbol w_c^{\mathrm{raw}})} \\
         & \rev{{}^{B}\!\boldsymbol w_c^{\mathrm{raw}}} =-\rev{\mathbf M_B^{\mathrm{n}}\boldsymbol \Phi^{-1}}\left(\mathbf{A}_1\dot{\boldsymbol{e}}_1 + \mathbf{A}_0\boldsymbol{e}_1 + \rev{\boldsymbol\nu}\right) \\
         & \rev{\boldsymbol\nu} \triangleq \rev{\boldsymbol\Phi \rev{\left(\mathbf M_B^{\mathrm n}\right)^{-1}}}
        \left(\rev{{}^{B}\!\boldsymbol w_{\mathrm{gra}}^{\mathrm n} \!+\! {}^{B}\!\boldsymbol w_{\mathrm{rot}}^{\mathrm n}}\right)
        +\hat{\boldsymbol{D}} + \rev{\boldsymbol a}
    \end{aligned}
    \right.
    \label{eq:cl-hofa-control-law}
\end{equation}
\begin{revblock}
    where $\mathbf A_0,\mathbf A_1\in\mathbb R^{6\times6}$ are selected such that
    $s^2\mathbf I_6+\mathbf A_1s+\mathbf A_0$ is Hurwitz,
    $\boldsymbol\nu\in\mathbb R^6$ is the auxiliary term, and
    $\hat{\boldsymbol D}\in\mathbb R^6$ is supplied by the LESO below.
    Here $\mathbf D_w=\operatorname{diag}(\boldsymbol f_{\max},\boldsymbol t_{\max})\succ0$ specifies the wrench limits;
    $\operatorname{Sat}_{\mathbf D_w}$ clips component $j$ to $\pm[\mathbf D_w]_{jj}$.
    The command clipping error is $\boldsymbol s_w\triangleq{}^B\!\boldsymbol w_c^{\mathrm{raw}}-{}^B\!\boldsymbol w_c^\star$.
\subsubsection{LESO Design}
    Let $\boldsymbol z_1\triangleq\boldsymbol e_1$,
    $\boldsymbol z_2\triangleq\dot{\boldsymbol e}_1$,
    $\boldsymbol z_3\triangleq\boldsymbol D$, and
    $\boldsymbol h_D\triangleq\dot{\boldsymbol D}$.
    The LESO is constructed in the standard form\textcolor{black}{~\cite{yang2009capabilities}}:
\end{revblock}
\begin{equation}
    \left\{
    \begin{aligned}
         & \dot{\hat{\boldsymbol{z}}}_1=\hat{\boldsymbol{z}}_2+\mathbf{L}_1\left(\boldsymbol{z}_1-\hat{\boldsymbol{z}}_1\right)
        \\
         & \begin{aligned}
               \dot{\hat{\boldsymbol{z}}}_2= &
               \hat{\boldsymbol{z}}_3+\mathbf{L}_2\left(\boldsymbol{z}_1-\hat{\boldsymbol{z}}_1\right)
               \\
                                             & +\rev{\boldsymbol a} \!+\! \boldsymbol\Phi \rev{\left(\mathbf M_B^{\mathrm n}\right)^{-1}}
               \left(\rev{
                   {}^{B}\!\boldsymbol w_c^{\mathrm{raw}} \!+\! {}^{B}\!\boldsymbol w_{\mathrm{gra}}^{\mathrm n} \!+\! {}^{B}\!\boldsymbol w_{\mathrm{rot}}^{\mathrm n}
               }\right)
           \end{aligned}
        \\
         & \dot{\hat{\boldsymbol{z}}}_3= \mathbf{L}_3\left(\boldsymbol{z}_1-\hat{\boldsymbol{z}}_1\right)
    \end{aligned}
    \right.
    \label{eq:cl-leso}
\end{equation}
\begin{revblock}
    where $\hat{\boldsymbol D}\triangleq\hat{\boldsymbol z}_3$ and, for $\boldsymbol s_w=\mathbf0$,
    $\hat{\boldsymbol z}_j$ estimates $\boldsymbol z_j$, $j=1,2,3$.
    The positive-definite diagonal gains $\mathbf L_1,\mathbf L_2,\mathbf L_3\in\mathbb R^{6\times6}$ are chosen such that
    $s^3\mathbf I_6+\mathbf L_1s^2+\mathbf L_2s+\mathbf L_3$ is Hurwitz.
\end{revblock}

\subsection{Control Allocation}
\subsubsection{Weighted Pseudoinverse Allocation}
\begin{revblock}
    Because the fault-dependent effectiveness matrix $\boldsymbol\Lambda$ is unavailable, CL-PFTC retains $\mathbf B_0$ and uses the weighted-PINV allocation
\end{revblock}
\begin{equation}
    \rev{\boldsymbol{u}_{f}^0}=
    \underbrace{
        \mathbf{W}^{-1}\mathbf{B}_{c}^{\top}\left(\mathbf{B}_c\mathbf{B}_{c}^{\top}\right)^{-1}}_{\rev{\mathbf{B}_{W}^{\dagger}\in\mathbb R^{18\times6}}}\rev{{}^{B}\!\boldsymbol w_c^\star}
    ,\mathbf{B}_c=\rev{\mathbf{B}_{0}}\mathbf{W}^{-1}
    \label{eq:cl-weighted-pseudoinverse-allocation}
\end{equation}
\begin{revblock}
    where $\boldsymbol u_f^0\in\mathbb R^{18}$ is the virtual actuator force command before constraint correction and
    $\mathbf W=\mathbf W^\top\succ0$ weights actuator effort.
    For full-row-rank $\mathbf B_c$, \eqref{eq:cl-weighted-pseudoinverse-allocation} minimizes
    $\|\mathbf W\boldsymbol u_f\|_2$ subject to $\mathbf B_0\boldsymbol u_f={}^B\!\boldsymbol w_c^\star$.
    The precomputed right inverse satisfies $\mathbf B_0\mathbf B_W^\dagger=\mathbf I_6$.
\end{revblock}

\subsubsection{Actuator Command Mapping}
By solving \eqref{eq:btau-aerodynamic-force-torque}, the initial reference tilt angles and rotor speeds are obtained as
\begin{equation}
    \left\{
    \begin{aligned}
         & \rev{\alpha_i^{0}} = \mathrm{atan2}\left(\rev{{}^{A_i}\! f_{\mathrm{ae},y}^0},-\rev{{}^{A_i}\!{f}_{\mathrm{ae},z}^0}\right)
        \\
         & \rev{\lambda_i^{0}} = \mathrm{arcsin}\left(\tfrac{-\rev{{}^{A_i}\! f_{\mathrm{ae},x}^0}}{\left \|\rev{{}^{A_i}\!\boldsymbol f_{\mathrm{ae}}^0}\right \|_2}\right)
        ,\;\rev{\dot{\sigma}_i^{0}} = \sqrt{\tfrac{\left \|\rev{{}^{A_i}\!\boldsymbol f_{\mathrm{ae}}^0}\right \|_2}{c_f}}
    \end{aligned}
    \right.
    \label{eq:cl-actuator-command-conversion}
\end{equation}
\rev{where ${}^{A_i}\!\boldsymbol f_{\mathrm{ae}}^{0}=\operatorname{col}({}^{A_i}\! f_{\mathrm{ae},x}^0,{}^{A_i}\! f_{\mathrm{ae},y}^0,{}^{A_i}\! f_{\mathrm{ae},z}^0)$}.

\begin{revblock}
    The nominal mapping in \eqref{eq:cl-weighted-pseudoinverse-allocation} and
    \eqref{eq:cl-actuator-command-conversion} does not enforce the MATC.
    For steady hovering at a prescribed roll angle with zero torque demand, no external disturbance,
    $\mathbf W=\mathbf I_{18}$, and inactive thrust limits, the uncorrected PINV solution yields
    $\sin\lambda_i=\sin\delta_i\sin\phi$ and the MATC-limited boundary
    $\phi_{\max}^{\mathrm{PINV}}=\arcsin(\tfrac{2}{\sqrt3}\sin\lambda_{\max})$.
    For $|\phi|>\phi_{\max}^{\mathrm{PINV}}$, at least one commanded $\lambda_i$ exceeds $\lambda_{\max}$.
\end{revblock}
\subsubsection{MATC Handling}
\begin{revblock}
    The MATC implied by \eqref{eq:cl-actuator-command-conversion} is
    \begin{equation}
        {}^{A_i}\!f_{\mathrm{ae},x} \geq -\sqrt{\left({}^{A_i}\!f_{\mathrm{ae},y}\right)^2 + \left({}^{A_i}\!f_{\mathrm{ae},z}\right)^2}\,\tan\left(\lambda_{\max}\right)
        \label{eq:cl-matc-force-constraint}
    \end{equation}

    Let
    $\boldsymbol f_x^0
        =({}^{A_1}\!f_{\mathrm{ae},x}^0,\ldots,{}^{A_6}\!f_{\mathrm{ae},x}^0)^\top
        =\mathbf S_x\boldsymbol u_f^0\in\mathbb R^6$,
    where $\mathbf S_x\in\mathbb R^{6\times18}$ selects the six local $x$-force components.
    The corresponding effectiveness matrix
    $\mathbf B_x=\mathbf B_0\mathbf S_x^\top\in\mathbb R^{6\times6}$ is
    \begin{equation}
        \mathbf B_x \!=\!
        \begin{bmatrix}
            1 \!      & \! -1 \!     & \! -\tfrac{1}{2} \!              & \! \tfrac{1}{2} \!               & \! \tfrac{1}{2} \!              & \! -\tfrac{1}{2}             \\[0.5mm]
            0 \!      & \! 0 \!      & -\tfrac{\sqrt{3}}{2} \!          & \! \tfrac{\sqrt{3}}{2} \!        & \! -\tfrac{\sqrt{3}}{2} \!      & \! \tfrac{\sqrt{3}}{2}       \\
            0 \!      & \! 0  \!     & \! 0 \!                          & \! 0 \!                          & \! 0 \!                         & \! 0                         \\
            c_\tau \! & \! c_\tau \! & \! -\tfrac{c_\tau}{2} \!         & \! -\tfrac{c_\tau}{2} \!         & \! -\tfrac{c_\tau}{2} \!        & \! -\tfrac{c_\tau}{2}        \\[0.5mm]
            0 \!      & \! 0  \!     & \! -\tfrac{\sqrt{3}c_\tau}{2} \! & \! -\tfrac{\sqrt{3}c_\tau}{2} \! & \! \tfrac{\sqrt{3}c_\tau}{2} \! & \! \tfrac{\sqrt{3}c_\tau}{2} \\
            0 \!      & \! 0  \!     & \! 0 \!                          & \! 0 \!                          & \! 0 \!                         & \! 0
        \end{bmatrix}
    \end{equation}

    A nullspace basis is
    \begin{equation}
        \mathbf N_x \!=\!
        \begin{bmatrix}
            1 \! & \!  0 \! & \! 1 \!  & \!  0 \! & \! 0 \! & \! 1 \\
            0 \! & \! 1 \!  & \!  0 \! & \! 1 \!  & \! 1 \! & \! 0
        \end{bmatrix}^{\top} \in \mathbb R^{6\times 2}
    \end{equation}
    Its columns redistribute the local $x$ forces within actuator groups
    $\{1,3,6\}$ and $\{2,4,5\}$, respectively, and satisfy
    $\mathbf B_x\mathbf N_x=\mathbf0_{6\times2}$.
    Hence, parameterize the corrected local $x$-force command as
    \begin{equation}
        \boldsymbol f_x^c = \boldsymbol f_x^0 + \mathbf N_x \boldsymbol \xi
    \end{equation}
    where $\boldsymbol\xi=\operatorname{col}(\xi_1,\xi_2)\in\mathbb R^2$.
    To enforce \eqref{eq:cl-matc-force-constraint}, choose
    \begin{equation}
        \left\{
        \begin{aligned}
            \xi_1 & = \max\left(0, \rho_{x,1},\rho_{x,3},\rho_{x,6}\right) \\
            \xi_2 & = \max\left(0, \rho_{x,2},\rho_{x,4},\rho_{x,5}\right)
        \end{aligned}
        \right.
        \label{eq:cl-matc-correction-magnitude}
    \end{equation}
    where $\rho_{x,i}\triangleq -\sqrt{\left({}^{A_i}\!f_{\mathrm{ae},y}^{0}\right)^2 \!+\! \left({}^{A_i}\!f_{\mathrm{ae},z}^{0}\right)^2}\tan\left(\lambda_{\max}\right) - {}^{A_i}\!f_{\mathrm{ae},x}^{0}$
    is the initial residual of \eqref{eq:cl-matc-force-constraint} for BTAU $i$.

    Finally, the actuator force command vector is corrected as
    \begin{equation}
        \boldsymbol u_f^c = \boldsymbol u_f^0 + \mathbf S_x^{\top}\mathbf N_x\boldsymbol \xi
        \label{eq:cl-matc-force-command}
    \end{equation}

    Because $\mathbf B_0\mathbf S_x^\top\mathbf N_x=\mathbf0$, the MATC correction preserves the nominal wrench and enforces \eqref{eq:cl-matc-force-constraint} for all six BTAUs.
    Under these hovering conditions, it extends the MATC-feasible range beyond the PINV boundary.
    Final actuator commands are reconstructed through \eqref{eq:cl-actuator-command-conversion}.
    The remaining actuator bounds in \eqref{eq:aws-actuator-force-set} are enforced by saturation, and the resulting realization mismatch enters $\boldsymbol D$ under the conditions below.
    Under unknown rotor failures, the correction need not lie in the nullspace of $\mathbf B_0\boldsymbol\Lambda$, and the resulting wrench mismatch enters $\boldsymbol D_f$ in \eqref{eq:cl-disturbance-decomposition}.
\end{revblock}
\subsection{Stability Analysis}
\begin{revblock}
Let $t_0=0$ and $0<t_1<\cdots<t_{N_f}$ denote the finitely many failure instants, 
and set $t_{N_f+1}=\infty$.
The analysis is conducted under Assumption~\ref{ass:attitude-domain} and the following assumptions:
\begin{assumption}[Inter-fault disturbance regularity]
    On each open inter-fault interval $(t_k,t_{k+1})$,
    the CL-PFTC lumped disturbance $\boldsymbol D$ is
    locally absolutely continuous and satisfies $\|\dot{\boldsymbol D}(t)\|_2\leq\overline h_D$,
    where $\overline h_D\geq0$ is finite and independent of $k$.
    \label{ass:cl-disturbance-regularity}
\end{assumption}

\begin{assumption}[Local command admissibility]
    The reference in \eqref{eq:cl-hofa-control-law} and \(\boldsymbol D\) lie in prescribed compact sets.
    Uniformly over these sets and the considered fault modes, 
    compatible zero-error tracking states lie in the attitude-error domain,
    with compensated raw wrenches strictly within command limits.
    The command map is uniformly locally Lipschitz near these states.
    \looseness=-1
    \label{ass:cl-local-command-admissibility}
\end{assumption}

Let $\widetilde{\boldsymbol D}=\boldsymbol D-\hat{\boldsymbol D}\in\mathbb R^6$ and
$\boldsymbol e=\operatorname{col}(\boldsymbol e_1,\dot{\boldsymbol e}_1)\in\mathbb R^{12}$.
While $\boldsymbol s_w=\mathbf0$, substituting \eqref{eq:cl-hofa-control-law} into \eqref{eq:cl-controller-error-dynamics} yields
\begin{equation}
    \ddot{\boldsymbol e}_1 + \mathbf A_1\dot{\boldsymbol e}_1 + \mathbf A_0\boldsymbol e_1 = \widetilde{\boldsymbol D}
    \label{eq:cl-closed-loop-tracking-error}
\end{equation}

\begin{remark}[Fault-induced error jumps]
\label{rem:cl-fault-jumps}
At a failure instant $t_k$, the pose, velocity, and LESO
states remain continuous, while $\boldsymbol D$ may undergo
a finite jump
$\Delta\boldsymbol D_k
=\boldsymbol D(t_k^+)-\boldsymbol D(t_k^-)$.
Since $\hat{\boldsymbol D}$ is continuous, the
disturbance-estimation error satisfies
\[
\widetilde{\boldsymbol D}(t_k^+)
=\widetilde{\boldsymbol D}(t_k^-)
+\Delta\boldsymbol D_k.
\]
\end{remark}

With
$\widetilde{\boldsymbol z}
    =\operatorname{col}(\widetilde{\boldsymbol z}_1,\widetilde{\boldsymbol z}_2,\widetilde{\boldsymbol z}_3)\in\mathbb R^{18}$,
$\widetilde{\boldsymbol z}_j=\boldsymbol z_j-\hat{\boldsymbol z}_j$, $j=1,2,3$,
define
$\boldsymbol\zeta=\operatorname{col}(\boldsymbol e,\widetilde{\boldsymbol z})\in\mathbb R^{30}$.
Under these conditions, \eqref{eq:cl-leso} and \eqref{eq:cl-closed-loop-tracking-error}
give the augmented closed-loop error system
\begin{equation}
\dot{\boldsymbol\zeta}=
\underbrace{\begin{+bmatrix}[
        colsep=4pt, rowsep=0pt,
        hline{3}={0.3pt,dashed}, vline{3}={0.3pt,dashed}]
    \mathbf0_6 & \mathbf I_6 & \mathbf0_6 & \mathbf0_6 & \mathbf0_6     \\
    -\mathbf A_0 & -\mathbf A_1 & \mathbf0_6 & \mathbf0_6 & \mathbf I_6 \\
    \mathbf0_6 & \mathbf0_6 & -\mathbf L_1 & \mathbf I_6 & \mathbf0_6   \\
    \mathbf0_6 & \mathbf0_6 & -\mathbf L_2 & \mathbf0_6 & \mathbf I_6   \\
    \mathbf0_6 & \mathbf0_6 & -\mathbf L_3 & \mathbf0_6 & \mathbf0_6
    \end{+bmatrix}}_{\mathbf A_{cl}}
\boldsymbol\zeta+
\underbrace{\begin{bmatrix}
        \mathbf0_6 \\ \mathbf0_6\\ \mathbf0_6\\ \mathbf0_6\\ \mathbf I_6
    \end{bmatrix}}_{\mathbf E_{cl}}
\boldsymbol h_D
\label{eq:cl-augmented-error-system}
\end{equation}
where the block triangular matrix $\mathbf A_{cl}$ is Hurwitz under the stated gain conditions.
Consequently, \eqref{eq:cl-augmented-error-system} is exponentially input-to-state stable (ISS) with respect to $\boldsymbol h_D$~\cite{sontagCharacterizationsInputtoState1995}.
By Remark~\ref{rem:cl-fault-jumps}, $\boldsymbol\zeta(t_k^+)=\boldsymbol\zeta(t_k^-) + \mathbf E_{cl}\Delta\boldsymbol D_k$.
For some constants $\kappa\geq1$ and $\alpha_{\mathrm{cl}}>0$ independent of the inter-fault interval,
\begin{equation}
    \begin{aligned}
        \|\boldsymbol\zeta(t)\|_2
        \leq{} &
        \kappa e^{-\alpha_{\mathrm{cl}} t}
        \|\boldsymbol\zeta(0^+)\|_2
        +\frac{\kappa}{\alpha_{\mathrm{cl}}}\overline h_D
        \\
                & +\kappa\sum_{0<t_k\leq t}
        e^{-\alpha_{\mathrm{cl}}(t-t_k)}
        \|\Delta\boldsymbol D_k\|_2 .
    \end{aligned}
    \label{eq:cl-inter-fault-error-bound}
\end{equation}

Assumption~\ref{ass:cl-local-command-admissibility} gives a uniform neighborhood of $\boldsymbol\zeta=\mathbf0$ within the attitude-error domain where the wrench command is unsaturated.
For sufficiently small initial augmented error, $\overline h_D$, and $\Delta\boldsymbol D_k$, \eqref{eq:cl-inter-fault-error-bound} excludes a first exit by the standard invariant-set argument~\cite{hindiAnalysisLinearSystems1998,huAnalysisDesignSaturation2002}.

Within this local regime, the jump term in \eqref{eq:cl-inter-fault-error-bound} decays exponentially after the final failure, so the tracking and observer errors are ultimately bounded, with
$\limsup_{t\to\infty}\|\boldsymbol e(t)\|_2 \leq\kappa\overline h_D/\alpha_{\mathrm{cl}}$.
If $\boldsymbol D$ becomes constant thereafter,
\eqref{eq:cl-augmented-error-system} gives exponential convergence of both errors to zero.

\end{revblock}

\section{AL-PFTC Design}\label{sec:al-pftc}
\begin{figure}[!htb]
    \centering
    \includegraphics[width=8.8cm]{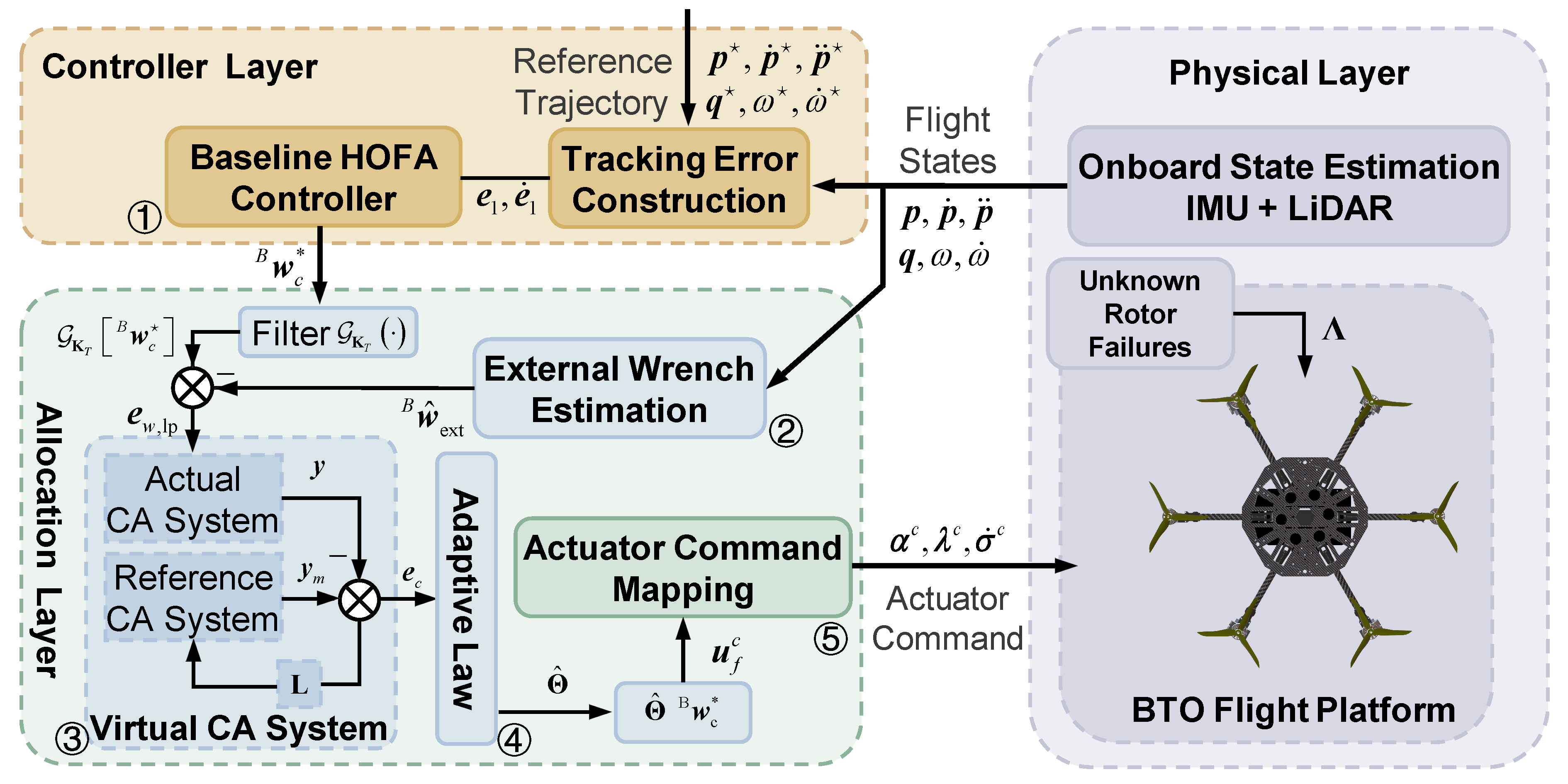}
    \caption{\rev{AL-PFTC architecture comprising
        \ding{172} baseline HOFA control~\eqref{eq:al-baseline-hofa-control-law},
        \ding{173} external-wrench estimation~\eqref{eq:al-torque-residual} and~\eqref{eq:al-force-estimator},
        \ding{174} virtual CA system~\eqref{eq:al-virtual-allocation-system},
        \ding{175} gated adaptive law~\eqref{eq:al-adaptive-law}, and
        \ding{176} actuator-command mapping~\eqref{eq:cl-actuator-command-conversion}.}}
    \label{fig:al-pftc-framework}

    \vspace{6pt}
    \begingroup
    \makeatletter\def\@captype{table}\makeatother
    \caption{\rev{Main Symbols and Notation for AL-PFTC}}
    \label{tab:al-pftc-notation}
    \centering
    \footnotesize
    \setlength{\tabcolsep}{2pt}
    \renewcommand{\arraystretch}{1.00}
    \begin{revtable}
        \begin{tabular}{@{}
            >{\centering\arraybackslash}m{0.14\columnwidth}
            >{\raggedright\arraybackslash}m{0.80\columnwidth}
            @{}}
            \toprule
            \textbf{Symbol} & \textbf{Definition}                                                           \\
            \midrule

            ${}^{B}\!\hat{\boldsymbol w}_{\mathrm{ext}}$
                            & External wrench estimate;
            ${}^{B}\!\hat{\boldsymbol w}_{\mathrm{ext}}\in\mathbb R^6$.                                     \\

            $\mathcal G_{\mathbf K_T}[\cdot]$
                            & Low-pass filter operator with bandwidth $k_\tau$ [$\mathrm{s}^{-1}$].         \\

            $\boldsymbol e_{w,\mathrm{lp}}$
                            & Filtered  wrench error;
            $\boldsymbol e_{w,\mathrm{lp}}\in\mathbb R^6$.                                                  \\

            $\boldsymbol e_c$
                            & Virtual CA error;
            $\boldsymbol e_c \in \mathbb R^6 $.                                                             \\

            $\hat{\boldsymbol\Theta}$
                            & Adaptive CA matrix;
            $\hat{\boldsymbol\Theta}\in\mathbb R^{18\times6}$.                                              \\

            $\mathbf Y_\Theta$
                            & Unconstrained adaptation direction of $\hat{\boldsymbol\Theta}$;
            $\mathbf Y_\Theta\in\mathbb R^{18\times6}$.                                                     \\

            $\mathbb B_{\Theta}$
                            & Common entrywise parameter box;
            $\mathbb B_{\Theta}\subset\mathbb R^{18\times6}$.                                               \\

            $\underline{\boldsymbol\Theta},\overline{\boldsymbol\Theta}$
                            & Lower and upper parameter limits;
            $\underline{\boldsymbol\Theta},\overline{\boldsymbol\Theta}\in\mathbb R^{18\times6}$.           \\

            $\boldsymbol e_{\mathcal G,k}$
                            & Filter commutation error;
            $\boldsymbol e_{\mathcal G,k}\in\mathbb R^6$.                                                   \\

            $\boldsymbol e_{w,a}$
                            & Mismatch between the commanded wrench and the external wrench;
            $\boldsymbol e_{w,a}\in\mathbb R^6$.                                                            \\

            $(\cdot)_{\mathrm{lp}}$
                            & Subscript denoting a quantity filtered by $\mathcal G_{\mathbf K_T}$.         \\

            $(\cdot)_{k}$
                            & Subscript denoting a quantity on the fault interval $\mathcal I_k$.           \\

            $[\cdot]_{a}$
                            & Notation denoting the $a$-th vector element.                                  \\

            $[\cdot]_{ab}$
                            & Notation denoting the matrix element in row $a$ and column $b$.               \\

            \bottomrule
        \end{tabular}
    \end{revtable}

    \endgroup
\end{figure}

\begin{revblock}
    As shown in Fig.~\ref{fig:al-pftc-framework}, AL-PFTC uses the nominal HOFA tracking law without LESO compensation
    and replaces the fixed allocators \eqref{eq:cl-weighted-pseudoinverse-allocation} and \eqref{eq:cl-matc-force-command}
    with an adaptive allocator driven by estimated wrench, without fault identification.
    Table~\ref{tab:al-pftc-notation} lists the notation.
    From \eqref{eq:cl-tracking-error-dynamics} and \eqref{eq:cl-inertia-model-uncertainty}, the baseline control law is
    \begin{equation}
        \begin{aligned}
            {}^{B}\!\boldsymbol w_c^{\star} & = \operatorname{Sat}_{\mathbf D_w}({}^{B}\!\boldsymbol w_c^{\mathrm{raw}}) \\
            {}^{B}\!\boldsymbol w_c^{\mathrm{raw}}
                                            & = -\mathbf M_B^{\mathrm n}\boldsymbol\Phi^{-1}
            \left(
            \mathbf A_1\dot{\boldsymbol e}_1
            +\mathbf A_0\boldsymbol e_1
            +\boldsymbol\nu_0
            \right),                                                                                                          \\
            \boldsymbol\nu_0
                                            & \triangleq
            \boldsymbol\Phi\left(\mathbf M_B^{\mathrm n}\right)^{-1}
            \left(
            {}^{B}\!\boldsymbol w_{\mathrm{gra}}^{\mathrm n}
            +{}^{B}\!\boldsymbol w_{\mathrm{rot}}^{\mathrm n}
            \right)
            +\boldsymbol a
        \end{aligned}
        \label{eq:al-baseline-hofa-control-law}
    \end{equation}
\end{revblock}

\subsection{\rev{External-Wrench Feedback and Adaptive Allocation}}
\subsubsection{\rev{External-Wrench Estimation}}
\rev{AL-PFTC estimates the combined aerodynamic and disturbance wrench from onboard motion measurements.
Based on \eqref{eq:bto-newton-euler-dynamics}, the external wrench is}
\begin{equation}
    \rev{{}^{B}\!\boldsymbol w_{\mathrm{ext}}
    \triangleq
    {}^{B}\!\boldsymbol w_{\mathrm{ae}}+{}^{B}\!\boldsymbol w_d
    =
    \operatorname{col}\left(
    {}^{B}\!\boldsymbol f_{\mathrm{ext}},
    {}^{B}\!\boldsymbol t_{\mathrm{ext}}
    \right)
    \in\mathbb R^6}
    \label{eq:al-ext-wrench-definition}
\end{equation}

To avoid differentiating angular velocity,
a momentum-based method is used to estimate the external torque.
Define angular momentum \rev{$\boldsymbol h = \mathbf J_B^{\mathrm n} \boldsymbol\omega \in \mathbb R^3$}
\rev{using nominal rigid-body parameters.}
\rev{Under the nominal model, \eqref{eq:bto-newton-euler-dynamics} gives}
\begin{equation}
    \rev{\dot{\boldsymbol h} \!=\! \boldsymbol t_0 \!+\! {}^{B}\!\boldsymbol t_{\mathrm{ext}},\;
    \boldsymbol t_0 \!=\! {}^{B}\!\boldsymbol t_{\mathrm{gra}}^{\mathrm{n}} \!+\! {}^{B}\!\boldsymbol t_{\mathrm{rot}}^{\mathrm{n}}
    \!-\! m[{}^{B}\!\boldsymbol r_{\mathrm{com}}^{\mathrm{n}}]_{\times}{}^{B}\!\mathbf R_E\ddot{\boldsymbol p}}
    \label{eq:al-angular-momentum-dynamics}
\end{equation}

Following \cite{tomic2017external}, define the residual:
\begin{equation}
    \rev{\boldsymbol \rho_w \!=\! \mathbf K_\tau\left(\boldsymbol h - \boldsymbol h(0)
    \!-\! \int_{0}^{t}\left(\boldsymbol t_0 \!+\! \boldsymbol \rho_w\right)\mathrm{d}\tau\right),\; \mathbf K_\tau \triangleq k_\tau\mathbf I_3}
    \label{eq:al-torque-residual}
\end{equation}
where $k_\tau > 0$ is the bandwidth.
Differentiating \eqref{eq:al-torque-residual} gives
\begin{equation}
    \rev{\dot{\boldsymbol \rho}_w} = \mathbf K_\tau\left({}^{B}\!\boldsymbol t_{\mathrm{ext}} - \rev{\boldsymbol \rho_w}\right)
    \label{eq:al-torque-estimator}
\end{equation}
Thus \rev{${}^{B}\!\hat{\boldsymbol t}_{\mathrm{ext}} = \rev{\boldsymbol \rho_w}$} estimates ${}^{B}\!\boldsymbol t_{\mathrm{ext}}$ through a low-pass filter.
\rev{The force channel uses the same bandwidth:}
\begin{equation}
    \rev{{}^{B}\!\dot{\hat{\boldsymbol f}}_{\mathrm{ext}} =\mathbf K_\tau
    \left(
    m{}^{B}\!\mathbf R_E\ddot{\boldsymbol p}
    -{}^{B}\!\boldsymbol f_{\mathrm{gra}}^{\mathrm{n}}
    -{}^{B}\!\boldsymbol f_{\mathrm{rot}}^{\mathrm{n}}
    -{}^{B}\!\hat{\boldsymbol f}_{\mathrm{ext}}
    \right)}
    \label{eq:al-force-estimator}
\end{equation}

\begin{revblock}
    Define ${}^{B}\!\hat{\boldsymbol w}_{\mathrm{ext}} = \operatorname{col}({}^{B}\!\hat{\boldsymbol f}_{\mathrm{ext}},{}^{B}\!\hat{\boldsymbol t}_{\mathrm{ext}})$.
    Collect the omitted force-coupling term
    $m[{}^B\!\boldsymbol r_{\mathrm{com}}]_\times\dot{\boldsymbol\omega}$
    and nominal-parameter mismatch in the input-equivalent residual $\boldsymbol\epsilon_{\rm in}\in\mathbb R^{6}$.
    The estimates in \eqref{eq:al-torque-residual} and \eqref{eq:al-force-estimator} satisfy:
    \[
        {}^{B}\!\dot{\hat{\boldsymbol w}}_{\mathrm{ext}}
        =\mathbf K_T\bigl({}^{B}\!\boldsymbol w_{\mathrm{ext}}
        +\boldsymbol\epsilon_{\rm in}-{}^{B}\!\hat{\boldsymbol w}_{\mathrm{ext}}\bigr),
        \; \mathbf K_T\triangleq k_\tau\mathbf I_6
    \]

    To match the estimator bandwidth,
    define the low-pass filter
    $\boldsymbol x_{\mathrm{lp}} \triangleq \mathcal G_{\mathbf K_T}(\boldsymbol x)$ by
    $\dot{\boldsymbol x}_{\mathrm{lp}}=\mathbf K_T(\boldsymbol x-\boldsymbol x_{\mathrm{lp}})$.
    With ${}^{B}\!\boldsymbol w_{c,\mathrm{lp}}^\star \triangleq\mathcal G_{\mathbf K_T}({}^{B}\!\boldsymbol w_c^\star)$,
    the filtered wrench error is then defined as
    \begin{equation}
        \boldsymbol e_{w,\mathrm{lp}} \triangleq {}^{B}\!\boldsymbol w_{c,\mathrm{lp}}^{\star} - {}^{B}\!\hat{\boldsymbol w}_{\mathrm{ext}}
        \label{eq:al-filtered-output-wrench-error}
    \end{equation}
    With ${}^B\!\boldsymbol w_{c,\mathrm{lp}}^\star(0)=\mathbf0$,
    the saturated input keeps the filtered command within the prescribed command box for all $t\ge0$.
    \subsubsection{Virtual Control-Allocation System and Adaptive Law}
    The virtual CA system and closed-loop reference model
    generate the adaptation error\textcolor{black}{~\cite{gibsonAdaptiveOutputFeedback2015}}:
\end{revblock}
\begin{equation}
    \left\{
    \begin{aligned}
        \dot{\boldsymbol y}   & = \mathbf A_m\boldsymbol y\mathbin{\rev{-}}\rev{\boldsymbol e_{w,\mathrm{lp}}}                       \\
        \dot{\boldsymbol y}_m & = \mathbf A_m\boldsymbol y_m + \mathbf L\left(\boldsymbol y_m - \boldsymbol y\right)
    \end{aligned}
    \right.
    \label{eq:al-virtual-allocation-system}
\end{equation}
\begin{revblock}
    where $\boldsymbol y,\boldsymbol y_m\in\mathbb R^6$ are the virtual CA and reference model states, respectively.
    Choose a Hurwitz matrix $\mathbf A_m\in\mathbb R^{6\times6}$ and a gain $\mathbf L\in\mathbb R^{6\times6}$
    so that $\mathbf A_L=\mathbf A_m+\mathbf L$ is Hurwitz.
    The adaptation error $\boldsymbol e_c\triangleq\boldsymbol y_m-\boldsymbol y\in\mathbb R^6$ satisfies
\end{revblock}
\begin{equation}
    \dot{\boldsymbol e}_c = \rev{\mathbf A_L} \boldsymbol e_c + \rev{\boldsymbol e_{w,\mathrm{lp}}}
    \label{eq:al-virtual-allocation-error}
\end{equation}

\begin{revblock}
    Define the adaptive CA matrix $\hat{\boldsymbol\Theta}(t)\in\mathbb R^{18\times6}$, which produces the force commands $\boldsymbol u_f^c = \hat{\boldsymbol\Theta}{}^B\!\boldsymbol w_c^\star$.
    Apply \eqref{eq:cl-actuator-command-conversion}
    to each three-component block of $\boldsymbol u_f^c$
    to obtain the actuator commands, then limited by the bounds
    in \eqref{eq:aws-actuator-force-set}.

    The unconstrained update direction for $\hat{\boldsymbol\Theta}$ is
\end{revblock}
\begin{equation}
    \rev{\mathbf Y_\Theta} \triangleq \rev{\mathbf B_0^\top} \mathbf P_c \boldsymbol e_c \rev{({}^{B}\!\boldsymbol w_{c,\mathrm{lp}}^{\star})^\top} \boldsymbol\Gamma_c,\;
    \rev{\hat{\boldsymbol\Theta}(0) = \mathbf{B}_{W}^{\dagger}}
    \label{eq:al-adaptive-update-direction}
\end{equation}
\begin{revblock}
    where $\boldsymbol\Gamma_c\in\mathbb R^{6\times6}$ is a positive diagonal adaptation gain.
    The symmetric positive-definite matrices $\mathbf P_c,\mathbf Q_c\in\mathbb R^{6\times6}$ satisfy the Lyapunov equation:
\end{revblock}
\begin{equation}
    \rev{\mathbf A_L^{\top}\mathbf P_c + \mathbf P_c\mathbf A_L = - \mathbf{Q}_c}
    \label{eq:al-lyapunov-equation}
\end{equation}

\begin{revblock}
    Let $\boldsymbol\Theta_0\triangleq\hat{\boldsymbol\Theta}(0)$.
    For elementwise half-widths $[\boldsymbol\Delta_\Theta]_{ab}$,
    $a=1,\ldots,18$, $b=1,\ldots,6$, define the parameter box
    \begin{equation}
        \mathbb B_\Theta\triangleq
        \left\{\boldsymbol\Theta:
        |[\boldsymbol\Theta]_{ab}-[\boldsymbol\Theta_0]_{ab}|
        \leq[\boldsymbol\Delta_\Theta]_{ab},\ \forall a,b\right\}
        \label{eq:al-elementwise-parameter-bounds}
    \end{equation}

    The lower and upper bounds are $\underline{\boldsymbol\Theta}\!=\!\boldsymbol\Theta_0\!-\!\boldsymbol\Delta_\Theta$
    and $\overline{\boldsymbol\Theta}\!=\!\boldsymbol\Theta_0\!+\!\boldsymbol\Delta_\Theta$.
    Outward components are suppressed at these bounds:
    \begin{equation}
        \begin{aligned}
            \relax [\dot{\hat{\boldsymbol \Theta}}]_{ab} & \!=\!
            \begin{cases}
                0, \!                         & \! [\hat{\boldsymbol \Theta}]_{ab} \!\geq\! [\overline{\boldsymbol\Theta}]_{ab} \;\text{and}\; [\mathbf Y_{\Theta}]_{ab}\!>\!0  \\
                0, \!                         & \! [\hat{\boldsymbol \Theta}]_{ab} \!\leq\! [\underline{\boldsymbol\Theta}]_{ab} \;\text{and}\; [\mathbf Y_{\Theta}]_{ab}\!<\!0 \\
                [\mathbf Y_{\Theta}]_{ab}, \! & \! \text{otherwise}
            \end{cases} \\
                                                         & = g_{ab}[\mathbf Y_{\Theta}]_{ab} \;\; \text{for}\; a = 1,\ldots,18,\; b = 1,\ldots,6
        \end{aligned}
        \label{eq:al-adaptive-law}
    \end{equation}

    This update preserves $\hat{\boldsymbol\Theta}(t)\in\mathbb B_\Theta$ for all $t\geq0$.
\end{revblock}
\subsubsection{Offline Bounds for the Allocation Matrix}
\begin{revblock}
    The offline box must retain a right inverse for each admissible fault.
    For $\boldsymbol\Lambda_k\in\mathbb L_P$ in \eqref{eq:aws-admissible-effectiveness-set},
    let $\mathbf B_k\triangleq\mathbf B_0\boldsymbol\Lambda_k$.
    The AWS construction ensures that $\mathbf B_k$ has full row rank.
    Define a right inverse
    \begin{equation}
        \boldsymbol\Theta_k=\boldsymbol\Theta_0+
        \mathbf B_k^\dagger(\mathbf I_6-\mathbf B_k\boldsymbol\Theta_0)
        \in\mathbb R^{18\times6}
        \label{eq:al-representative-right-inverse}
    \end{equation}
    which satisfies $\mathbf B_k\boldsymbol\Theta_k=\mathbf I_6$.

    Choose the elementwise half-widths so that $\mathbb B_\Theta$ contains $\boldsymbol\Theta_k$ for every $\boldsymbol\Lambda_k\in\mathbb L_P$:
    \begin{equation}
        [\boldsymbol\Delta_\Theta]_{ab}\geq
        \max_{\boldsymbol\Lambda_k\in\mathbb L_P}
        \left|[\boldsymbol\Theta_k]_{ab}-[\boldsymbol\Theta_0]_{ab}\right|
        \label{eq:al-elementwise-parameter-half-width}
    \end{equation}
    $\mathbf B_k^\dagger$ and $\boldsymbol\Theta_k$ depend continuously on $\boldsymbol\Lambda_k$.
    Compactness of $\mathbb L_P$ therefore makes the maxima in \eqref{eq:al-elementwise-parameter-half-width} finite.

    To avoid computing 108 separate entrywise maxima,
    a conservative choice satisfying \eqref{eq:al-elementwise-parameter-half-width}
    uses six columnwise radii:
    \begin{equation}
        [\boldsymbol\Delta_\Theta]_{ab}=\kappa_\Theta R_{\theta,b}
        \geq\max_{\boldsymbol\Lambda_k\in\mathbb L_P}
        \|\boldsymbol\theta_{k,b}-\boldsymbol\theta_{0,b}\|_2
        \label{eq:al-columnwise-parameter-radius}
    \end{equation}
    where $\boldsymbol\theta_{k,b},\boldsymbol\theta_{0,b}\in\mathbb R^{18}$
    are the $b$-th columns and $\kappa_\Theta\geq1$ is a conservative margin.
    Bound each $R_{\theta,b}$ using a lower bound on the smallest singular value at
    each total-failure endpoint and the maximum norm of the affine residual
    over the corresponding box vertices.
    This uses positive-semidefinite monotonicity of $\mathbf B_k\mathbf B_k^\top$
    and convexity of the squared residual norm.
    \looseness=-1\par
\end{revblock}
\subsection{Stability Analysis}
\label{subsec:al-stability}
\begin{revblock}
    On each $\mathcal I_k=[t_k,t_{k+1})$,
    $\boldsymbol\Lambda_k\in\mathbb L_P$ is constant.
    Define the allocation mismatch
    $\mathbf M_k\triangleq
    \mathbf I_6-\mathbf B_k\hat{\boldsymbol\Theta}$.
    The analysis uses Assumption~\ref{ass:attitude-domain}
    and the following residual bound.
    \begin{assumption}[Disturbance boundedness]
        \label{ass:al-residual-boundedness}
        The residual disturbances entering the allocation and physical tracking-error dynamics, including model errors in the wrench estimator, are uniformly bounded on a prescribed candidate error domain \(\mathcal O\),
        under the prescribed parameter and command bounds.
    \end{assumption}

    \subsubsection{Augmented Allocation Error Dynamics}
    \label{subsubsec:al-augmented-dynamics}
    To analyze \eqref{eq:al-virtual-allocation-error},
    we derive a decomposition of the filtered wrench error $\boldsymbol e_{w,\mathrm{lp}}$.
    On $\mathcal I_k$, define the realization residual as
    \begin{equation}
        \boldsymbol r_{a,k}  = \mathbf B_k\hat{\boldsymbol\Theta}{}^{B}\!\boldsymbol w_c^\star-{}^{B}\!\boldsymbol w_{\mathrm{ext}}
        \label{eq:al-realization-residual}
    \end{equation}
    which includes actuator realization mismatch and external disturbances.
    Assumption~\ref{ass:al-residual-boundedness} gives a uniform bound
    $\overline r_a\ge0$ on $\|\boldsymbol r_{a,k}\|_2$ over $\mathcal O$.
    The filter definition, \eqref{eq:al-filtered-output-wrench-error},
   and \eqref{eq:al-realization-residual} give
    \begin{equation}
        \dot{\boldsymbol e}_{w,\mathrm{lp}} = -k_\tau \boldsymbol e_{w,\mathrm{lp}} + k_\tau\mathbf M_k{}^{B}\!\boldsymbol w_c^\star + k_\tau(\boldsymbol r_{a,k} - \boldsymbol\epsilon_{\rm in})
        \label{eq:al-filtered-wrench-mismatch-dynamics}
    \end{equation}

    To separate the adaptive mismatch from the residual input,
    introduce the auxiliary filter $\boldsymbol\upsilon\in\mathbb R^6$ by
    \begin{equation}
        \dot{\boldsymbol\upsilon}
        =-k_\tau\boldsymbol\upsilon
        +k_\tau\mathbf M_k{}^{B}\!\boldsymbol w_c^\star
        \label{eq:al-auxiliary-mismatch-filter}
    \end{equation}
    with finite initial state and no reset at faults.
    Define the filtered residual
    $\boldsymbol r_{\rm lp}\triangleq\boldsymbol e_{w,\mathrm{lp}}-\boldsymbol\upsilon$.
    Subtracting \eqref{eq:al-auxiliary-mismatch-filter}
    from \eqref{eq:al-filtered-wrench-mismatch-dynamics} gives
    \begin{equation}
        \dot{\boldsymbol r}_{\rm lp} =-k_\tau\boldsymbol r_{\rm lp}+k_\tau(\boldsymbol r_{a,k}-\boldsymbol\epsilon_{\rm in})
        \label{eq:al-joint-estimator-residual}
    \end{equation}

    Since $\mathbf M_k$ varies with the adaptive parameters,
    it generally cannot be taken outside the filter.
    To express $\boldsymbol\upsilon$ using the filtered command, define the
    filter commutation error $\boldsymbol e_{\mathcal G,k} \triangleq \mathbf M_k{}^{B}\!\boldsymbol w_{c,\mathrm{lp}}^\star - \boldsymbol\upsilon$.
    Differentiating this error and using the command-filter dynamics
    and \eqref{eq:al-auxiliary-mismatch-filter} gives
    \begin{equation}
        \begin{aligned}
            \dot{\boldsymbol e}_{\mathcal G,k}
            & = \dot{\mathbf M}_k{}^{B}\!\boldsymbol w_{c,\mathrm{lp}}^\star
            + \mathbf M_k{}^{B}\!\dot{\boldsymbol w}_{c,\mathrm{lp}}^\star
            - \dot{\boldsymbol\upsilon} \\
            & = -k_\tau\boldsymbol e_{\mathcal G,k}
            + \dot{\mathbf M}_k{}^{B}\!\boldsymbol w_{c,\mathrm{lp}}^\star
        \end{aligned}
        \label{eq:al-filtered-error-expression}
    \end{equation}

    For the remaining adaptive term, the elementwise law
    \eqref{eq:al-adaptive-law} gives, for $a=1,\ldots,18$,
    \begin{equation}
        [\dot{\hat{\boldsymbol{\Theta}}}{}^{B}\!\boldsymbol w_{c,\mathrm{lp}}^\star]_a
        = [\mathbf{B}_0^\top \mathbf P_c \boldsymbol e_c]_a
        \underbrace{\sum_{b=1}^{6}g_{ab}[\mathbf \Gamma_c]_{bb}([{}^{B}\!\boldsymbol w_{c,\mathrm{lp}}^\star]_b)^2}_{[\mathbf{D}_g]_{aa}}
        \label{eq:al-elementwise-allocator-error}
    \end{equation}
    where $\mathbf D_g\in\mathbb R^{18\times18}$ is diagonal.
    The saturation in \eqref{eq:al-baseline-hofa-control-law} and gate in \eqref{eq:al-adaptive-law} imply $0\preceq\mathbf D_g\preceq
    \overline\varpi_c\mathbf I_{18}$, where
    $\overline\varpi_c\triangleq
    \operatorname{tr}(\boldsymbol\Gamma_c\mathbf D_w^2)$.

    Since $\dot{\mathbf{M}}_k = -\mathbf{B}_0\boldsymbol{\Lambda}_k\dot{\hat{\boldsymbol{\Theta}}}$ on $\mathcal I_k$, \eqref{eq:al-elementwise-allocator-error} gives:
    \begin{equation}
        \dot{\mathbf M}_k{}^{B}\!\boldsymbol w_{c,\mathrm{lp}}^\star =
        -\mathbf{L}_{g,k}\mathbf P_c\boldsymbol e_c,\; \mathbf{L}_{g,k}\triangleq \mathbf B_0\boldsymbol\Lambda_k\mathbf D_g\mathbf B_0^\top
        \label{eq:al-elementwise-sector}
    \end{equation}

    The definitions above give
    $\boldsymbol e_{w,\mathrm{lp}}=\boldsymbol r_{\rm lp}
    +\mathbf M_k{}^{B}\!\boldsymbol w_{c,\mathrm{lp}}^\star
    -\boldsymbol e_{\mathcal G,k}$.
    Substituting this identity into \eqref{eq:al-virtual-allocation-error}
    and \eqref{eq:al-elementwise-sector} into
    \eqref{eq:al-filtered-error-expression} yields the coupled dynamics
    \begin{align}
        \dot{\boldsymbol e}_c
        & = \mathbf A_L\boldsymbol e_c+\boldsymbol r_{\rm lp}
        +\mathbf M_k{}^{B}\!\boldsymbol w_{c,\mathrm{lp}}^\star
        -\boldsymbol e_{\mathcal G,k}
        \label{eq:al-augmented-allocation-error-dynamics} \\
        \dot{\boldsymbol e}_{\mathcal G,k}
        & = -k_\tau\boldsymbol e_{\mathcal G,k}
        -\mathbf L_{g,k}\mathbf P_c\boldsymbol e_c
        \label{eq:al-filter-commutation-error-dynamics}
    \end{align}

    \subsubsection{Allocation Error Stability}
    \label{subsubsec:al-allocator-uub}
    Define $\tilde{\boldsymbol\Theta}_k\triangleq\boldsymbol\Theta_k-\hat{\boldsymbol\Theta}$,
    so that $\mathbf B_k\tilde{\boldsymbol\Theta}_k=\mathbf M_k$.
    Define $\mathbf R_k \triangleq \mathbf B_0\boldsymbol \Lambda_k \mathbf B_0^\top \succ \mathbf 0$ and $\mathbf R_0 = \mathbf B_0\mathbf B_0^\top$.
    Define the augmented allocation error $\boldsymbol x_a\triangleq\operatorname{col}(\boldsymbol e_c,\boldsymbol e_{\mathcal G,k}) \in \mathbb R^{12}$.
    Consider the Lyapunov function
    \begin{equation}
        V_{a,k}=\boldsymbol x_a^\top\mathbf S_{a,k}\boldsymbol x_a+W_k,\;
        W_k=\operatorname{tr}(\widetilde{\boldsymbol\Theta}_k^\top\boldsymbol\Lambda_k\widetilde{\boldsymbol\Theta}_k\boldsymbol\Gamma_c^{-1})
        \label{eq:al-allocation-storage}
    \end{equation}
    where $\mathbf S_{a,k}=\operatorname{blkdiag}(\mathbf P_c,c_{\mathcal G}\mathbf R_k^{-1})\in\mathbb R^{12\times 12}$, with $c_{\mathcal G}>0$.
    The box $\mathbb B_\Theta$ gives the finite bound $\overline W_k=\max_{\hat{\boldsymbol\Theta}\in\mathbb B_\Theta}W_k$, computable from its elementwise endpoints.

    Define the matrices
    \[
        \begin{aligned}
            \mathbf H_{g,k}  =\begin{bmatrix}
                \mathbf Q_c & \mathbf C_{g,k} \\
                \mathbf C_{g,k}^\top & 2c_{\mathcal G}k_\tau\mathbf R_k^{-1}
            \end{bmatrix},\;
            \mathbf C_{g,k}  =\mathbf P_c+c_{\mathcal G}\mathbf P_c\mathbf L_{g,k}\mathbf R_k^{-1}
        \end{aligned}
    \]

    \begin{lemma}[Allocation-layer dissipation certificate]
        \label{lem:al-allocation-dissipation}
        Let $\alpha_a>0$ satisfy $\mathbf Q_c-\alpha_a\mathbf P_c\succ\mathbf0$.
        The dissipation inequality
        \begin{equation}
            \mathbf H_{g,k}\succeq\alpha_a\mathbf S_{a,k},
            \qquad t\in\mathcal I_k,
            \label{eq:al-allocation-certificate}
        \end{equation}
        holds for all admissible fault modes and gate values if
        \begin{equation}
            c_{\mathcal G}=\overline\varpi_c^{-1},\qquad
            k_\tau>\frac{\alpha_a}{2}+2\kappa_0(\alpha_a)\overline\varpi_c
            \label{eq:al-filter-bandwidth-condition}
        \end{equation}
        where $\kappa_0(\alpha_a)\triangleq
        \|(\mathbf Q_c-\alpha_a\mathbf P_c)^{-1/2}
        \mathbf P_c\mathbf R_0^{1/2}\|_2^2$.
    \end{lemma}
    \begin{proof}
        For $c_{\mathcal G}>0$, apply the Schur complement to
        \eqref{eq:al-allocation-certificate}:
        \begin{equation}
            c_{\mathcal G}(2k_\tau-\alpha_a)\mathbf R_k^{-1}-\mathbf C_{g,k}^\top(\mathbf Q_c-\alpha_a\mathbf P_c)^{-1}\mathbf C_{g,k}\succeq\mathbf0
            \label{eq:al-schur-complement-condition}
        \end{equation}

        A congruence transformation with $\mathbf R_k^{1/2}$ gives
        \begin{equation}
            \left\|
            (\mathbf Q_c-\alpha_a\mathbf P_c)^{-1/2}\mathbf C_{g,k}\mathbf R_k^{1/2}
            \right\|_2^2 \leq c_{\mathcal G}(2k_\tau-\alpha_a)
            \label{eq:al-schur-norm-condition}
        \end{equation}
        Its left side factors as
        \[
            \left\| (\mathbf Q_c-\alpha_a\mathbf P_c)^{-1/2}\mathbf P_c\mathbf R_k^{1/2}(\mathbf I_6+c_{\mathcal G}\mathbf R_k^{-1/2}
            \mathbf L_{g,k}\mathbf R_k^{-1/2}) \right\|_2^2
        \]
        By \eqref{eq:al-elementwise-allocator-error} and \eqref{eq:al-elementwise-sector},
        $\mathbf0\preceq\mathbf R_k^{-1/2}\mathbf L_{g,k}\mathbf R_k^{-1/2}
        \preceq\overline\varpi_c\mathbf I_6$.
        Moreover, $\mathbf R_k\preceq\mathbf R_0$ for
        $\boldsymbol\Lambda_k\in\mathbb L_P$.
        Thus, \eqref{eq:al-schur-norm-condition} is ensured by
        \[
            k_\tau\ge\frac{\alpha_a}{2}
            +\frac{\kappa_0(\alpha_a)}{2c_{\mathcal G}}
            (1+c_{\mathcal G}\overline\varpi_c)^2.
        \]
        The right-hand side is minimized at
        $c_{\mathcal G}=\overline\varpi_c^{-1}$;
        the strict inequality in \eqref{eq:al-filter-bandwidth-condition}
        therefore ensures \eqref{eq:al-allocation-certificate}.
    \end{proof}

    \emph{Numerical verification.}
    For the considered platform, choose
    \[
        \begin{aligned}
            &\mathbf A_L\!=\!-\operatorname{diag}(130,130,90,30,30,70), \mathbf Q_c\!=\!\mathbf I_6,  \\
            &\boldsymbol\Gamma_c\!=\!\operatorname{diag}(15,15,15,10,10,5), \mathbf D_w\!=\!\operatorname{diag}(30,30,60,5,5,5).
        \end{aligned}
    \]
    Then, \eqref{eq:al-filter-bandwidth-condition} holds for some
    $\alpha_a>0$ whenever $k_\tau>31\,\mathrm{s}^{-1}$.

    \begin{theorem}[Allocation-error boundedness]
        \label{thm:al-allocation-boundedness}
        Under Assumption~\ref{ass:al-residual-boundedness}
        and the conditions of Lemma~\ref{lem:al-allocation-dissipation},
        let $\overline r_c\ge0$ uniformly bound
        $\|\boldsymbol r_{\rm lp}(t)\|_{\mathbf P_c}$
        before a possible exit from $\mathcal O$,
        including the filter initialization transient, where
        $\|\boldsymbol v\|_{\mathbf P_c}^2
        =\boldsymbol v^\top\mathbf P_c\boldsymbol v$.
        On $\mathcal I_k$, the following holds while the physical error remains in $\mathcal O$:
        \begin{equation}
            \begin{gathered}
                V_{a,k}(t) \leq e^{-\tfrac{\alpha_a}{2}(t-t_k)}V_{a,k}(t_k^+) + \overline V_{a,k}[1-e^{-\tfrac{\alpha_a}{2}(t-t_k)}]\\
                \overline V_{a,k}=\overline W_k+4\overline r_c^2/\alpha_a^2.
            \end{gathered}
            \label{eq:al-allocation-comparison}
        \end{equation}
        On the final interval, if the physical error remains in $\mathcal O$, $\boldsymbol e_c$ and $\boldsymbol e_{\mathcal G,k}$ are uniformly ultimately bounded (UUB) with
        \begin{equation}
            \begin{aligned}
                \limsup_{t\to\infty}\|\boldsymbol e_c(t)\|_2^2 & \leq\overline V_{a,k}\underline{p}_c^{-1},              \\
                \limsup_{t\to\infty}\|\boldsymbol e_{\mathcal G,k}(t)\|_2^2
                                                               & \leq \overline r_k c_{\mathcal G}^{-1}\overline V_{a,k}
            \end{aligned}
            \label{eq:al-allocation-ultimate-bounds}
        \end{equation}
        where $\underline{p}_c=\lambda_{\min}(\mathbf P_c)$
        and $\overline r_k=\lambda_{\max}(\mathbf R_k)$.
    \end{theorem}

    \begin{proof}
        By Assumption~\ref{ass:al-residual-boundedness}, the stable filter
        \eqref{eq:al-joint-estimator-residual} with finite initial state
        yields a uniform bound on $\boldsymbol r_{\rm lp}$
        before exit from $\mathcal O$.
        Since $\boldsymbol\Theta_k\in\mathbb B_\Theta$,
        the elementwise gate \eqref{eq:al-adaptive-law} satisfies
        \begin{equation}
            \left([\hat{\boldsymbol \Theta}]_{ab}-[\boldsymbol\Theta_k]_{ab}\right)
            \left([\dot{\hat{\boldsymbol \Theta}}]_{ab}-[\mathbf{Y}_{\Theta}]_{ab}\right)\leq 0
            \label{eq:al-gradient-condition}
        \end{equation}
        Using $\dot{\tilde{\boldsymbol\Theta}}_k=-\dot{\hat{\boldsymbol\Theta}}$
        and $\mathbf B_k\tilde{\boldsymbol\Theta}_k=\mathbf M_k$,
        \eqref{eq:al-gradient-condition} and \eqref{eq:al-adaptive-update-direction} give
        \begin{equation}
            \begin{aligned}
                \dot W_k
                =-2\operatorname{tr}\!\left(\tilde{\boldsymbol\Theta}_k^\top
                \boldsymbol\Lambda_k\dot{\hat{\boldsymbol\Theta}}\boldsymbol\Gamma_c^{-1}\right)
                 & \leq-2\operatorname{tr}\!\left(\tilde{\boldsymbol\Theta}_k^\top
                \boldsymbol\Lambda_k\mathbf Y_\Theta\boldsymbol\Gamma_c^{-1}\right) \\
                 & =-2\boldsymbol e_c^\top\mathbf P_c
                \mathbf M_k{}^{B}\!\boldsymbol w_{c,\mathrm{lp}}^\star
            \end{aligned}
            \label{eq:al-parameter-dissipation}
        \end{equation}

        Combining \eqref{eq:al-parameter-dissipation},
        \eqref{eq:al-augmented-allocation-error-dynamics}, and
        \eqref{eq:al-filter-commutation-error-dynamics}
        cancels the allocation-mismatch term in $\dot V_{a,k}$.
        Then \eqref{eq:al-allocation-certificate}
        and Young's inequality give
        \begin{equation}
            \begin{aligned}
                \dot V_{a,k}
                 & \leq-\boldsymbol x_a^\top\mathbf H_{g,k}\boldsymbol x_a
                +2\boldsymbol e_c^\top\mathbf P_c\boldsymbol r_{\rm lp}                      \\
                 & \leq-\frac{\alpha_a}{2}\boldsymbol x_a^\top\mathbf S_{a,k}\boldsymbol x_a
                +\frac{2}{\alpha_a}\|\boldsymbol r_{\rm lp}\|_{\mathbf P_c}^2
            \end{aligned}
            \label{eq:al-allocation-dissipation}
        \end{equation}
        Since $\boldsymbol x_a^\top\mathbf S_{a,k}\boldsymbol x_a
            =V_{a,k}-W_k\geq V_{a,k}-\overline W_k$,
        the comparison lemma gives \eqref{eq:al-allocation-comparison}.
        The definition of $\mathbf S_{a,k}$ then gives \eqref{eq:al-allocation-ultimate-bounds}.
    \end{proof}

    Theorem~\ref{thm:al-allocation-boundedness} bounds the allocation errors
    while the physical error remains in $\mathcal O$.
    The following joint analysis establishes domain invariance
    and physical regulation.
    \subsubsection{Physical Tracking Error Stability}
    \label{subsubsec:al-local-regulation}
    Consider a final fixed fault mode $k$
    and a constant reference with nominal wrench
    ${}^B\!\boldsymbol w_h\triangleq
    \left.{}^B\!\boldsymbol w_c^{\mathrm{raw}}\right|_{\boldsymbol e=\mathbf0}
    \ne\mathbf0$ satisfying $\|\mathbf D_w^{-1}{}^B\!\boldsymbol w_h\|_\infty<1$.
    To examine the coupling between adaptive allocation and post-fault regulation, 
    the local analysis uses the nominal rigid-body model
    with exact static actuator realization.

    \begin{assumption}[Post-fault disturbance conditions]
        \label{ass:al-local-residual-regularity}
        $T$ is the start of the local analysis in the final fixed fault mode.
        For the post-fault regulation task, the disturbance $\boldsymbol r_{a,k}$
        is locally absolutely continuous on $[T,\infty)$, with
        $\|\dot{\boldsymbol r}_{a,k}(t)\|_2\le\overline\nu$ almost everywhere, and
        $\int_T^\infty\|\dot{\boldsymbol r}_{a,k}(t)\|_2^2\,\mathrm dt\le E_\nu$,
        for finite nonnegative bounds $\overline\nu$ and $E_\nu$.
    \end{assumption}
    These bounds admit nonzero constant disturbances and finite-duration smooth transitions.

    From \eqref{eq:al-realization-residual}, the actual wrench mismatch is
    \begin{equation}
        \boldsymbol e_{w,a} = {}^B\!\boldsymbol w_c^\star-{}^B\!\boldsymbol w_{\rm ext} = \mathbf M_k{}^B\!\boldsymbol w_c^\star+\boldsymbol r_{a,k}
        \label{eq:al-actual-wrench-mismatch}
    \end{equation}
    With task command ${}^B\!\boldsymbol w_{\mathrm{task}} = {}^B\!\boldsymbol w_c^\star-{}^B\!\boldsymbol w_h$, define
    $\boldsymbol\iota_k=\mathbf M_k{}^B\!\boldsymbol w_h+\boldsymbol r_{a,k}$, giving
    \[
        \boldsymbol e_{w,a}=\boldsymbol\iota_k
        +\mathbf M_k{}^B\!\boldsymbol w_{\mathrm{task}}
    \]

    To distinguish reference wrench compensation from allocation changes in the complementary directions, introduce
    \begin{equation}
        \begin{aligned}
            \mathbf Z_k     & =\mathbf B_k(\hat{\boldsymbol\Theta}-\boldsymbol\Theta_0)\mathbf\Pi_h, & \mathbf\Pi_h & =\mathbf I_6-{}^B\!\boldsymbol w_h\boldsymbol v_h^\top                 \\
            \boldsymbol v_h & =\boldsymbol\Gamma_c{}^B\!\boldsymbol w_h/\beta_h,                     & \beta_h      & = ({}^B\!\boldsymbol w_h)^\top\boldsymbol\Gamma_c{}^B\!\boldsymbol w_h
        \end{aligned}
        \label{eq:al-reachable-task-coordinates}
    \end{equation}
    where $\mathbf Z_k \in \mathbb R^{6\times 6}$ describes changes in the effective allocation map on complementary wrench directions, with $\mathbf Z_k{}^B\!\boldsymbol w_h=\mathbf0$.
    $\mathbf\Pi_h\in\mathbb R^{6\times6}$ is the $\boldsymbol\Gamma_c$-weighted orthogonal projection onto the complement of the reference wrench direction,
    with $({}^B\!\boldsymbol w_h)^\top \boldsymbol\Gamma_c\mathbf\Pi_h=\mathbf0$.
    Any wrench $\boldsymbol w\in\mathbb R^6$ can therefore be written as
    $\boldsymbol w ={}^B\!\boldsymbol w_h(\boldsymbol v_h^\top\boldsymbol w) +\mathbf\Pi_h\boldsymbol w$.

    Define $\boldsymbol\ell_h \triangleq {}^B\!\boldsymbol w_{c,\rm lp}^\star-{}^B\!\boldsymbol w_h$.
    On an interval with unblocked updates, \eqref{eq:al-adaptive-update-direction} gives
    \begin{equation}
        \begin{aligned}
            \dot{\mathbf Z}_k & =\mathbf R_k\mathbf P_c\boldsymbol e_c
            \boldsymbol\ell_h^\top\boldsymbol\Gamma_c\mathbf\Pi_h       \\
            \dot{\boldsymbol\iota}_k
                              & =-\mathbf R_k\mathbf P_c\boldsymbol e_c
            (\beta_h+\boldsymbol\ell_h^\top\boldsymbol\Gamma_c{}^B\!\boldsymbol w_h)
            +\dot{\boldsymbol r}_{a,k}
        \end{aligned}
        \label{eq:al-reachable-adaptive-identities}
    \end{equation}

    To check parameter-box margins from these coordinates, define
    $\mathbf K_k=\mathbf B_0^\top\mathbf R_k^{-1}$,
    so $\mathbf K_k\mathbf B_k\mathbf B_0^\top=\mathbf B_0^\top$.
    If updates have remained unblocked since
    $\hat{\boldsymbol\Theta}(0)=\boldsymbol\Theta_0$,
    their accumulated increments retain the left factor
    $\mathbf B_0^\top$ and satisfy
    $\hat{\boldsymbol\Theta}-\boldsymbol\Theta_0 = \mathbf K_k\mathbf B_k (\hat{\boldsymbol\Theta}-\boldsymbol\Theta_0)$.
    Equation~\eqref{eq:al-reachable-task-coordinates} then gives
    \begin{equation}
        \hat{\boldsymbol\Theta}=\boldsymbol\Theta_0+\mathbf K_k
        \bigl(\mathbf Z_k + \bigl(
        (\mathbf I_6-\mathbf B_k\boldsymbol\Theta_0){}^B\!\boldsymbol w_h +\boldsymbol r_{a,k}-\boldsymbol\iota_k
        \bigr)\boldsymbol v_h^\top
        \bigr)
        \label{eq:al-reachable-parameter-reconstruction}
    \end{equation}
    At entry $T$, $\hat{\boldsymbol\Theta}(T)$ must satisfy
    \eqref{eq:al-reachable-parameter-reconstruction}.
    Unblocked updates preserve this identity.

    The same coordinates give the actual wrench mismatch:
    \[
        \begin{aligned}
            \boldsymbol e_{w,a}
            \!=\!\boldsymbol\iota_k\!+\!((\mathbf I_6\!-\!\mathbf B_k\boldsymbol\Theta_0)\mathbf\Pi_h\!-\!\mathbf Z_k
            \!+\!(\boldsymbol\iota_k\!-\!\boldsymbol r_{a,k})\boldsymbol v_h^\top)
            {}^B\!\boldsymbol w_{\mathrm{task}}
        \end{aligned}
    \]

    Define the augmented error $\boldsymbol x_k=\operatorname{col}(\boldsymbol e,\boldsymbol e_c,
        \boldsymbol\iota_k,\boldsymbol e_{w,\rm lp},\boldsymbol\ell_h)\in\mathbb R^{36}$
    and the command clipping error $\boldsymbol s_w\triangleq{}^B\!\boldsymbol w_c^{\mathrm{raw}}-{}^B\!\boldsymbol w_c^\star$.
    Thus ${}^B\!\boldsymbol w_c^{\mathrm{raw}}-{}^B\!\boldsymbol w_{\mathrm{ext}}
    =\boldsymbol s_w+\boldsymbol e_{w,a}$.
    The nominal observer residual is
    $\boldsymbol\epsilon_{\mathrm{in}}=\operatorname{col}
        (m[{}^B\!\boldsymbol r_{\mathrm{com}}^{\mathrm n}]_\times\dot{\boldsymbol\omega},\mathbf0_3)$,
    with $\dot{\boldsymbol\omega}$ evaluated from \eqref{eq:bto-newton-euler-dynamics}.
    Combining \eqref{eq:cl-tracking-error-dynamics}, \eqref{eq:al-baseline-hofa-control-law},
    \eqref{eq:al-virtual-allocation-error}, \eqref{eq:al-filtered-wrench-mismatch-dynamics},
    and \eqref{eq:al-reachable-adaptive-identities}, using the mismatch above and the low-pass filter
    $\dot{\boldsymbol\ell}_h=k_\tau({}^B\!\boldsymbol w_c^\star-{}^B\!\boldsymbol w_h-\boldsymbol\ell_h)$, gives
    \begin{equation}
        \dot{\boldsymbol x}_k
        =\boldsymbol\psi_k(\boldsymbol x_k,\mathbf Z_k,\boldsymbol r_{a,k})
        +\mathbf H_r\dot{\boldsymbol r}_{a,k}
        \label{eq:al-reachable-normal-form}
    \end{equation}
    where $\boldsymbol\psi_k\in\mathbb R^{36}$ collects the nominal dynamics, including command saturation.
    $\mathbf H_r\!=\!\operatorname{col}(\mathbf0_{18\times6},\mathbf I_6,\mathbf0_{12\times6})\in\mathbb R^{36\times 6}$.

    To formulate a joint dissipation condition, separate
    the terms linear in $\boldsymbol x_k$ from the nonlinear remainder.
    The strict reference margin makes saturation inactive near $\boldsymbol x_k=\mathbf0$.
    Since $\boldsymbol\psi_k(\mathbf0,\mathbf Z_k,\boldsymbol r_{a,k})=\mathbf0$,
    define the linear part and exact remainder in
    \eqref{eq:al-reachable-normal-form} by
    \begin{equation}
        \begin{aligned}
            \boldsymbol\Psi_k(\mathbf Z_k,\boldsymbol r_{a,k})
             & =
            \left.
            \frac{\partial \boldsymbol\psi_k
                (\boldsymbol x_k,\mathbf Z_k,\boldsymbol r_{a,k})}
            {\partial \boldsymbol x_k}
            \right|_{\boldsymbol x_k=\mathbf0} \in \mathbb R^{36\times 36}                               \\
            \boldsymbol n_k
             & =\boldsymbol\psi_k(\boldsymbol x_k,\mathbf Z_k,\boldsymbol r_{a,k}) 
            -\boldsymbol\Psi_k\boldsymbol x_k \in \mathbb R^{36}
        \end{aligned}
    \end{equation}
    Here $\mathbf Z_k$ and $\boldsymbol r_{a,k}$ are held fixed
    only for the partial derivative.
    For constant $\mathbf P_k\succ0$, define the Lyapunov function:
    \begin{equation}
        U_k=\boldsymbol x_k^\top\mathbf P_k\boldsymbol x_k
        \label{eq:al-physical-storage}
    \end{equation}

    Consider the candidate set $U_k\le\overline U_k$,
    $\|\mathbf Z_k\|_2\le\overline Z_k$,
    $\|\boldsymbol r_{a,k}\|_2\le\overline r_a$.
    For $a_c,b_\ell>0$, define 
    \[
        \begin{gathered}
            q_{Z,k}=\boldsymbol x_k^\top\mathbf Q_{Z,k}\boldsymbol x_k 
                =a_c\|\mathbf R_k\mathbf P_c\boldsymbol e_c\|_2^2 +b_\ell\|\mathbf\Pi_h^\top\boldsymbol\Gamma_c\boldsymbol\ell_h\|_2^2 \\
        \mathbf Q_{Z,k}\succeq0, \qquad \kappa_{Z,k}=(2\sqrt{a_cb_\ell})^{-1}
        \end{gathered} 
    \]

    \begin{lemma}[Sufficient conditions for local joint dissipation]
        \label{lem:al-local-joint-dissipation}
        With $\operatorname{He}(\mathbf X)=\mathbf X+\mathbf X^\top$,
        let $\gamma_k>0$ and choose joint dissipation rates $0<\lambda_{\mathrm{j},k}<\alpha_{\mathrm{j},k}$.
        If
        \begin{equation}
            \begin{gathered}
                \operatorname{He}(\mathbf P_k\boldsymbol\Psi_k)+\alpha_{\mathrm{j},k}\mathbf P_k+\mathbf Q_{Z,k}
                +\gamma_k^{-1}\mathbf P_k\mathbf H_r\mathbf H_r^\top\mathbf P_k\preceq\mathbf0,  \\
                2\boldsymbol x_k^\top\mathbf P_k\boldsymbol n_k
                \le(\alpha_{\mathrm{j},k}-\lambda_{\mathrm{j},k})U_k
            \end{gathered}
            \label{eq:al-reachable-dissipation-certificate}
        \end{equation}
        hold uniformly on the candidate set,
        then the unblocked joint dynamics \eqref{eq:al-reachable-adaptive-identities}
        and \eqref{eq:al-reachable-normal-form} satisfy
        \begin{subequations}
            \label{eq:al-local-dissipation-estimates}
            \begin{align}
                \dot U_k&\le-\lambda_{\mathrm{j},k} U_k-q_{Z,k}
                    +\gamma_k\|\dot{\boldsymbol r}_{a,k}\|_2^2,
                    \label{eq:al-local-storage-dissipation}\\
                \|\dot{\mathbf Z}_k\|_2&\le\kappa_{Z,k}q_{Z,k}
                    \label{eq:al-local-allocation-drift-rate}
            \end{align}
        \end{subequations}
        almost everywhere while the state remains in this set.

        Independently, on a compact candidate set inside the smooth domain
        of the nominal mechanics and the valid attitude chart, there is a uniform $C_k\ge0$ such that
        \begin{equation}
            2\boldsymbol x_k^\top\mathbf P_k\boldsymbol n_k
            \le C_kU_k^{3/2}
            \label{eq:al-reachable-remainder-bound}
        \end{equation}
        including any active clipping region.
        Hence, the second inequality in \eqref{eq:al-reachable-dissipation-certificate}
        is ensured by $C_k\sqrt{\overline U_k}\le\alpha_{\mathrm{j},k}-\lambda_{\mathrm{j},k}$.
    \end{lemma}
    \begin{proof}
        Differentiating $U_k$ along \eqref{eq:al-reachable-normal-form}
        and applying \eqref{eq:al-reachable-dissipation-certificate}
        and Young's inequality gives the first estimate.
        From \eqref{eq:al-reachable-adaptive-identities},
        the outer-product norm and weighted Young's inequality give
        \[
            \|\dot{\mathbf Z}_k\|_2
            =\|\mathbf R_k\mathbf P_c\boldsymbol e_c\|_2
            \|\mathbf\Pi_h^\top\boldsymbol\Gamma_c\boldsymbol\ell_h\|_2
            \le\kappa_{Z,k}q_{Z,k}
        \]

        For the nonlinear check, define
        $\mu_w=1-\|\mathbf D_w^{-1}{}^B\!\boldsymbol w_h\|_\infty>0$.
        ${}^B\!\boldsymbol w_c^{\rm raw}$ in \eqref{eq:al-baseline-hofa-control-law}
        is locally Lipschitz in $\boldsymbol x_k$ and equals ${}^B\!\boldsymbol w_h$ at $\boldsymbol x_k=\mathbf0$.
        Compactness and local Lipschitz continuity, together with
        $\mathbf P_k\succ0$, give a finite $L_w\ge0$ such that
        \[
            \|\mathbf D_w^{-1}({}^B\!\boldsymbol w_c^{\rm raw}
            -{}^B\!\boldsymbol w_h)\|_2\le L_w\sqrt{U_k}.
        \]

        First, normalized saturation projects onto a box containing
        the Euclidean ball of radius $\mu_w$ centered at
        $\mathbf D_w^{-1}{}^B\!\boldsymbol w_h$.
        The distance to the box is no greater than the distance to this ball.
        With $(a)_+=\max\{a,0\}$, this gives
        \begin{subequations}
            \label{eq:al-command-clipping-bound}
            \begin{equation}
                \|\mathbf D_w^{-1}\boldsymbol s_w\|_2
                \le (L_w\sqrt{U_k}-\mu_w)_+.
                \label{eq:al-command-clipping-distance}
            \end{equation}

            Second, applying $(a-\mu_w)_+\le a^2/(4\mu_w)$ for $a\ge0$
            with $a=L_w\sqrt{U_k}$ yields
            \begin{equation}
                \|\mathbf D_w^{-1}\boldsymbol s_w\|_2
                \le\frac{L_w^2}{4\mu_w}U_k.
                \label{eq:al-command-clipping-quadratic}
            \end{equation}
        \end{subequations}

        The smooth nominal terms have a uniform remainder of order
        $\|\boldsymbol x_k\|_2^2$.
        The wrench-mismatch decomposition above and the candidate-set
        bounds uniformly bound $\mathbf M_k$.
        Thus the matrices multiplying $\boldsymbol s_w$ are uniformly
        bounded throughout this set.
        Since $U_k$ is equivalent to $\|\boldsymbol x_k\|_2^2$,
        \eqref{eq:al-command-clipping-quadratic} makes the clipping contribution
        of the same order.
        Together these estimates give \eqref{eq:al-reachable-remainder-bound},
        and $U_k\le\overline U_k$ yields the stated sufficient check.
    \end{proof}

    \begin{theorem}[Local uniform ultimate boundedness]
        \label{thm:al-local-regulation-uub}
        Under Assumptions~\ref{ass:al-residual-boundedness}
        and~\ref{ass:al-local-residual-regularity}, suppose that, for constants
        $\gamma_k>0$ and $0<\lambda_{\mathrm{j},k}<\alpha_{\mathrm{j},k}$,
        \eqref{eq:al-reachable-dissipation-certificate} holds uniformly
        on the candidate set. Throughout this set, the reconstructed parameters
        lie strictly inside $\mathbb B_\Theta$, the actuator forces allocated
        from the saturated command satisfy the actuator constraints with strict
        margins, and the physical error projection lies strictly in $\mathcal O$
        within the attitude chart. Command clipping is allowed on this set.
        At the entry time $T$, suppose
        \eqref{eq:al-reachable-parameter-reconstruction} holds.
        Let $U_T\ge U_k(T)$ and $Z_T\ge\|\mathbf Z_k(T)\|_2$.
        With $\overline\nu$ and $E_\nu$ in
        Assumption~\ref{ass:al-local-residual-regularity}, suppose
        \begin{equation}
            \begin{aligned}
                \max\{U_T,\gamma_k\overline\nu^2/\lambda_{\mathrm{j},k}\} & <\overline U_k, \\
                Z_T+\kappa_{Z,k}(U_T+\gamma_kE_\nu)          & <\overline Z_k
            \end{aligned}
            \label{eq:al-reachable-entry-budgets}
        \end{equation}
        Then the solution remains in the candidate set for all $t\ge T$,
        and the augmented error $\boldsymbol x_k$ is UUB, with
        \begin{equation}
            \limsup_{t\to\infty}\|\boldsymbol x_k(t)\|_2^2
            \le\gamma_k\overline\nu^2\lambda_{\mathrm{j},k}^{-1}\underline{p}_k^{-1}
            \label{eq:al-local-physical-uub}
        \end{equation}
        where $\underline{p}_k$ is the minimum eigenvalue of $\mathbf P_k$.
    \end{theorem}
    \begin{proof}
        Until a possible first exit, the strict margins keep the adaptive update unblocked
        and preserve \eqref{eq:al-reachable-parameter-reconstruction} from $T$.
        Lemma~\ref{lem:al-local-joint-dissipation} gives the dissipation and drift estimates.
        Since $q_{Z,k}\ge0$ and
        $\|\dot{\boldsymbol r}_{a,k}\|_2\le\overline\nu$,
        exponential comparison gives
        $U_k(t)\le\max\{U_T,\gamma_k\overline\nu^2/\lambda_{\mathrm{j},k}\}$.

        To bound the accumulated drift of $\mathbf Z_k$, retain
        $q_{Z,k}$ and integrate
        \eqref{eq:al-local-storage-dissipation} from $T$ to $t$.
        Dropping the nonpositive terms $-U_k(t)$ and
        $-\lambda_{\mathrm{j},k}\int_T^t U_k(s)\,\mathrm ds$ gives
        \begin{equation}
            \begin{aligned}
                \int_T^t q_{Z,k}(s)\,\mathrm ds
                 & \le U_k(T)+\gamma_k\int_T^t\|\dot{\boldsymbol r}_{a,k}(s)\|_2^2\,\mathrm ds \\
                 & \le U_T+\gamma_k E_\nu
            \end{aligned}
            \label{eq:al-reachable-drift-integral}
        \end{equation}

        Since $\mathbf Z_k(t)=\mathbf Z_k(T)
        +\int_T^t\dot{\mathbf Z}_k(s)\,\mathrm ds$,
        the triangle inequality, \eqref{eq:al-local-allocation-drift-rate},
        and \eqref{eq:al-reachable-drift-integral} give
        \[
            \begin{aligned}
                \|\mathbf Z_k(t)\|_2
                 & \le Z_T+\int_T^t\|\dot{\mathbf Z}_k(s)\|_2\,\mathrm ds \\
                 & \le Z_T+\kappa_{Z,k}(U_T+\gamma_k E_\nu)
            \end{aligned}
        \]
        The strict inequalities in \eqref{eq:al-reachable-entry-budgets}
        keep $U_k$ and $\|\mathbf Z_k\|_2$ below their respective
        boundaries. The candidate-set margins keep the physical
        error and parameters within the domain of
        Assumption~\ref{ass:al-residual-boundedness},
        so its residual bound remains valid.
        These bounds exclude a first exit and hold for all $t\ge T$.
        Exponential comparison and
        $U_k\ge\underline{p}_k\|\boldsymbol x_k\|_2^2$
        give \eqref{eq:al-local-physical-uub}.
    \end{proof}
    The entry time $T$ may coincide with the final fault transition or follow
    a finite recovery transient, provided \eqref{eq:al-reachable-parameter-reconstruction}, \eqref{eq:al-reachable-entry-budgets} hold.

    \emph{Numerical illustration.}
    Consider nominal hover under $\overline{\boldsymbol\Lambda}_{1,2}$
    with the preceding numerical configuration and $k_\tau=150\,\mathrm{s}^{-1}$.
    For $\alpha_{\mathrm{j},k}=0.20\,\mathrm{s}^{-1}$ and $\lambda_{\mathrm{j},k}=0.10\,\mathrm{s}^{-1}$,
    the computed bound
    $C_k\sqrt{\overline U_k}\approx0.086\,\mathrm{s}^{-1}<\alpha_{\mathrm{j},k}-\lambda_{\mathrm{j},k}$
    satisfies the nonlinear-remainder condition in Lemma~\ref{lem:al-local-joint-dissipation}.

    Theorem 2 provides sufficient conditions for local post-fault regulation,
    with conservatism arising in part from uniform nonlinear-remainder bounds
    and norm-based estimates of the accumulated drift of $Z_k$.
    These conditions do not characterize the limits of achievable task performance.
    Further experiments in Section~\ref{sec:experiment} complement this analysis with experimental evaluations of
    fault recovery, trajectory tracking, and task execution under selected
    fault and disturbance conditions.

    \begin{remark}[Fault compensation mechanisms]
        \label{rem:al-cl-al-accommodation}
        CL-PFTC retains the nominal allocator and compensates fault-induced
        wrench mismatch through the LESO-based controller layer.
        For the nominal weighted-PINV branch before constraint correction,
        define
        $\mathbf G_k^{\mathrm{sc}}
            =\mathbf D_w^{-1}\mathbf B_k\boldsymbol\Theta_0\mathbf D_w$.
        Under $\overline{\boldsymbol\Lambda}_{4,5}$, the considered configuration
        gives $\sigma_{\min}(\mathbf G_k^{\mathrm{sc}})=0.463$,
        corresponding to an ideal maximum amplification of approximately $2.16$
        for normalized commands over all wrench directions, despite post-failure wrench feasibility.
        AL-PFTC updates the allocation map using estimated-wrench feedback;
        $\boldsymbol\iota_k$ captures
        allocation mismatch and residual disturbance jointly.
    \end{remark}

    \begin{remark}[Relation to existing adaptive methods]
        \label{rem:al-adaptive-control-positioning}
        Conventional model-reference adaptive control (MRAC) adapts controller parameters relative to a
        reference model; the $\mathcal L_1$ scheme
        in\textcolor{black}{~\cite{parkAdaptiveFaultTolerant2021}} estimates additive plant
        uncertainty and input effectiveness.
        AL-PFTC retains the nominal HOFA law and adapts only
        $\hat{\boldsymbol\Theta}$ through $\boldsymbol e_c$, driven by the difference between
        the filtered commanded wrench and the estimated external wrench.

        The virtual allocation system and closed-loop reference model
        follow\textcolor{black}{~\cite{tohidi2020adaptive,tohidiDiscreteAdaptiveControl2021}},
        shaping allocation-error dynamics.
        For AWS-admissible abrupt total rotor failures, six bounds on column radii
        are computed offline from right inverses.
        Including these right inverses in the parameter box ensures~\eqref{eq:al-gradient-condition},
        preserving cancellation of the mismatch cross term in the allocation
        Lyapunov derivative bound.
        The analysis retains estimator residuals and the dynamics of the filter commutation error.
        \looseness=-1\par

        Theorem~\ref{thm:al-allocation-boundedness} bounds augmented allocation
        errors while the physical error remains in $\mathcal O$.
        Theorem~\ref{thm:al-local-regulation-uub} establishes domain
        invariance and local UUB for constant-reference regulation
        in a final fixed fault mode, under the stated assumptions on nominal dynamics,
        exact static actuator realization, disturbances, and entry into the candidate set.
        Its candidate set keeps parameters strictly inside the box,
        leaving adaptation unblocked.
    \end{remark}

    \begin{remark}[Applicability to other platforms]
        \label{rem:al-platform-applicability}
        The architecture can be reused for systems admitting
        $\boldsymbol v=\mathbf B_{P,0}\boldsymbol\Lambda_k\boldsymbol u$,
        with $\mathbf B_{P,0}\in\mathbb R^{m\times n_P}$ and $n_P>m$.
        The admissible effectiveness matrices are nonnegative, diagonal,
        and piecewise constant, forming a compact set on which
        $\mathbf B_{P,0}\boldsymbol\Lambda_k$ retains full row rank.
        The parameter box must contain a right inverse for each mode.
        Each implementation requires checks of command and residual bounds,
        actuator feasibility, and allocation dissipation.
        The wrench map, estimator, actuator realization, and physical
        stability and entry conditions remain platform dependent.
    \end{remark}

\end{revblock}

\section{Simulations}\label{sec:simulations}
\rev{Two complementary Simscape Multibody studies assess 
(i) platform-level fault recovery under a shared CL-PFTC baseline and 
(ii) how CL-PFTC and AL-PFTC accommodate disturbances and rotor failures on the BTO.}
\rev{All simulations use the rigid-body parameters in Table~\ref{tab:aws-nominal-parameters} and the actuator constraints in \eqref{eq:aws-actuator-force-set}.}
\begin{revblock}
Position and attitude errors and their root-mean-square errors (RMSEs) quantify tracking performance:
\begin{equation}
    \left\{
        \begin{aligned}
            e_p(t) &\triangleq \|\boldsymbol p^\star - \boldsymbol p\|_2,\;
            e_\theta(t) \triangleq 2\arccos |(\boldsymbol q^\star)^\top\boldsymbol q | \\
            \mathcal R_{\mathcal T}(e(t)) &\triangleq \sqrt{\frac{1}{t_b-t_a}\int_{\mathcal T} e^2\mathrm{d}t}
        \end{aligned}
    \right.
    \label{eq:simulation-metric-tracking-errors}
\end{equation}
where $\mathcal T\triangleq [t_a,t_b]$ is the evaluation interval.
\end{revblock}
\begin{revblock}
\subsection{Fault-Recovery Comparison of Multirotor Configurations}
Within each roll case, the BTO, UTO, and CCU use identical rigid-body parameters, CL-PFTC gains, initial conditions, and hover commands, differing only in their actuator-force models and command mappings.
The total failure $\overline{\boldsymbol\Lambda}_4$ is injected at $t=t_f$ to evaluate fault recovery during level hovering and hovering at a roll angle of $45^\circ$.

Table~\ref{tab:simulation-platform-comparison} reports the BTO and UTO RMSEs over $\mathcal T_{s1}=[t_f,t_f+4\,\mathrm{s}]$.
The CCU is excluded because it cannot realize the $45^\circ$ hover and reaches the position-error safety threshold after $\overline{\boldsymbol\Lambda}_4$ (Fig.~\ref{fig:simulation-flight-platform-comparison}).
At $0^\circ$, the BTO and UTO have negligible fault-free errors and similar post-failure responses.
At $45^\circ$, the BTO reduces position and attitude RMSEs by 37.0\% and 23.5\%, respectively, with smaller peak transients than the UTO.

\begin{figure}[!b]
    \centering
    \begingroup
    \makeatletter\def\@captype{table}\makeatother
    \caption{\rev{Tracking RMSEs: fault-free (FF) and rotor~\#4 failure}}
    \label{tab:simulation-platform-comparison}
    \centering
    \footnotesize
    \renewcommand{\arraystretch}{0.9}
    \begin{threeparttable}[c]
        \begin{revtable}
        \begin{tabular}{@{}cccccc@{}}
            \toprule
            \multirow[c]{2}{*}[-0.7ex]{Roll}
            & \multirow[c]{2}{*}[-0.7ex]{Metric}
            & \multicolumn{2}{c}{BTO}
            & \multicolumn{2}{c}{UTO} \\
            \cmidrule(lr){3-4}\cmidrule(lr){5-6}
            & & $\mathbf{FF}$ &  $\overline{\boldsymbol\Lambda}_4$ & $\mathbf{FF}$ & $\overline{\boldsymbol\Lambda}_4$ \\
            \midrule
            \multirow[c]{2}{*}[-0.2ex]{$0^{\circ}$}
            & $\mathcal{R}_{\mathcal T_{s1}}(e_p)$ [m]
            & $<10^{-7}$ & 0.016
            & $<10^{-7}$ & 0.017 \\

            & $\mathcal{R}_{\mathcal T_{s1}}(e_\theta)$ [deg]
            & $<10^{-7}$ & 3.14
            & $<10^{-7}$ & 3.13 \\

            \multirow[c]{2}{*}[-0.2ex]{$45^{\circ}$}
            & $\mathcal{R}_{\mathcal T_{s1}}(e_p)$ [m]
            & $<10^{-7}$ & 0.017
            & $<10^{-7}$ & 0.027 \\

            & $\mathcal{R}_{\mathcal T_{s1}}(e_\theta)$ [deg]
            & $<10^{-7}$ & 2.64
            & $<10^{-7}$ & 3.45 \\

            \bottomrule
        \end{tabular}
    \end{revtable}
    \end{threeparttable}

    \endgroup

    \vspace{6pt}
    \includegraphics[width=75mm]{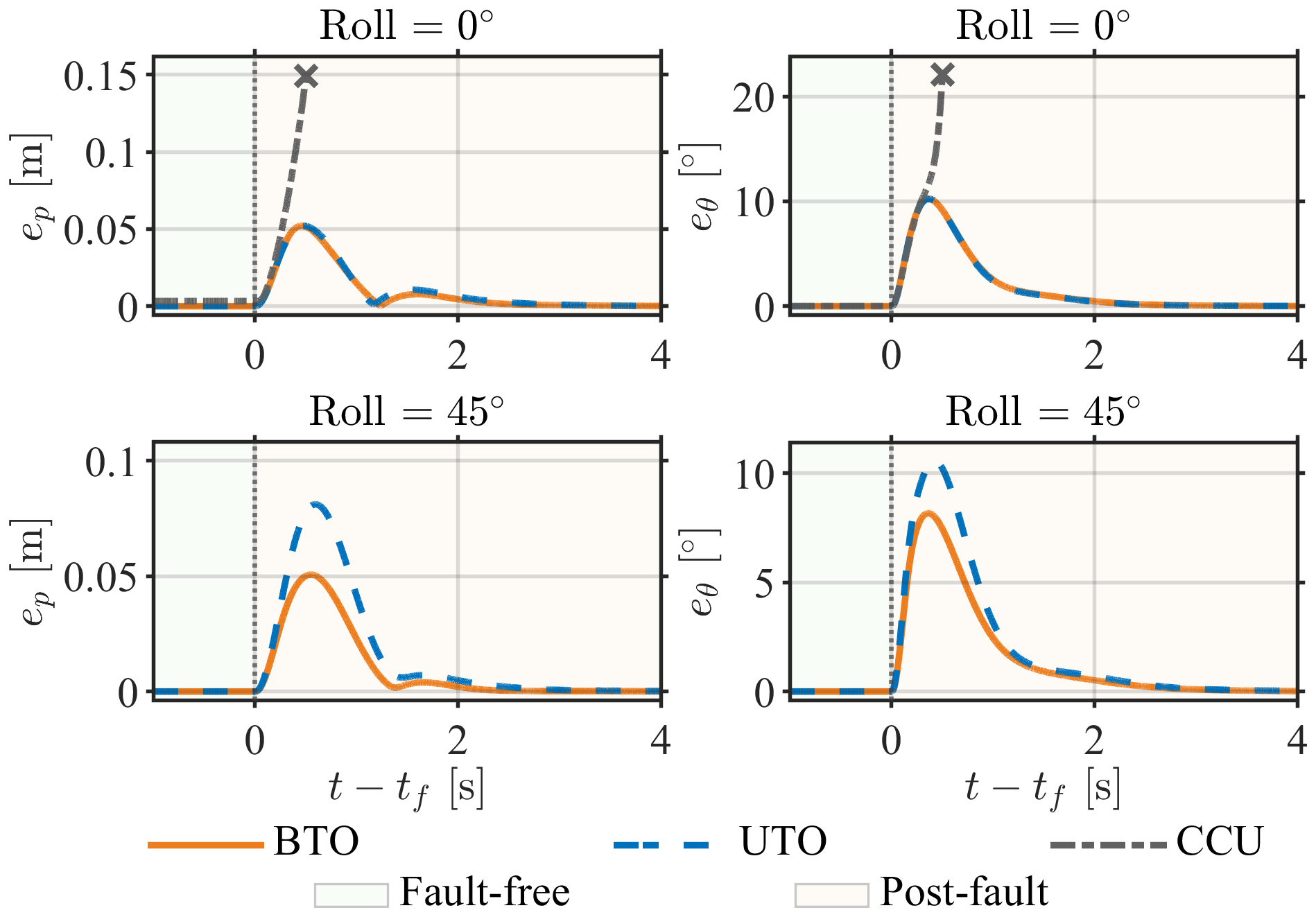}
    \caption{\rev{Fault recovery of the BTO, UTO, and CCU with CL-PFTC after total rotor~\#4 failure at $0^\circ$ and $45^\circ$ roll. The dotted line marks $t=t_f$; crosses mark when the position error first reaches the safety threshold.}}
    \label{fig:simulation-flight-platform-comparison}
\end{figure}
\end{revblock}
\subsection{\rev{Fault-Recovery Comparison of CL-PFTC and AL-PFTC}}
\label{subsec:simulation-cl-al-comparison}
\begin{revblock}
On the same BTO, CL-PFTC and AL-PFTC track
\eqref{eq:simulation-reference-trajectory} under the disturbance
and failure sequence below.
The fault $\overline{\boldsymbol\Lambda}_{4,5}$ passes
the horizontal-hover screening in
Section~\ref{sec:aws-fault-analysis} and illustrates attenuation
through nominal allocation in
Remark~\ref{rem:al-cl-al-accommodation}.
\end{revblock}
\begin{equation}
\begin{aligned}
\rev{\boldsymbol p^\star}(t)
&=
\begin{bmatrix}
A_x k(t)\sin(\omega_1 t) &
A_y k(t)\cos(\omega_1 t) &
A_z
\end{bmatrix}^{\top}
\\[-1mm]
\rev{\boldsymbol\vartheta^\star}(t)
&=
k(t)
\begin{bmatrix}
A_r\cos(\omega_2 t) &
A_p\sin(\omega_2 t) &
A_{\psi}t
\end{bmatrix}^{\top}
\end{aligned}
\label{eq:simulation-reference-trajectory}
\end{equation}
\rev{where $\boldsymbol\vartheta^\star(t)$
denotes the desired Euler-angle vector,} 
$k(t)=1-e^{-\rev{\beta_s} t^3}$ and $\rev{\beta_s}=0.05~\mathrm{s^{-3}}$. The trajectory parameters are \rev{$A_x=A_y=0.8~\mathrm{m}$}, \rev{$A_z=-0.8~\mathrm{m}$}, $\omega_1=\omega_2=\tfrac{\pi}{6}~\mathrm{rad/s}$, \rev{$A_r=A_p=\tfrac{\pi}{9}~\mathrm{rad}$}, and $A_{\psi}=0.035~\mathrm{rad/s}$.
\begin{revblock}
An external wrench disturbance ${}^B\!\boldsymbol w_d\triangleq \operatorname{col}({}^B\!\boldsymbol f_d,{}^B\!\boldsymbol t_d)$ is introduced at $t=2~\mathrm{s}$ and smoothly increased over $0.5~\mathrm{s}$ to
${}^B\!\boldsymbol f_d=[5,5,5]^\top\mathrm{N}$ and
${}^B\!\boldsymbol t_d=[0.5,0.5,0.5]^\top\mathrm{N\cdot m}$, after which it remains constant.
At $t=5~\mathrm{s}$, $\overline{\boldsymbol \Lambda}_{4,5}$ is injected while the disturbance persists.

Since the simulated and nominal aircraft models use identical physical parameters, the model-error term $\boldsymbol D_m$ is zero, isolating disturbance rejection and fault-induced allocation mismatch from nominal-model uncertainty.
The residual wrench
${}^B\!\boldsymbol w_r\triangleq
\operatorname{col}({}^B\!\boldsymbol f_r,{}^B\!\boldsymbol t_r)\in\mathbb R^6$
quantifies the uncompensated wrench mismatch in each closed-loop system:
\begin{equation}
    \begin{aligned}
        {}^B\!\boldsymbol w_{r,\mathrm{CL}}
        &= {}^B\!\boldsymbol w_{c,\mathrm{CL}}^\star
        + \mathbf M_B^\mathrm{n}\boldsymbol\Phi^{-1}\hat{\boldsymbol D}
        - {}^B\!\boldsymbol w_{\mathrm{ae},\mathrm{CL}}
        - {}^B\!\boldsymbol w_d,\\
        {}^B\!\boldsymbol w_{r,\mathrm{AL}}
        &= {}^B\!\boldsymbol w_{c,\mathrm{AL}}^\star
        - {}^B\!\boldsymbol w_{\mathrm{ae},\mathrm{AL}}
        - {}^B\!\boldsymbol w_d .
    \end{aligned}
\label{eq:simulation-residual-disturbance}
\end{equation}

For $j\in\{\mathrm{CL},\mathrm{AL}\}$,
${}^B\!\boldsymbol w_{\mathrm{ae},j}$ is reconstructed from realized actuator states.
For CL-PFTC, the reference wrench for this offline metric is formed by adding the LESO correction $\mathbf M_B^\mathrm{n}\boldsymbol\Phi^{-1}\hat{\boldsymbol D}$ to the commanded wrench ${}^B\!\boldsymbol w_{c,\mathrm{CL}}^\star$.
For AL-PFTC, the commanded wrench ${}^B\!\boldsymbol w_{c,\mathrm{AL}}^\star$ is used directly.
For AL-PFTC, ${}^B\!\boldsymbol w_{r,\mathrm{AL}}=\boldsymbol e_{w,a}$ in \eqref{eq:al-actual-wrench-mismatch}.
These residuals are offline evaluation metrics, not feedback signals.
The results are shown in Fig.~\ref{fig:simulation-method-comparison} and Table~\ref{tab:simulation-method-comparison} over the evaluation interval $\mathcal T_{s2}=[5,18]~\mathrm{s}$.

\begin{figure}[!b]
    \centering
    \includegraphics[width=75mm]{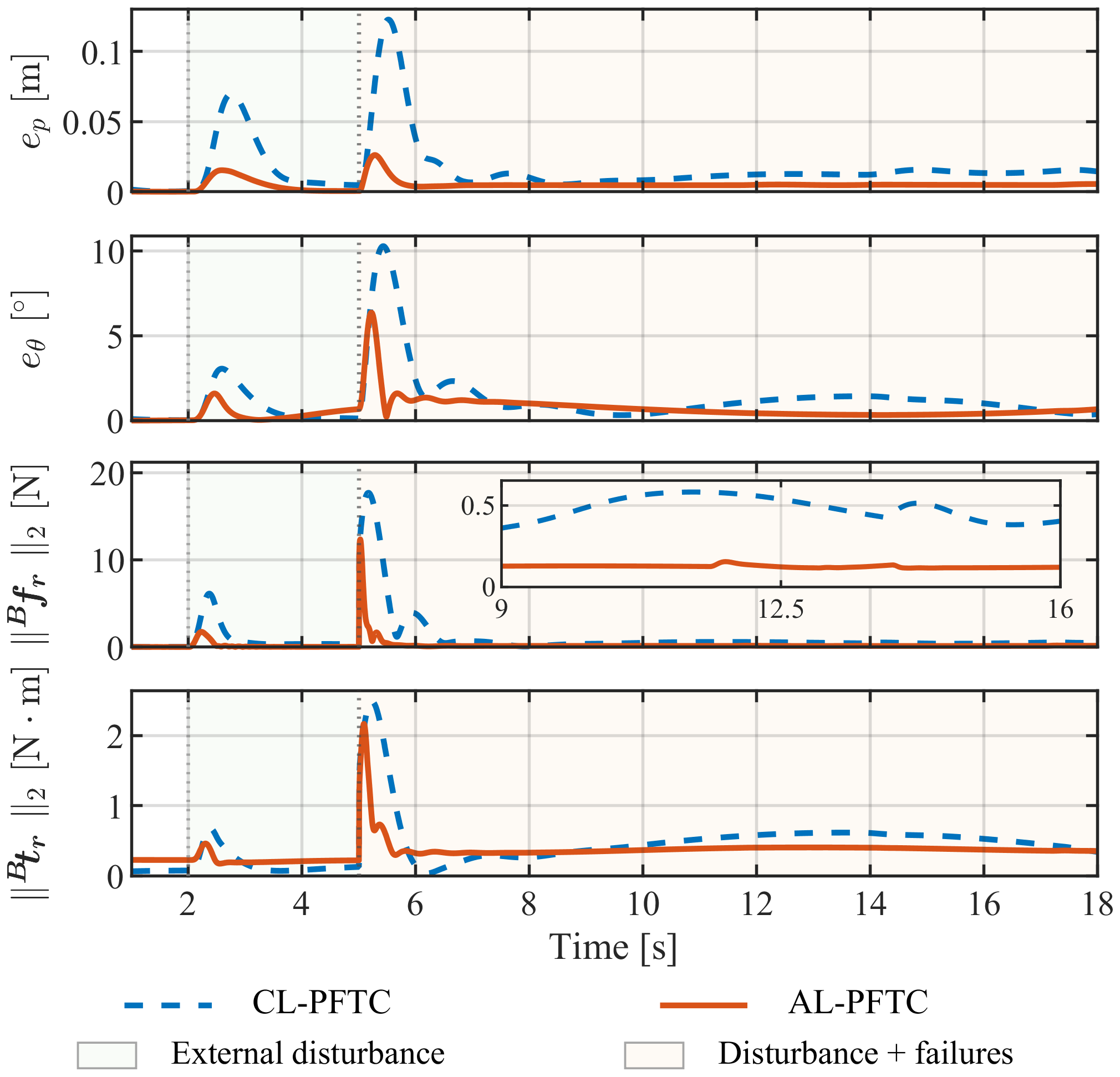}
    \caption{\rev{CL-PFTC and AL-PFTC responses, from top to bottom: position error,
        attitude error, force residual, and torque residual.
        Dotted lines mark disturbance onset at $2~\mathrm{s}$ and total failures
        of rotors~\#4 and \#5 at $5~\mathrm{s}$;
        shading marks periods with disturbance alone and with additional rotor failures.}}
    \label{fig:simulation-method-comparison}
\end{figure}

Figure~\ref{fig:simulation-method-comparison} shows smaller post-failure tracking errors and residual-wrench transients with AL-PFTC than with CL-PFTC.
The RMSEs of position error, attitude error, force residual, and torque residual decrease by 76.5\%, 52.9\%, 69.6\%, and 32.4\%, respectively (Table~\ref{tab:simulation-method-comparison}).
These results support the intended role of adaptive allocation in reducing persistent post-failure wrench mismatch.

\begin{table}[!htb]
    \caption{\rev{Post-failure RMSEs and reductions relative to CL-PFTC}}
    \label{tab:simulation-method-comparison}
    \centering
    \footnotesize
    \setlength{\tabcolsep}{3.0pt}
    \renewcommand{\arraystretch}{1.00}
    \begin{revtable}
    \begin{tabular}{@{}lccc@{}}
        \toprule
        \multirow{2}{*}[-0.7ex]{Metric}
        & \multicolumn{2}{c}{Method}
        & \multirow{2}{*}[-0.7ex]{AL-PFTC reduction} \\
        \cmidrule(lr){2-3}
        & AL-PFTC & CL-PFTC & \\
        \midrule
        $\mathcal{R}_{\mathcal T_{s2}}(e_p)\; [\mathrm{m}]$ 
        & 0.0062 & 0.0264 & 76.5\% \\

        $\mathcal{R}_{\mathcal T_{s2}}(e_\theta)\; [\mathrm{deg}]$ 
        & 1.0598 & 2.2506 & 52.9\% \\

        $\mathcal{R}_{\mathcal T_{s2}}(\|{}^B\!\boldsymbol f_r\|_2)\; [\mathrm{N}]$
        & 0.8989 & 2.9576 & 69.6\% \\

        $\mathcal{R}_{\mathcal T_{s2}}(\|{}^B\!\boldsymbol t_r\|_2)\; [\mathrm{N\cdot m}]$ 
        & 0.4362 & 0.6450 & 32.4\% \\
        \bottomrule
    \end{tabular}     
    \end{revtable}
\end{table}
\end{revblock}
 
\section{Experiments}\label{sec:experiment}
\begin{figure}[!htb]
  \centering
  \includegraphics[width=80mm]{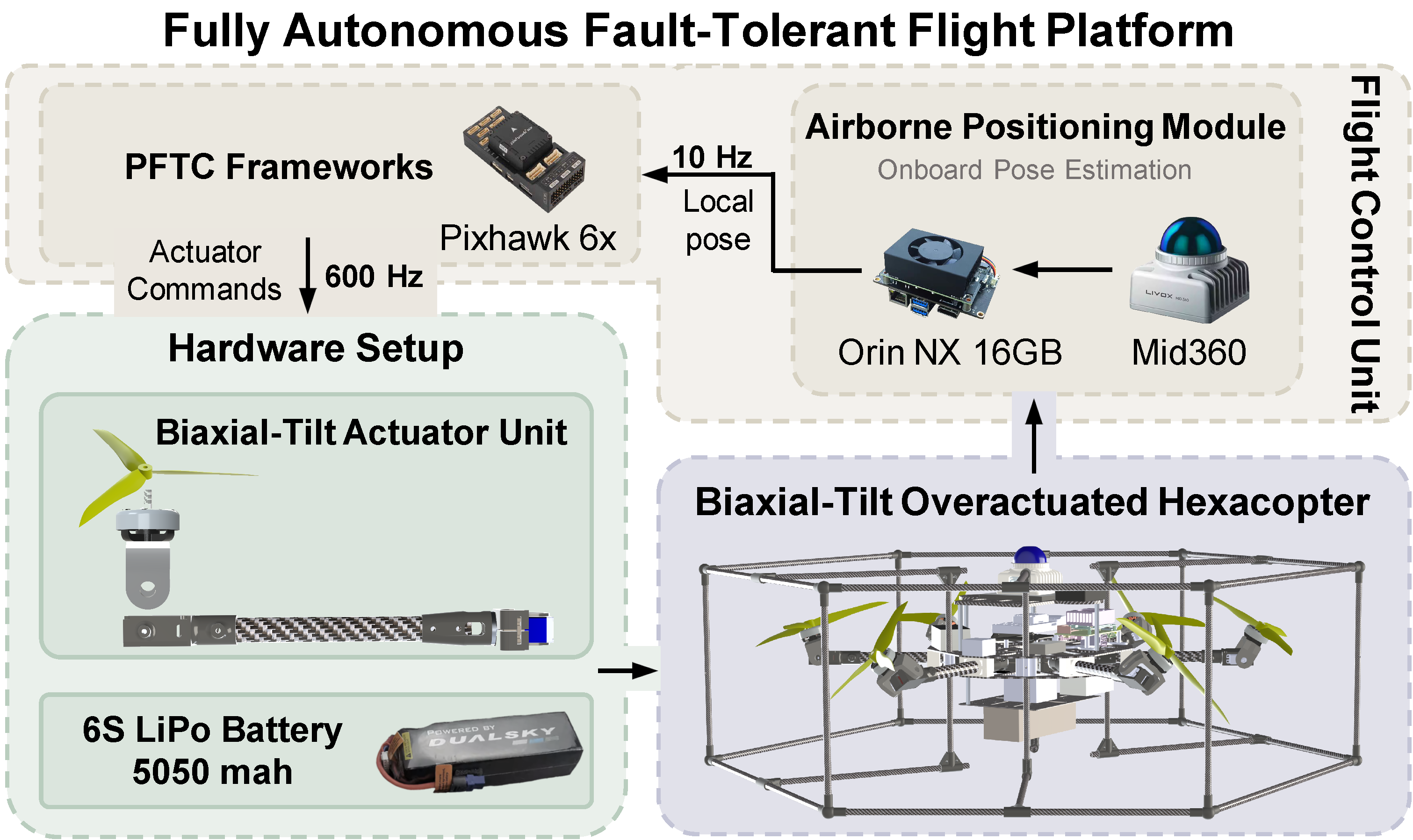}
  \caption{Experimental BTO platform and onboard control architecture.
  State estimation and flight control run onboard without external positioning.}
  \label{fig:experiment-flight-platform}
\end{figure}

\begin{revblock}
  Building on the simulations, hovering tests compare the PFTC methods and BTO/UTO configurations.
  Indoor 6-DOF tracking compares the methods on the BTO, while outdoor tracking and autonomous tasks assess the BTO with AL-PFTC under abrupt total failures of selected rotors.
  Fig.~\ref{fig:experiment-flight-platform} shows the experimental platform and onboard control architecture.
\end{revblock}

\subsection{Experimental Setup}
1) \textit{Flight Platform}:
The BTO serves as the primary experimental platform.
\rev{The platform is powered by a 6S, 5050-mAh lithium-polymer battery.
  With the onboard perception and computation modules operating,
  the typical flight endurance is approximately 5 min under
  fault-free hovering conditions.}
The UTO is realized on the same platform by locking the outer tilt axes of all BTAUs.
All other airframe, sensing, computing, and power hardware remains unchanged.
\rev{For the UTO, the CL-PFTC allocation and actuator command mappings in \eqref{eq:cl-weighted-pseudoinverse-allocation} and \eqref{eq:cl-actuator-command-conversion}
are adapted to the uniaxial force model in \eqref{eq:aws-uto-ccu-force-models}, and the MATC correction in \eqref{eq:cl-matc-force-command} is omitted.
In AL-PFTC, the initial allocation matrix is adjusted accordingly.}

2) \textit{Flight-Control Unit}:
Both PFTC methods are deployed on a resource-constrained \textit{STM32H753} flight-control unit (FCU).
\rev{The controller layer generates the desired aerodynamic wrench,
while the control-allocation layer maps this desired wrench to the rotor-speed and biaxial-tilt commands.
Both layers are updated at $600~\mathrm{Hz}$}.

3) \textit{State Estimation}:
An \textit{NVIDIA Jetson Orin NX (16 GB)} module and a \textit{Livox Mid-360} LiDAR are mounted on the aircraft.
\rev{FAST-LIVO2\textcolor{black}{ \cite{zheng2025fastlivo2}} operates in LiDAR--inertial mode with visual updates disabled, providing onboard pose estimates at 10~Hz.}
\rev{These pose estimates are subsequently fused with measurements from the flight-controller IMU through an extended Kalman filter (EKF) to obtain the local state estimate.}

\begin{revblock}
4) \textit{Fault Injection}:
  Abrupt total rotor failures are experimentally emulated by issuing an active-braking command directly to the electronic
  speed controller (ESC) of the selected rotor at the prescribed fault-injection instant.
  Neither method uses the braking trigger in its controller or estimator computations.

  Table~\ref{tab:exp_computation_time} reports onboard execution times for the independently scheduled controller and allocation tasks, including mutual task preemption and interrupt overhead.
  All recorded maxima remain well below the common task period, leaving sufficient real-time margin on the tested STM32H753 FCU.
\end{revblock}
\begin{table}[!htb]
  \centering
  \caption{\rev{Onboard execution times of CL-PFTC and AL-PFTC on the STM32H753 flight controller}}
  \label{tab:exp_computation_time}
  \footnotesize
  \setlength{\tabcolsep}{3.2pt}
  \renewcommand{\arraystretch}{1.00}
  \begin{revtable}
    \begin{tabular}{@{}lcccc@{}}
      \toprule
      \multirow{2}{*}[-0.5ex]{Computation}
       & \multicolumn{2}{c}{CL-PFTC}
       & \multicolumn{2}{c}{AL-PFTC}                                             \\
      \cmidrule(lr){2-3}\cmidrule(lr){4-5}
       & Mean [$\mu$s]                & Max [$\mu$s]  & Mean [$\mu$s] & Max [$\mu$s] \\
      \midrule

      \multicolumn{5}{@{}l}{\textit{Controller layer}}                           \\
      \textbf{Layer total}
       & \textbf{110.49}             & \textbf{255}
       & \textbf{93.18}              & \textbf{144}                              \\
      \quad HOFA controller
       & 92.63                       & 132          & 93.18        & 144         \\
      \quad LESO
       & 17.86                       & 153          & --           & --          \\

      \midrule
      \multicolumn{5}{@{}l}{\textit{Allocation layer}}                           \\
      \textbf{Layer total}
       & \textbf{57.38}              & \textbf{145}
       & \textbf{123.00}             & \textbf{209}                              \\
      \quad Actuator processing
       & 51.65                       & 140          & 47.82        & 138         \\
      \quad Weighted PINV
       & 5.73                        & 16           & --           & --          \\
      \quad Adaptive allocation
       & --                          & --           & 75.18        & 92          \\
      \bottomrule
    \end{tabular}
  \end{revtable}
\end{table}
\begin{revblock}
  
For each comparison in the following experiments, identical HOFA gains, actuator limits, and command wrench bounds are used.
For concision, the following abbreviations are used:

$\mathrm{BTO}_{\mathrm{AL}}$: The BTO configuration using AL-PFTC.

$\mathrm{BTO}_{\mathrm{CL}}$: The BTO configuration using CL-PFTC.

$\mathrm{UTO}_{\mathrm{AL}}$: The UTO configuration using AL-PFTC.

$\mathrm{UTO}_{\mathrm{CL}}$: The UTO configuration using CL-PFTC.
\end{revblock}
\subsection{Basic Experiments}
\subsubsection{Hovering}
\begin{revblock}
  The four combinations of configuration and control method are evaluated under the selected rotor failures.
  For these experiments, each trial is evaluated using
  the following criterion:
  \begin{equation}
    \max_{t\in \mathcal T_h} e_p(t) < 0.5~\mathrm{m},
    \;
    \max_{t\in \mathcal T_h} e_\theta(t) \leq 150^\circ
    \label{eq:experiment_hovering_stability_condition}
  \end{equation}
  where $\mathcal T_h\triangleq[t_f,t_f+3~\mathrm{s}]$ and $t_f$ denotes the fault-injection instant.
  The available flight space determines the $0.5~\mathrm{m}$ position-error threshold.
  The $150^\circ$ attitude bound corresponds to $\varepsilon_q=\cos75^\circ$ in the theoretical nonsingularity condition.
  A checkmark $\checkmark$ indicates that \eqref{eq:experiment_hovering_stability_condition} is satisfied; otherwise, a cross $\times$ is shown.
\end{revblock}

\begin{table}[!t]
  \centering

  \caption{Hovering results under different rotor-failure conditions at a roll angle of $0^\circ$}
  \label{tab:experiment-horizontal-hovering-stability}
  \resizebox{0.95\columnwidth}{!}{%
    \renewcommand{\arraystretch}{1.3}
    \begin{tabular}{lccccccccc}
      \toprule
      Fault
       & $\overline{\boldsymbol{\Lambda}}_1$
       & $\overline{\boldsymbol{\Lambda}}_4$
       & $\overline{\boldsymbol{\Lambda}}_5$
       & $\overline{\boldsymbol{\Lambda}}_{1,2}$
       & $\overline{\boldsymbol{\Lambda}}_{3,4}$
       & $\overline{\boldsymbol{\Lambda}}_{1,3}$
       & $\overline{\boldsymbol{\Lambda}}_{1,6}$
       & $\overline{\boldsymbol{\Lambda}}_{3,6}$
       & $\overline{\boldsymbol{\Lambda}}_{1,3,6}$                                                     \\
      \midrule
      $\mathrm{BTO}_{\mathrm{CL}}$
       & \checkmark                                & \checkmark & \checkmark & \checkmark & \checkmark
       & $\times$                                  & $\times$   & $\times$   & $\times$                \\
      $\mathrm{BTO}_{\mathrm{AL}}$
       & \checkmark                                & \checkmark & \checkmark & \checkmark & \checkmark
       & \checkmark                                & \checkmark & \checkmark & \checkmark              \\
      $\mathrm{UTO}_{\mathrm{CL}}$
       & \checkmark                                & \checkmark & \checkmark & \checkmark & \checkmark
       & $\times$                                  & $\times$   & $\times$   & $\times$                \\
      $\mathrm{UTO}_{\mathrm{AL}}$
       & \checkmark                                & \checkmark & \checkmark & \checkmark & \checkmark
       & \checkmark                                & \checkmark & \checkmark & \checkmark              \\
      \bottomrule
    \end{tabular}%
  }

  \par\vspace{1.5ex}

  \caption{Hovering results under different rotor-failure conditions at a roll angle of $45^\circ$}
  \label{tab:experiment-tilted-hovering-stability}
  \resizebox{0.95\columnwidth}{!}{%
    \renewcommand{\arraystretch}{1.3}
    \begin{tabular}{lccccccccc}
      \toprule
      Fault
       & $\overline{\boldsymbol{\Lambda}}_1$
       & $\overline{\boldsymbol{\Lambda}}_4$
       & $\overline{\boldsymbol{\Lambda}}_5$
       & $\overline{\boldsymbol{\Lambda}}_{1,2}$
       & $\overline{\boldsymbol{\Lambda}}_{3,4}$
       & $\overline{\boldsymbol{\Lambda}}_{1,3}$
       & $\overline{\boldsymbol{\Lambda}}_{1,6}$
       & $\overline{\boldsymbol{\Lambda}}_{3,6}$
       & $\overline{\boldsymbol{\Lambda}}_{1,3,6}$                                                            \\
      \midrule
      $\mathrm{BTO}_{\mathrm{CL}}$
       & \checkmark                                & \checkmark & \checkmark
       & $\times$                                  & $\times$   & $\times$   & $\times$ & $\times$ & $\times$ \\
      $\mathrm{BTO}_{\mathrm{AL}}$
       & \checkmark                                & \checkmark & \checkmark
       & \checkmark                                & \checkmark & \checkmark
       & \checkmark                                & \checkmark & \checkmark                                  \\
      $\mathrm{UTO}_{\mathrm{CL}}$
       & $\times$                                  & $\times$   & \checkmark
       & $\times$                                  & $\times$   & $\times$   & $\times$ & $\times$ & $\times$ \\
      $\mathrm{UTO}_{\mathrm{AL}}$
       & $\times$                                  & \checkmark & \checkmark
       & $\times$                                  & \checkmark & $\times$
       & \checkmark                                & $\times$   & \checkmark                                  \\
      \bottomrule
    \end{tabular}%
  }
\end{table}

\begin{figure}[!tb]
  \centering
  \includegraphics[width=85mm]{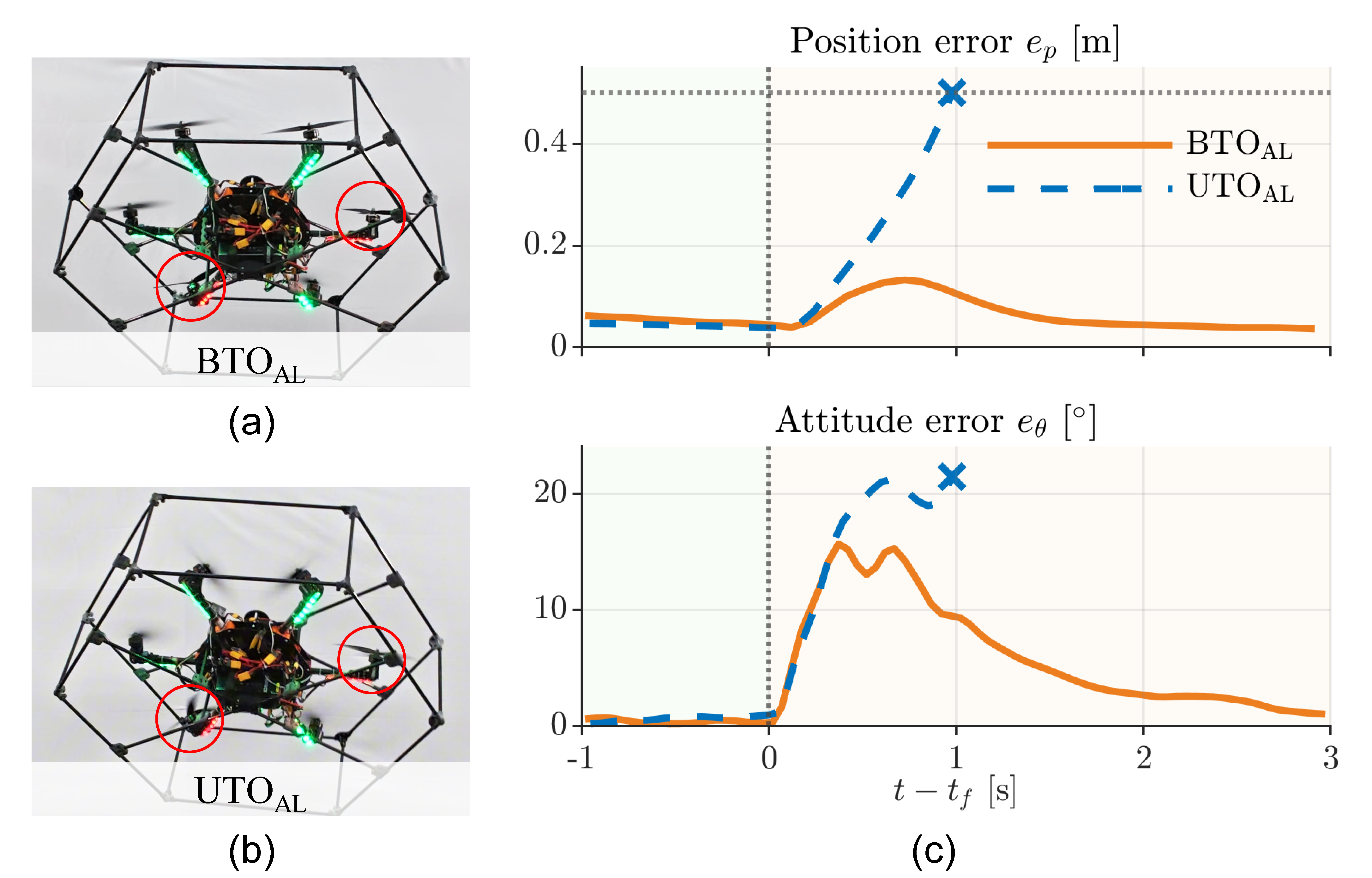}
  \caption{\rev{Hovering} recovery \rev{comparison} under
    \rev{$\overline{\boldsymbol{\Lambda}}_{1,3}$ at a roll angle of $45^\circ$}.
    (a) $\mathrm{BTO}_{\mathrm{AL}}$. (b) $\mathrm{UTO}_{\mathrm{AL}}$.
    \rev{(c) Tracking error curves.
    The vertical dotted lines indicate fault injection, the horizontal
    dotted line denotes the position-error threshold, and the crosses mark
    the termination of the UTO trial.}}
  \label{fig:experiment-hovering}
\end{figure}

\begin{revblock}
At $0^\circ$, both AL-PFTC combinations satisfy \eqref{eq:experiment_hovering_stability_condition} in all nine tested cases, whereas both CL-PFTC combinations do so in five.
CL-PFTC does not meet the criterion in the three tested double failures with failed rotors $120^\circ$ apart or in the triple failure $\overline{\boldsymbol\Lambda}_{1,3,6}$.
These four fault cases are retained by the task-based screening in Section~\ref{sec:aws-fault-analysis}.

At $45^\circ$, $\mathrm{BTO}_{\mathrm{AL}}$, $\mathrm{UTO}_{\mathrm{AL}}$, $\mathrm{BTO}_{\mathrm{CL}}$, and $\mathrm{UTO}_{\mathrm{CL}}$ satisfy the criterion in $9/9$, $5/9$, $3/9$, and $1/9$ tested cases, respectively.
Tilted hovering requires a sustained lateral force component in the body frame and changes actuator loading, increasing sensitivity to failed-rotor locations.
More BTO cases meet the criterion, consistent with greater freedom to redistribute the wrench using the additional tilt axis.

Figure~\ref{fig:experiment-hovering} shows the representative $\overline{\boldsymbol\Lambda}_{1,3}$ comparison at $45^\circ$.
With AL-PFTC, the BTO satisfies the recovery criterion, whereas the UTO exceeds the position-error bound and the trial is terminated.
\end{revblock}

\subsubsection{Indoor Trajectory Tracking}
\begin{figure}[!b]
  \centering
  \includegraphics[width=80mm]{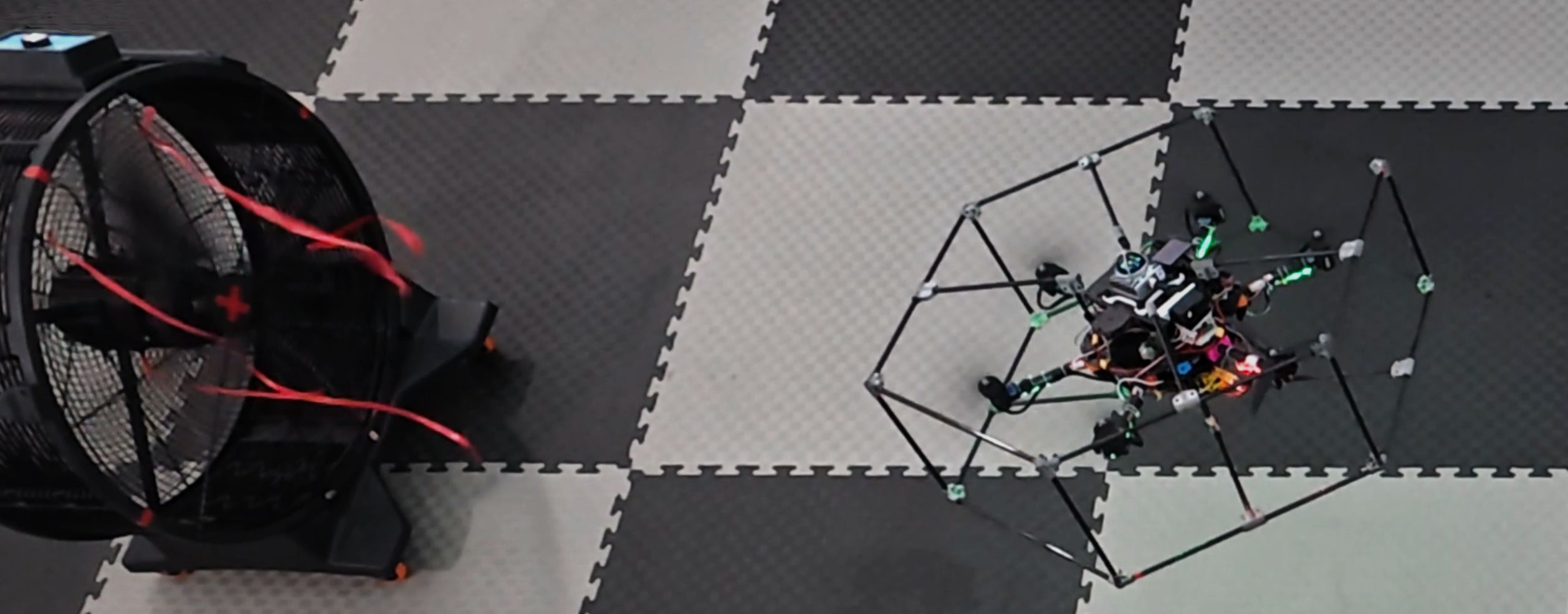}
  \caption{Experimental setup for indoor trajectory tracking.}
  \label{fig:experiment-indoor-experimental-scenario}

  \vspace{6pt}
  \includegraphics[width=85mm]{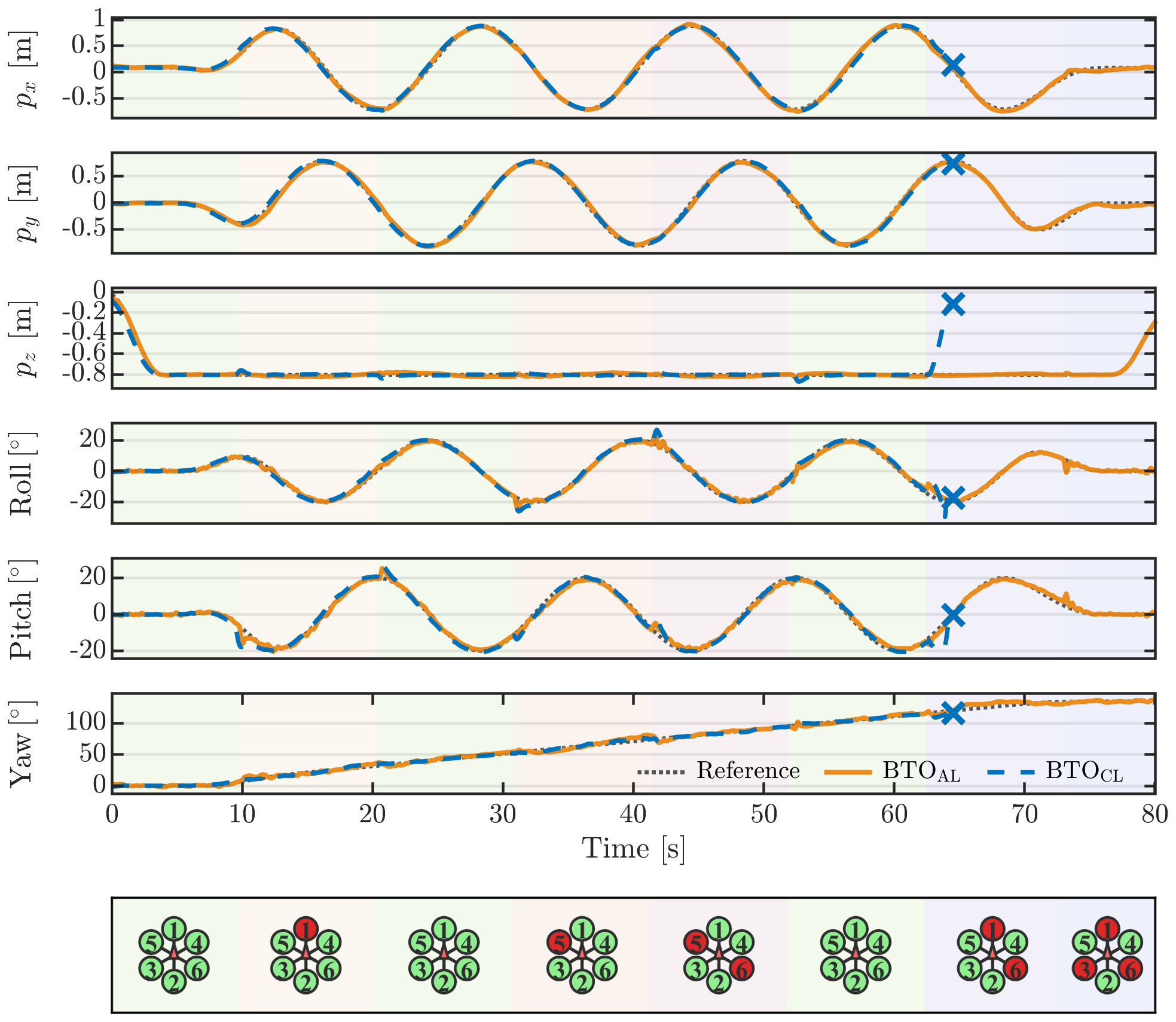}
  \caption{\rev{Indoor 6-DOF trajectory tracking results of $\mathrm{BTO}_{\mathrm{AL}}$ and $\mathrm{BTO}_{\mathrm{CL}}$
    under the prescribed rotor failure sequence.
    Crosses mark termination of the failed $\mathrm{BTO}_{\mathrm{CL}}$ trial and the bottom bar shows the fault stages.
    For visualization, the records are aligned at the trajectory start time.}}
  \label{fig:experiment-indoor-platform-comparison}
\end{figure}
\begin{revblock}
  This experiment compares the proposed AL-PFTC method with the CL-PFTC reference on the same BTO platform during 6-DOF trajectory tracking.
  Both methods follow the trajectory form in \eqref{eq:simulation-reference-trajectory}, with $\omega_1=\omega_2=\tfrac{\pi}{8}~\mathrm{rad/s}$,
  under the following prescribed sequence of braking conditions:
  \[
    \mathbf{FF}
    \rightarrow
    \overline{\boldsymbol{\Lambda}}_1
    \rightarrow
    \mathbf{FF}
    \rightarrow
    \overline{\boldsymbol{\Lambda}}_5
    \rightarrow
    \overline{\boldsymbol{\Lambda}}_{5,6}
    \rightarrow
    \mathbf{FF}
    \rightarrow
    \overline{\boldsymbol{\Lambda}}_{1,6}
    \rightarrow
    \overline{\boldsymbol{\Lambda}}_{1,3,6}
  \]
  where $\mathbf{FF}$ denotes that no active-braking command is issued.
  Each stage has a nominal duration of $10~\mathrm{s}$, giving a planned trial duration of $80~\mathrm{s}$.
  The experimental scenario is shown in Fig.~\ref{fig:experiment-indoor-experimental-scenario}.
  All indoor trajectory trials are independently conducted under approximately $5~\mathrm{m/s}$ lateral airflow.

  Fig.~\ref{fig:experiment-indoor-platform-comparison} compares the tracking responses of $\mathrm{BTO}_{\mathrm{AL}}$ and $\mathrm{BTO}_{\mathrm{CL}}$.
  Both maintain tracking through $\overline{\boldsymbol{\Lambda}}_{5,6}$ and the subsequent $\mathbf{FF}$ stage.
  Following the application of $\overline{\boldsymbol{\Lambda}}_{1,6}$, $\mathrm{BTO}_{\mathrm{CL}}$ departs rapidly from the reference and its trial is terminated.
  In the AL-PFTC trial, the BTO continues tracking through the final $\overline{\boldsymbol{\Lambda}}_{1,3,6}$ stage and completes the prescribed $80~\mathrm{s}$ trial.

\end{revblock}
\begin{figure}[!tb]
  \centering

  \begin{minipage}{\linewidth}
    \centering
    \footnotesize
    \makeatletter
    \def\@captype{table}
    \makeatother
    \caption{\rev{RMSEs of tracking and filtered wrench errors for $\mathrm{BTO}_{\mathrm{AL}}$}}
    \label{tab:experiment-indoor-bto-performance}
    \renewcommand{\arraystretch}{1.00}
    \begin{revtable}
    \begin{tabular}{lcc}
      \toprule
      Metric
       & Fault-free
       & Fault sequence \\
      \midrule

      $\mathcal{R}_{\mathcal T_{s3}}(e_p)$ [m]
       & 0.0176
       & 0.0403         \\

      $\mathcal{R}_{\mathcal T_{s3}}(e_\theta)$ [deg]
       & 1.34
       & 3.20           \\

      $\mathcal{R}_{\mathcal T_{s3}}(\|\boldsymbol e_{w,F}\|_2)$ [N]
       & 0.868
       & 2.366          \\

      $\mathcal{R}_{\mathcal T_{s3}}(\|\boldsymbol e_{w,T}\|_2)$ [$\mathrm{N\cdot m}$]
       & 0.215
       & 0.683          \\

      \bottomrule
    \end{tabular}%
    \end{revtable}
  \end{minipage}

  \vspace{6pt}
  \includegraphics[width=80mm]{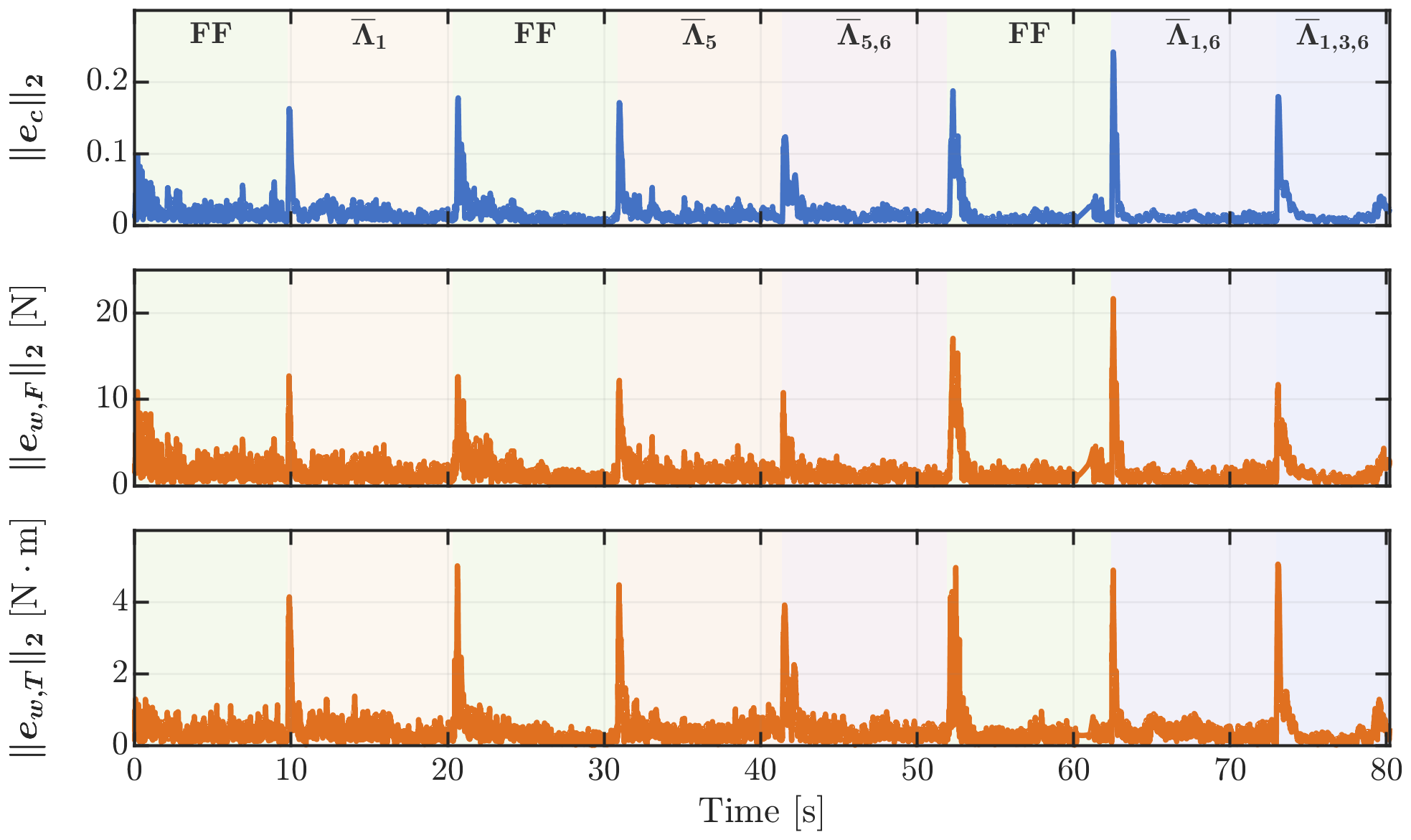}
  \caption{\rev{Allocation-layer responses} of $\mathrm{BTO}_{\mathrm{AL}}$
    \rev{during} indoor \rev{6-DOF} trajectory tracking \rev{under the prescribed
    rotor-failure sequence}.}
  \label{fig:experiment-indoor-bto-allo-state}

\end{figure}

\begin{figure}[!tb]
  \centering
  \includegraphics[width=80mm]{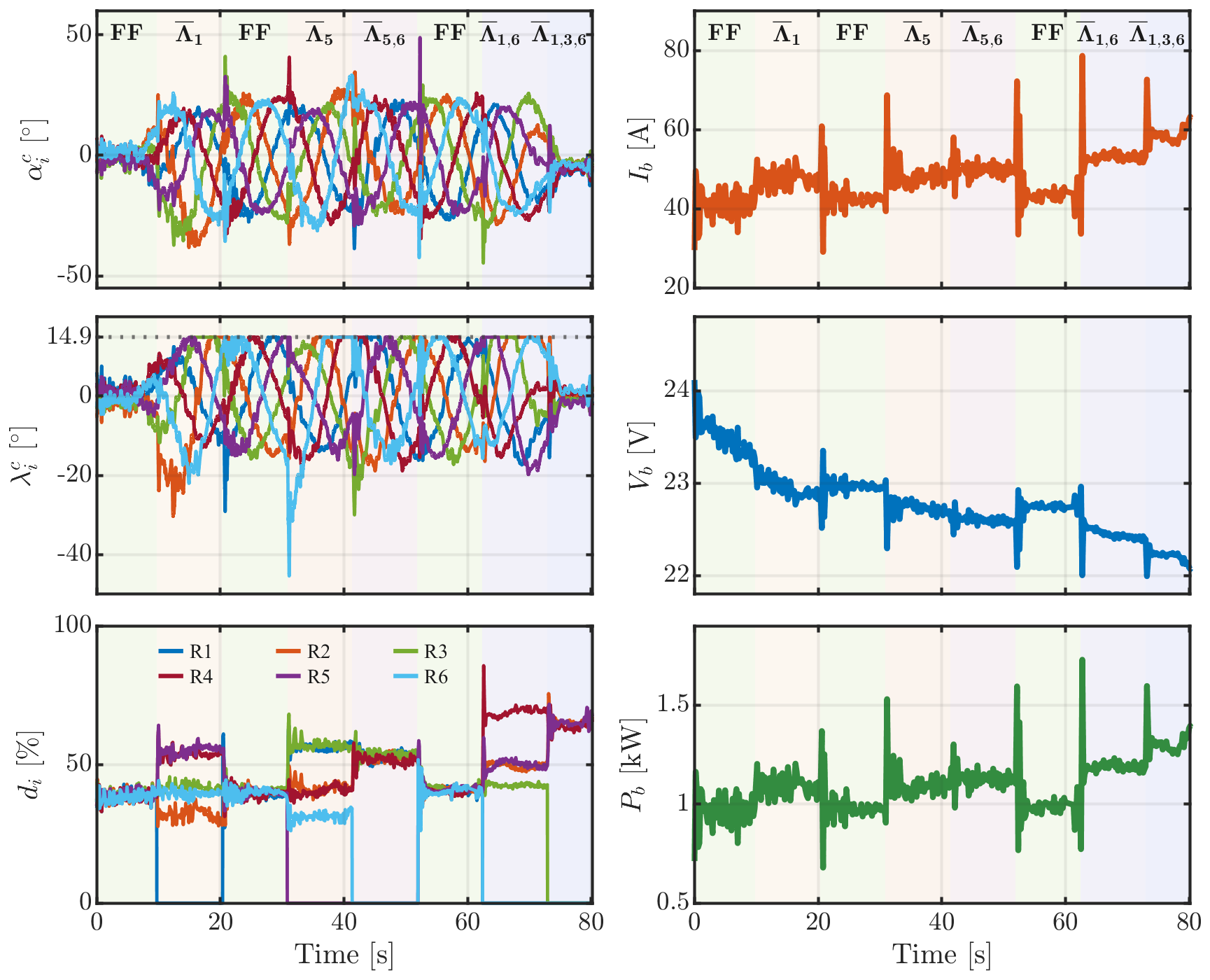}
  \caption{Actuator \rev{and electrical responses during the indoor flight.
  From top to bottom, the left column shows inner and outer servo
  angle} commands \rev{and motor duty;
  the right column shows} battery \rev{current, voltage, and power}.}
  \label{fig:experiment-indoor-bto-actuator-power-combined}
\end{figure}
\begin{revblock}
  Figs.~\ref{fig:experiment-indoor-bto-allo-state} and \ref{fig:experiment-indoor-bto-actuator-power-combined} provide supporting diagnostics for the $\mathrm{BTO}_{\mathrm{AL}}$ trial.
  Let $\boldsymbol e_{w,F}$ and $\boldsymbol e_{w,T}$ denote the force and torque components of $\boldsymbol e_{w,\mathrm{lp}}$, respectively.
  The virtual allocation error $\|\boldsymbol e_c\|_2$ and the filtered wrench errors $\|\boldsymbol e_{w,F}\|_2$ and $\|\boldsymbol e_{w,T}\|_2$ exhibit transient increases near changes in the braking commands and subsequently decrease.
  The outer tilt commands $\lambda_i^c$ repeatedly reach the imposed limit of $0.26~\mathrm{rad}$ ($14.9^\circ$), while the vehicle continues tracking through the final stage.

  Table~\ref{tab:experiment-indoor-bto-performance} compares RMSEs of tracking and filtered wrench errors for $\mathrm{BTO}_{\mathrm{AL}}$ under the prescribed braking sequence and in a separate fault-free trial.
  Both records are evaluated over $\mathcal T_{s3}=[0,80]~\mathrm{s}$.
  The position and attitude RMSEs are $0.0176~\mathrm{m}$ and $1.34^\circ$ in the fault-free trial, and $0.0403~\mathrm{m}$ and $3.20^\circ$ under the braking sequence.
\end{revblock}
\subsubsection{Outdoor Figure-8 Tracking}
\begin{figure}[!tb]
  \centering
  \includegraphics[width=80mm]{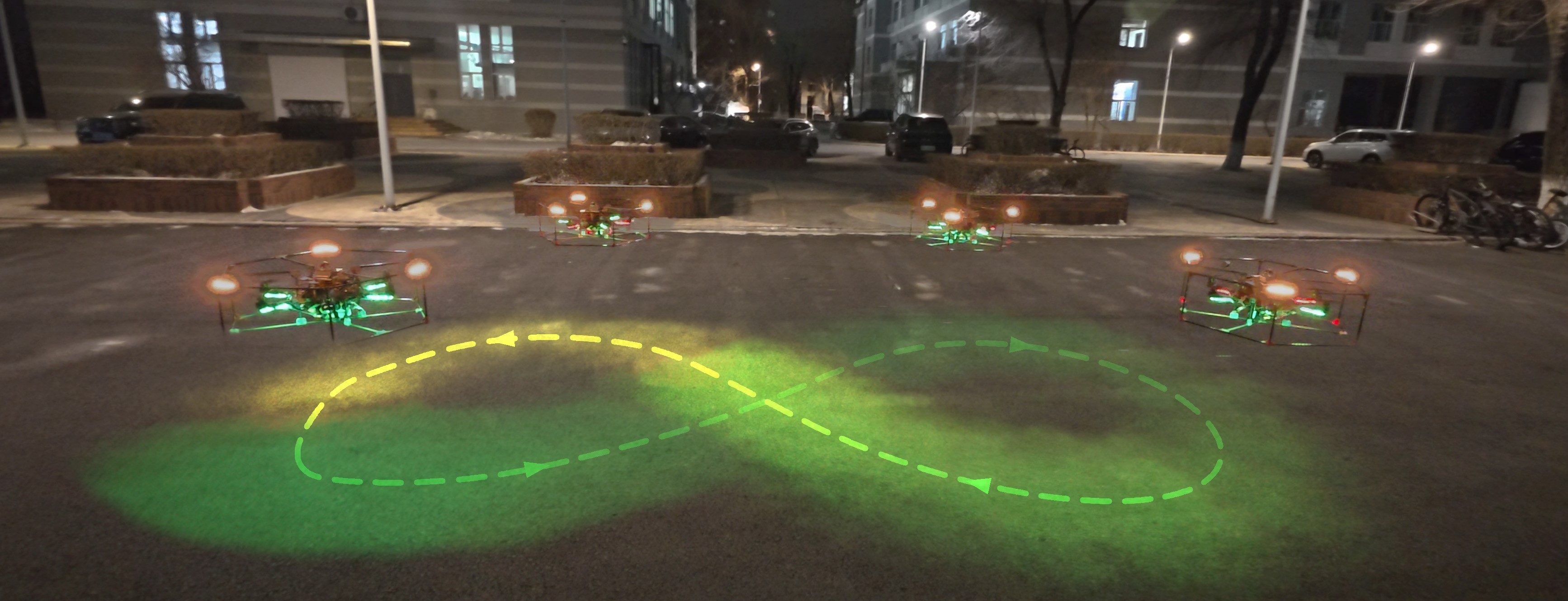}
  \caption{\rev{Multi-frame composite} of the outdoor \rev{Figure-8 trajectory-tracking} experiment \rev{under successive} rotor failures.
    \rev{The dashed curve and arrowheads show the flight path and direction}.}
  \label{fig:experiment-outdoor-figure-eight-tracking-real-photo}

  \vspace{6pt}
  \includegraphics[width=80mm]{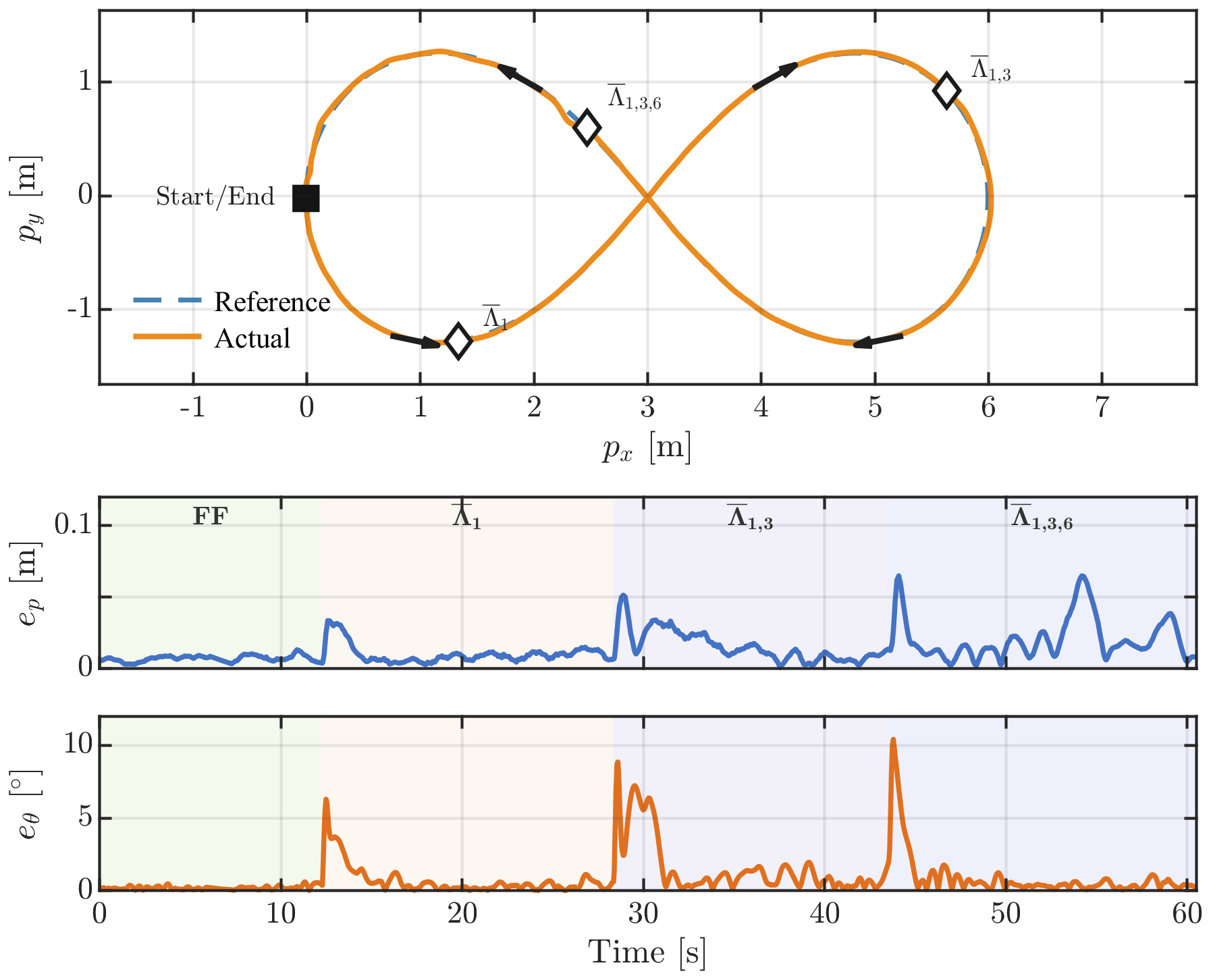}
  \caption{\rev{Outdoor Figure-8 tracking of
    $\mathrm{BTO}_{\mathrm{AL}}$ under successive rotor failures.
    The top panel compares the reference trajectory with the trajectory estimated onboard;
    arrows indicate the flight direction, and diamonds
    mark the fault-injection locations.
    The middle and bottom panels show the position and attitude
    tracking errors, respectively.}}
  \label{fig:experiment-outdoor-figure-eight-tracking}
\end{figure}

As shown in Fig.~\ref{fig:experiment-outdoor-figure-eight-tracking-real-photo}, to further evaluate $\mathrm{BTO}_{\mathrm{AL}}$ under realistic outdoor conditions, 
an outdoor Figure-8 tracking experiment is conducted
with wind speeds of $3.4$--$5.4~\mathrm{m/s}$ and \rev{a forecast
temperature of approximately $-20\,^{\circ}\mathrm{C}$}.
\rev{Low temperature increases air density, alters propeller loading, and affects battery, actuator, and sensing behavior.
Together with wind, these effects are treated as combined environmental and
electromechanical uncertainties.}

The successive fault conditions are:
\[
  \mathbf{FF}
  \rightarrow
  \overline{\boldsymbol{\Lambda}}_1
  \rightarrow
  \overline{\boldsymbol{\Lambda}}_{1,3}
  \rightarrow
  \overline{\boldsymbol{\Lambda}}_{1,3,6}
\]
\rev{with fault injections marked by diamonds in Fig.~\ref{fig:experiment-outdoor-figure-eight-tracking}.
The reference is generated offline,
and the actual trajectory is obtained from the onboard state estimate.}

As shown in Fig.~\ref{fig:experiment-outdoor-figure-eight-tracking},
$\mathrm{BTO}_{\mathrm{AL}}$ completes the entire Figure-8 trajectory under the successive rotor-failure conditions.
\rev{Over the evaluation interval $\mathcal T_{s4}=[0,60]~\mathrm{s}$, the resulting tracking errors are
      $\mathcal{R}_{\mathcal T_{s4}}(e_p)=0.0178~\mathrm{m}$ and $\mathcal{R}_{\mathcal T_{s4}}(e_\theta)=1.70^\circ$.
        No persistent tracking drift is observed after the transition to
      $\overline{\boldsymbol{\Lambda}}_{1,3,6}$, indicating that
      $\mathrm{BTO}_{\mathrm{AL}}$ retains bounded trajectory-tracking
      performance under the tested conditions of low temperature, wind, and
      multiple rotor failures.}

  \subsection{Flight Task Experiments}
  We assess the mission-level fault tolerance of $\mathrm{BTO}_{\mathrm{AL}}$ through two autonomous tasks:
  traversal through a narrow frame at nonzero roll and contact-based aerial writing.
  Beyond hovering and trajectory tracking, these tasks introduce geometric clearance or contact constraints and unmodeled disturbances.

  \begin{figure}[!tb]
    \centering
    \includegraphics[width=80mm]{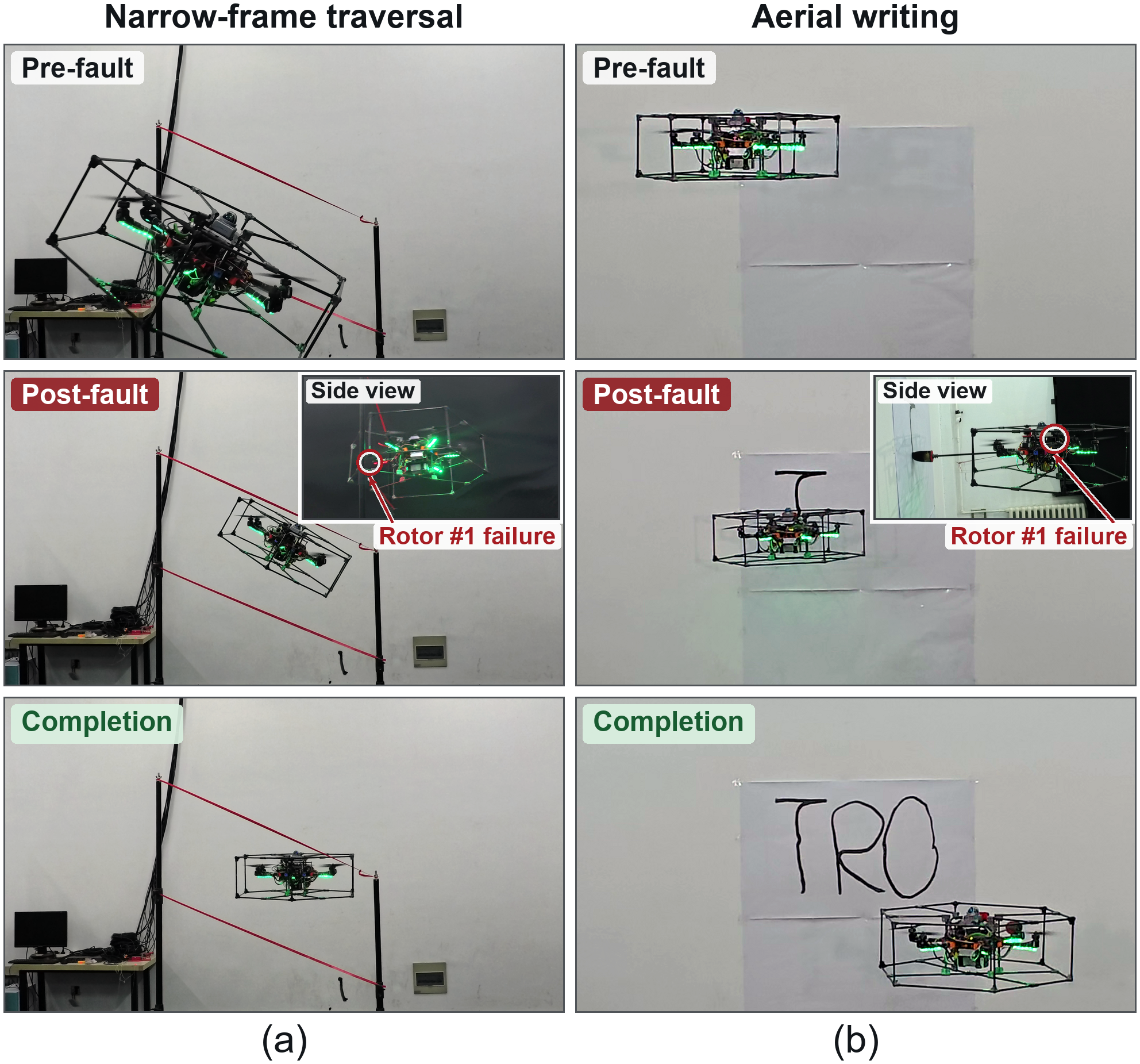}
    \caption{\rev{Task-level experiments with $\mathrm{BTO}_{\mathrm{AL}}$ under an abrupt total failure of rotor~\#1.
      (a) Traversal through a narrow frame at nonzero roll.
      (b) Contact-based aerial writing on a vertical surface.
      Rows show the pre-fault, post-fault, and completion stages, and the middle-row insets identify the failed rotor.}}
    \label{fig:experiment-flight-task-real-photo}
  \end{figure}
  \subsubsection{Traversal}

  In Fig.~\ref{fig:experiment-flight-task-real-photo}(a), $\mathrm{BTO}_{\mathrm{AL}}$ enters the frame at a prescribed $20^\circ$ roll with all rotors operational.
  The opening measures $1.0~\mathrm{m}\times0.4~\mathrm{m}$.

  \begin{revblock}
    Before the vehicle clears the opening, rotor~\#1 is actively braked to impose $\overline{\boldsymbol{\Lambda}}_1$ and remains inactive through completion.
    This timing tests fault accommodation while the attitude and geometric clearance constraints are simultaneously active.
  \end{revblock}

  \rev{Over the full traversal interval $\mathcal T_{s5}$, the position and attitude RMSEs are $0.017~\mathrm{m}$ and $2.0^\circ$, respectively.}
  Despite the in-frame rotor failure, $\mathrm{BTO}_{\mathrm{AL}}$ maintains position and attitude tracking and completes the traversal without task interruption.
  \subsubsection{Aerial Writing}

  Here, $\mathrm{BTO}_{\mathrm{AL}}$ autonomously executes an approach--write--return sequence to draw the target shape on a vertical wall.
\begin{revblock}
The fault-injection trial begins with all rotors operational.
After writing begins but before the target shape is completed, 
rotor~\#1 is disabled by active braking and remains inactive throughout the remaining writing and return phases.
\end{revblock}

The task is subject to two main sources of unknown and time-varying disturbance:

\textit{Aerodynamic proximity disturbance}:
        The flight task is performed approximately $0.5~\mathrm{m}$ from the wall, where wall-induced aerodynamic effects become non-negligible and may degrade flight stability and tracking accuracy.
  
\begin{figure}[!htb]
  \centering
  \includegraphics[width=80mm]{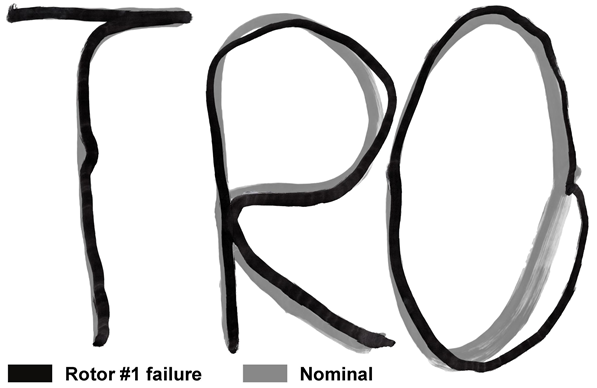}
  \caption{Aerial writing outcomes for $\mathrm{BTO}_\mathrm{AL}$ in the fault-free trial (gray) and the fault-injection trial ($\overline{\boldsymbol{\Lambda}}_1$, black).}
  \label{figAerialWritingResult}
\end{figure}

\textit{Contact-induced friction disturbance}:
  The writing tool is made of soft sponge material and no force sensor is installed.
  As a result, the contact force and friction are unmeasured and time-varying.
  In addition, the pen-up and pen-down operations introduce abrupt changes in the contact disturbance.
\begin{revblock}
  
  Performance is assessed through vehicle position tracking and the completed ink pattern.
  Over the complete approach--write--return interval
  $\mathcal{T}_{s6}$, the vehicle position RMSE
  $\mathcal{R}_{\mathcal{T}_{s6}}(e_p)$ is $0.0070~\mathrm{m}$
  in the fault-free trial and $0.0074~\mathrm{m}$ in the fault-injection trial with
  $\overline{\boldsymbol{\Lambda}}_1$.
  These RMSEs quantify vehicle position tracking, not pen-tip or
  ink-pattern accuracy, and indicate comparable tracking over
  both complete trials.

  The completed ink regions are compared as spatial point sets
  without temporal matching.
  Let $A_{\mathrm F}$ and $A_{\mathrm N}$ denote the segmented
  wall-plane samples under $\overline{\boldsymbol{\Lambda}}_1$ and
  fault-free conditions, respectively.
  Both sets contain the coordinates of all segmented ink pixels,
converted to meters using the measured dimensions and aligned
by translation at the upper-left starting positions.
  With
  $D(A,B)\triangleq
  \left(\min_{b\in B}\|a-b\|_2\right)_{a\in A}$
  denoting the directed nearest-neighbor distance samples, define the symmetric Hausdorff distance (HD) and the 95th-percentile Hausdorff distance, $\mathrm{HD}_{95}$, as
  \begin{equation}
    \begin{aligned}
      \mathrm{HD}
       & \!=\!
      \max\!
      \left\{
      \max D(A_{\mathrm F},A_{\mathrm N}),
      \max D(A_{\mathrm N},A_{\mathrm F})
      \right\}
      \\
      \mathrm{HD}_{95}
       & \!=\!
      \max\!
      \left\{
      Q_{0.95}\!\left(D(A_{\mathrm F},A_{\mathrm N})\right),
      Q_{0.95}\!\left(D(A_{\mathrm N},A_{\mathrm F})\right)
      \right\}
    \end{aligned}
    \label{eq:experiment-writing-image-metrics}
  \end{equation}
  Here, $Q_{0.95}$ is the empirical 95th percentile.
  $\mathrm{HD}$ measures the maximum discrepancy, whereas
  $\mathrm{HD}_{95}$ is less sensitive to
  isolated segmentation noise and local ink irregularities.

  The resulting $\mathrm{HD}=0.0744~\mathrm{m}$ and
  $\mathrm{HD}_{95}=0.0437~\mathrm{m}$ quantify the ink-pattern
  discrepancy.
  Figure~\ref{figAerialWritingResult} shows that the overall ``TRO''
  geometry is retained, with the largest deviations near the lower
  oblique stroke of ``R'' and the right curved segment of ``O''.
  Together with the comparable vehicle position RMSE, these results show
  completion of the tested contact-based writing task under the
  sustained failure $\overline{\boldsymbol{\Lambda}}_1$, with
  predominantly localized ink-pattern differences.
\end{revblock}

\subsection{Discussion}
\begin{revblock}
  The hovering comparisons in
  Tables~\ref{tab:experiment-horizontal-hovering-stability}
  and \ref{tab:experiment-tilted-hovering-stability}
  show that the same screened fault cases can yield different
  closed-loop outcomes.
  At horizontal hover, AL-PFTC satisfies the criterion in more tested cases than CL-PFTC on both platforms.
  At $45^\circ$, this advantage persists on each platform, while
  the BTO satisfies the criterion in more cases than the UTO
  under both methods.
  With CL-PFTC, the compensating commands still pass through the
  nominal allocation map after rotor effectiveness changes.
  AL-PFTC instead adjusts the command distribution using wrench
  feedback, providing a mechanism consistent with the different
  outcomes on the same platform.
  Configuration differences become more evident under
  nonzero-attitude demands.
  The indoor comparison on the same BTO in
  Fig.~\ref{fig:experiment-indoor-platform-comparison}
  extends the method comparison to continuous tracking under
  successive failures.
  The smaller residual wrench errors in
  Fig.~\ref{fig:simulation-method-comparison}
  provide complementary support for this interpretation.

  The traversal and writing trials place fault accommodation in
  operations whose motion requirements persist after failure.
  In traversal, the fault is imposed before the vehicle clears the
  opening, requiring continued attitude tracking within the available
  geometric clearance.
  In writing, rotor loss occurs during wall interaction, with
  unmeasured contact forces contributing to the combined wrench
  defined in \eqref{eq:al-ext-wrench-definition}.
  The allocation update uses the corresponding filtered wrench error
  in \eqref{eq:al-filtered-output-wrench-error} without separately
  identifying the fault and contact contributions.
  These trials therefore extend the evaluation from hover recovery
  to maintaining prescribed motion under geometric and contact constraints.
\end{revblock}
\section{Conclusion}
\begin{revblock}
This work developed AL-PFTC to support continued independent position
and attitude control after abrupt total rotor failures without explicit
fault identification.
The method adapts the command distribution using external-wrench
estimates obtained from onboard motion measurements.
Wrench-reserve analysis at horizontal hover defined a fault set for the
adaptive design, and common offline parameter bounds were constructed
for the corresponding effectiveness set.
Accounting for estimator residuals and filter dynamics, the analysis
yielded an explicit sufficient bandwidth condition for allocation-error
dissipation.
Under the stated conditions, the joint analysis further established
local UUB for constant-reference pose regulation in the final fixed
fault mode.

Experiments on the reconfigurable hexacopter showed that AL-PFTC
satisfied the prescribed hovering criterion in more tested fault cases
than CL-PFTC in both the BTO and UTO configurations.
At $45^\circ$ roll, the BTO met this criterion in more cases than the
UTO under both methods.
In the indoor comparison, the BTO with AL-PFTC completed the prescribed
6-DOF tracking trial under wind disturbance, whereas the CL-PFTC trial
was terminated.
Further AL-PFTC trials demonstrated outdoor tracking under successive
failures, as well as narrow-frame traversal at nonzero roll and
contact-based aerial writing under unmeasured contact disturbances.
In both task demonstrations, a total rotor failure was injected during
execution and sustained until completion.
The AL-PFTC controller and allocator operated at 600~Hz on the
STM32H753 without online optimization.

These results support adaptive allocation as an effective means of
using retained actuation capability for continued task execution under
the tested conditions.
Future work will integrate AL-PFTC with environment-aware onboard
perception to support task execution in dynamic and unstructured
environments.
\end{revblock}

\begingroup
\ifnum\SHOWCHANGES=1
  \makeatletter
  \newcommand{\marknewreference}[1]{\@namedef{newreference@#1}{}}
  \marknewreference{sontagCharacterizationsInputtoState1995}
  \marknewreference{hindiAnalysisLinearSystems1998}
  \marknewreference{huAnalysisDesignSaturation2002}
  \marknewreference{falconi2016hexacopter}
  \marknewreference{du2023distributed}
  \marknewreference{wang2018adaptive}
  \marknewreference{shen2015inertiafree}
  \marknewreference{shen2017robust}
  \newcommand{\referencecolor}[1]{%
    \par
    \ifcsname newreference@#1\endcsname
      \color{red}%
    \else
      \color{black}%
    \fi
  }
  \let\originalbibitem\@bibitem
  \def\@bibitem#1{\referencecolor{#1}\originalbibitem{#1}}
  \let\originallbibitem\@lbibitem
  \def\@lbibitem[#1]#2{\referencecolor{#2}\originallbibitem[#1]{#2}}
  \makeatother
\fi
\bibliographystyle{IEEEtran}
\bibliography{Section1} 
\endgroup

\end{document}